%% file: main.tex
\def\year{2019}\relax
\documentclass[letterpaper]{article} %DO NOT CHANGE THIS
\usepackage{aaai19}  %Required
\usepackage{times}  %Required
\usepackage{helvet}  %Required
\usepackage{courier}  %Required
\usepackage{url}  %Required
\usepackage{graphicx}  %Required
\usepackage{algorithm,algorithmic}
\input{macros.tex} %input not allowed!
\begin{document}
\title{Iterative Classroom Teaching}
\author{
Teresa Yeo \\ LIONS, EPFL \\ teresa.yeo@epfl.ch
\And 
Parameswaran Kamalaruban \\ LIONS, EPFL \\ kamalaruban.parameswaran@epfl.ch
\And 
Adish Singla \\ MPI-SWS  \\ adishs@mpi-sws.org  %\\ Saarbr{\"u}cken, Germany
\And
Arpit Merchant \\ MPI-SWS \\ arpitdm@mpi-sws.org
\AND 
Thibault Asselborn \\ CHILI Lab, EPFL \\ thibault.asselborn@epfl.ch
\And 
Louis Faucon \\ CHILI Lab, EPFL \\ louis.faucon@epfl.ch
\And 
Pierre Dillenbourg \\ CHILI Lab, EPFL \\ pierre.dillenbourg@epfl.ch
\And 
Volkan Cevher \\ LIONS, EPFL \\ volkan.cevher@epfl.ch}
\maketitle

%%%%%%%%%%%%%%%%%%%%%%%%%%%%%%%%%%%%%%%%%%%%%%%%%%%%%%%%%
%%%%%%%%%%%%%%%%%%%%%%%%%%%%%%%%%%%%%%%%%%%%%%%%%%%%%%%%%
\input{0_abstract}
\input{1_introduction}
\input{2_model}
\input{3_average-learner}
\input{4_worst-learner}
\input{5_partition}
\input{6_experiments}

\input{7_conclusion}

\vspace{2mm}
{\bfseries Acknowledgments.} This work was supported in part by the Swiss National Science Foundation (SNSF) under grant number 407540\_167319, CR21I1\_162757 and NCCR Robotics.

%%%%%%%%%%%%%%%%%%%%%%%%%%%%%%%%%%%%%%%%%%%%%%%%%%%%%%%%%
\bibliography{refs}
\bibliographystyle{aaai}

%%%%%%%%%%%%%%%%%%%%%%%%%%%%%%%%%%%%%%%%%%%%%%%%%%%%%%%%%
\onecolumn
\appendix
\input{8.1_appendix_additional-robust-settings}

\input{8.2_appendix_proofs}
\input{8.3_appendix_rescalable-pool-based-teaching}

%%%%%%%%%%%%%%%%%%%%%%%%%%%%%%%%%%%%%%%%%%%%%%%%%%%%%%%%%
\end{document}

%% file: macros.tex
\usepackage{microtype}
\usepackage{graphicx}
\usepackage{subcaption}
\usepackage{booktabs} % for professional tables

\usepackage{hyperref}

\usepackage{amsfonts}       % blackboard math symbols
\usepackage{amsmath}
\usepackage{amsthm}

\usepackage{enumitem}
\newcommand{\bc}[1]{\left\{{#1}\right\}}
\newcommand{\br}[1]{\left({#1}\right)}
\newcommand{\bs}[1]{\left[{#1}\right]}
\newcommand{\abs}[1]{\left| {#1} \right|}

\newcommand{\floor}[1]{\left\lfloor #1 \right\rfloor}

\newcommand{\ip}[2]{\left\langle{#1},{#2}\right\rangle}
\newcommand{\norm}[1]{\left\| {#1} \right\|}

\newtheorem{theorem}{Theorem}

\newtheorem{remark}{Remark}

\allowdisplaybreaks

\makeatletter
\newtheorem*{rep@theorem}{\rep@title}
\newcommand{\newreptheorem}[2]{%
\newenvironment{rep#1}[1]{%
 \def\rep@title{#2 \ref{##1}}%
 \begin{rep@theorem}}%
 {\end{rep@theorem}}}
\makeatother

\newreptheorem{lemma}{Lemma}
\newreptheorem{theorem}{Theorem}
\newreptheorem{claim}{Claim}
\newreptheorem{proposition}{Proposition}

\DeclareMathOperator*{\argmin}{arg\,min}
\DeclareMathOperator*{\argmax}{arg\,max}

%% file: 0_abstract.tex
%!TEX root = main.tex
%%%%%%%%%%%%%%%%%%%%%%%%%%%%%%%%%%%%%%%%%%%%%%%%%%%%%%%%%
%%%%%%%%%%%%%%%%%%%%%%%%%%%%%%%%%%%%%%%%%%%%%%%%%%%%%%%%%
\begin{abstract}
We consider the machine teaching problem in a classroom-like setting wherein the teacher has to deliver the same examples to a diverse group of students. Their diversity stems from differences in their initial internal states as well as their learning rates. We prove that a teacher with full knowledge about the learning dynamics of the students can teach a target concept to the entire classroom using $\mathcal{O} \br{\min\bc{d,N} \log \frac{1}{\epsilon}}$ examples, where $d$ is the ambient dimension of the problem, $N$ is the number of learners, and $\epsilon$ is the accuracy parameter. We show the robustness of our teaching strategy when the teacher has limited knowledge of the learners' internal dynamics as provided by a noisy oracle. Further, we study the trade-off between the learners' workload and the teacher's cost in teaching the target concept. Our experiments validate our theoretical results and suggest that appropriately partitioning the classroom into homogenous groups provides a balance between these two objectives.
\end{abstract}

%% file: 1_introduction.tex
\vspace{-3mm}
\section{Introduction}\label{sec.introduction}

%********* new-introduction ********* \\
Machine teaching considers the inverse problem of machine learning. Given a learning model and a target, the teacher aims to find an optimal set of training examples for the learner \cite{zhu2018overview,liu2017iterative}. Machine teaching provides a rigorous formalism for various real-world applications such as personalized education and intelligent tutoring systems \cite{rafferty2016faster,patil2014optimal}, imitation learning \cite{cakmak2012algorithmic,haug2018teachingrisk}, program synthesis \cite{mayer2017proactive}, adversarial machine learning \cite{mei2015using}, and human-in-the-loop systems \cite{singla2014near,singla2013actively}.\footnote{\url{http://teaching-machines.cc/nips2017/}}

\vspace{-2.5mm}
\paragraph{Individual teaching}
Most of the research in this domain thus far, has focused on teaching a single student in the batch setting. Here, the teacher constructs an optimal training set (e.g., of minimum size) for a fixed learning model and a target concept and gives it to the student in a single interaction \cite{goldman1995complexity,zilles2011models,zhu2013machine,doliwa2014recursive}. Recently, there has been interest in studying the interactive setting \cite{liu2017iterative,zhu2018overview,chen2018understanding,hunziker2018teachingmultiple}, wherein the teacher focuses on finding an optimal sequence of examples to meet the needs of the student under consideration, which is, in fact, the natural expectation in a personalized teaching environment \cite{koedinger1997intelligent}. \cite{liu2017iterative} introduced the iterative machine teaching setting wherein the teacher has full knowledge of the internal state of the student at every time step using which she designs the subsequent optimal example. They show that such an ``omniscient" teacher can help a single student approximately learn the target concept using $\mathcal{O} \br{\log \frac{1}{\epsilon}}$ training examples (where $\epsilon$ is the accuracy parameter) as compared to $\mathcal{O} \br{\frac{1}{\epsilon}}$ examples chosen randomly by the stochastic gradient descent (SGD) teacher.
%cakmak2014eliciting,

\vspace{-2.5mm}
\paragraph{Classroom teaching}
In real-world classrooms, the teacher is restricted to providing the same examples to a large class of academically-diverse students. Customizing a teaching strategy for a specific student may not guarantee optimal performance of the entire class. Alternatively, teachers may constitute a partitioning of the students so as to maximize intra-group homogeneity while balancing the orchestration costs of managing parallel activities. \cite{Zhu} propose methods for explicitly constructing a minimal training set for teaching a class of batch learners based on a minimax teaching criterion. They also study optimal class partitioning based on prior distributions of the learners. However, they do not consider an interactive teaching setting.

\begin{figure*}[t!]
    \captionsetup[subfigure]{font=scriptsize,labelfont=scriptsize}
    \centering
    \begin{subfigure}[t]{0.25\textwidth}
        \includegraphics[width=\textwidth]{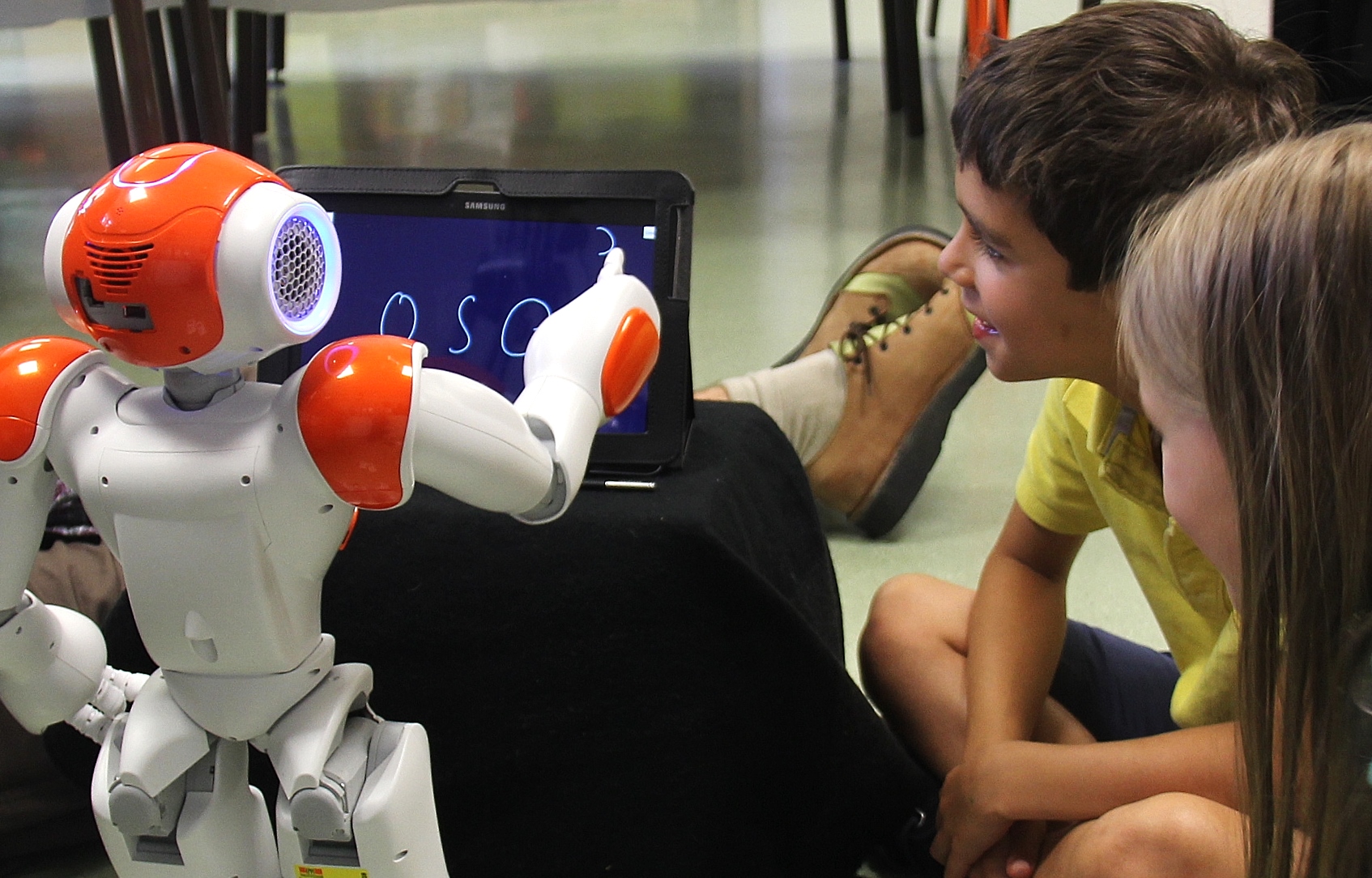}
        \caption{A Nao robot writing on a digital tablet.}
        \label{fig-cowriter}
    \end{subfigure}
    \qquad \qquad
    \begin{subfigure}[t]{0.50\textwidth}
    	\centering
        \includegraphics[width=\textwidth,height=3cm]{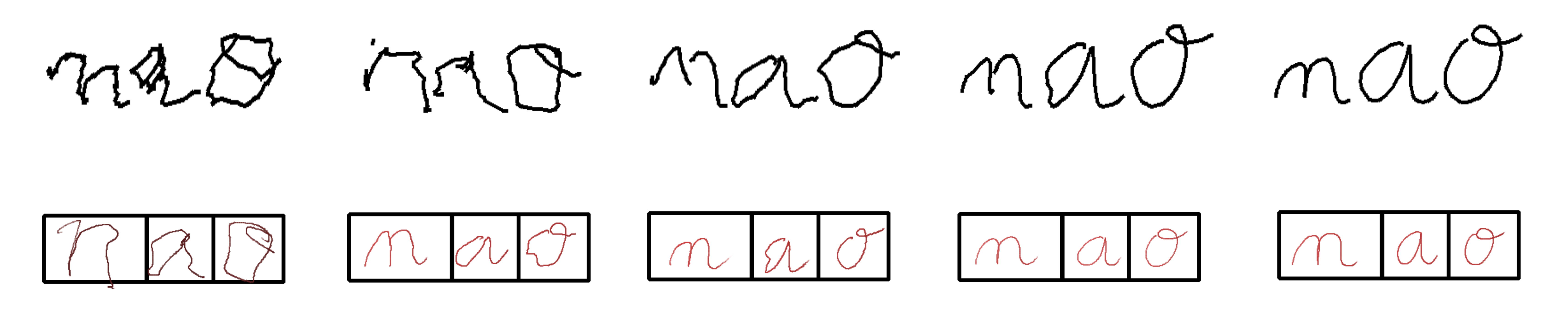}
        \captionsetup{format=hang}
        \caption{Example of five interactions for writing the word "nao". The top row shows the writing of the robot, the bottom row shows the child's writing.}
        \label{fig-cowriter-seq}
    \end{subfigure}
    \caption{The robot writes iteratively adapted to the handwriting profile of the child; if the child's handwriting is shaky, the robot too writes with a shaky handwriting. In correcting the robot's handwriting, the child works towards remediating theirs.}
    \vspace{-2mm}
    %The robot iteratively proposes a handwriting
    \label{fig:nao}
\end{figure*}

%In addition to giving a confidence boost, the robot iteratively proposes a handwriting adapted to the handwriting profile of the child. For example, if the child's handwriting is shaky, the robot too writes with a shaky handwriting. In correcting the robot's handwriting, the child consequently works towards remediating his/her own (mirrored) defaults. In section \ref{sec.experiments.realworld.cowriter}, we show the effectiveness of our teaching algorithm in choosing this sequence of examples.
%There is therefore a teacher hidden in the learner (robot). We propose to train this hidden teacher with our machine teaching algorithm.

\vspace{-1mm}
\subsection{Overview of our Approach}

% \begin{figure*}[h]
%     \captionsetup[subfigure]{font=scriptsize,labelfont=scriptsize}
%     \centering
%     \begin{subfigure}[t]{0.28\textwidth}
%         \includegraphics[width=\textwidth]{./plots/ICT_Picture_1a.pdf}
%         \captionsetup{format=hang}
%         \caption{Individual teaching (IT)}
%         \label{fig-cartoon-it}
%     \end{subfigure}
%     \hspace{7mm}
%     \begin{subfigure}[t]{0.28\textwidth}
%         \includegraphics[width=\textwidth]{./plots/ICT_Picture_1b.pdf}
%         \captionsetup{format=hang}
%         \caption{Classroom teaching (CT)}
%         \label{fig-cartoon-ct}
%     \end{subfigure}
%     \hspace{7mm}
%     \begin{subfigure}[t]{0.28\textwidth}
%         \includegraphics[width=\textwidth]{./plots/ICT_Picture_1c.pdf}
%         \captionsetup{format=hang}
%         \caption{Classroom teaching with partitioning (CTwP)}
%         \label{fig-cartoon-ct-opt-part}
%     \end{subfigure}
%     \caption{Comparisons between individual teaching (IT) and classroom teaching (CT and CTwP) paradigms.}
%     \label{fig-cartoon-representations}
% \end{figure*}

In this paper, we study the problem of designing optimal teaching examples for a classroom of iterative learners. We refer to this new paradigm as \emph{iterative classroom teaching} (CT). We focus on online projected gradient descent learners under squared loss function. The learning dynamics comprise of the learning rates and the initial states which are different for different students. At each time step, the teacher constructs the next training example based on information regarding the students' learning dynamics. We focus on the following teaching objectives motivated by real-world classroom settings, where at the end of the teaching process:
\begin{enumerate}[label=(\roman*)]
\item all learners in the class converge to the target model (\textit{cf.} Eq.~\eqref{teach-obj-all-eq}),
\item the class on average converges to the target model (\textit{cf.} Eq.~\eqref{teach-obj-avg-eq}).
\end{enumerate}

\paragraph{Contributions}
We first consider that setting wherein at all times, the teacher has complete knowledge of the learning rates, loss function, and full observability of the internal states of all the students in the class. A naive approach here would be to apply \cite{liu2017iterative}'s omniscient teaching strategy individually for each student in the class. This would require $\mathcal{O} \br{N \log \frac{1}{\epsilon}}$ teaching examples, where $N$ is the number of students in the classroom. We present a teaching strategy that can achieve a convergence rate of $\mathcal{O} \br{k \log \frac{1}{\epsilon}}$, where $k$ is the rank of the subspace in which the students of the classroom lie (i.e. $k \leq \min\bc{d,N}$, where $d$ is the ambient dimension of the problem). We also prove the robustness of our algorithm in noisy and incomplete information settings.

%In Section 4, we extend our analysis to cater to more practical scenarios (such as teaching human learners) where the teacher has limited observability of the current state of the classroom or has limited knowledge about the learning rates of the students.

%\paragraph{Outline of the paper}
We then explore the idea of partitioning the classroom into smaller groups of homogenous students based on either their learning ability or prior knowledge. We also validate our theoretical results on a simulated classroom of learners and demonstrate their practical applications to the task of teaching how to classify between butterflies and moths. Further, we show the applicability of our teaching strategy to the task of teaching children how to write (\textit{cf}. Figure \ref{fig:nao}).

%% file: 2_model.tex
\section{The Model}\label{sec.model}

%We consider the following model for the iterative classroom teaching problem.

In this section, we consider a stylized model to derive a solid understanding for the dynamics of the learners. This simplicity of our model will then allow us to gain insights into classroom partitioning (i.e., how to create classrooms), and then explore the key trade-offs between the learners' workload as well as the teacher's orchestration costs. By orchestration costs, we mean the number of examples the teacher needs to teach the class.
%We will then demonstrate how our algorithm helps children acquire handwriting skills with robots, where the children's cohort are chosen from theory. 

% \subsection{Preliminary Notation}
\vspace{-1.5mm}
\paragraph{Notation}
Define $\bc{a_i}_{i=1}^N := \bc{a_1, \dots, a_N}$ as a set of $N$ elements and $\bs{N} := \bc{1, \dots , N}$ as the index set. For a given matrix $A$, denote $\lambda_i \br{A}$ and $e_i \br{A}$ to be the $i$-th largest eigenvalue of $A$ and the corresponding eigenvector respectively. $\norm{\cdot}$ denotes the Euclidean norm unless otherwise specified. The projection operation on a set $\mathcal{W}$ for any element $y$ is defined as follows:
%Consider a probability space denoted by $(\Sigma, \mathcal{F}, P)$. Let $H_t\subset\mathcal{F}$ be the history of a random process up until time $t$. For a random variable $X$, define $\mathbb{E}_t\bs{X}:=\mathbb{E}\bs{X|H_t}$.

\vspace{-1.5mm}
\begin{align*}
\texttt{Proj}_{\mathcal{W}}\br{y} ~:=~& \argmin_{x \in \mathcal{W}} \norm{x - y}^2 %\\
% ~=~& \min\bc{1,\frac{\norm{\mathcal{W}}}{\norm{y}}} y .
\end{align*}

% \subsection{Model Parameters}
\paragraph{Parameters}
In \textit{synthesis-based teaching} \cite{liu2017iterative}, $\mathcal{X} = \bc{x \in \mathbb{R}^d, \norm{x} \leq D_{\mathcal{X}}}$ represents the feature space and the label set is given by $\mathcal{Y} = \mathbb{R} \, \text{(for regression) or } \{1, 2, \ldots, m\} \, \text{(for classification)}$. A training example is denoted by $(x, y) \in \mathcal{X} \times \mathcal{Y}$.
Further, we define the feasible hypothesis space by $\mathcal{W} = \bc{w \in \mathbb{R}^d, \norm{w} \leq D_{\mathcal{W}}}$, and denote the target hypothesis by $w^*$.

\paragraph{Classroom}
%We study the case when this is a constant value and also when it dynamically changes with time.
The classroom consists of $N$ students. Each student $j \in \bs{N}$ has two internal parameters: i) the learning rate (at time $t$) represented by $\eta_j^t$, and ii) the initial internal state given by $w^0_j \in \mathcal{W}$. At each time step $t$, the classroom receives a labelled training example $(x^t, y^t) \in \mathcal{X} \times \mathcal{Y}$ and each student $j$ performs a projected gradient descent step as follows:
\[
w^{t+1}_j ~=~ \texttt{Proj}_{\mathcal{W}}\br{w^{t}_j - \eta_j^t \frac{\partial \ell\br{\pi(w^t_j, x^t), y^t}}{\partial w^t_j}},
\]
where $\ell$ is the loss function and $\pi(w^t_j, x^t)$ is the student's label for example $x^t$. We restrict our analysis to the linear regression case where $\pi(w^t_j, x^t) = \ip{w^t_j}{x^t}$ and $\ell\br{\ip{w^t_j}{x^t}, y^t} = \frac{1}{2}\br{\ip{w^t_j}{x^t} - y^t}^2$.

\paragraph{Teacher}
The teacher, over a series of iterations, interacts with the students in the classroom and guides them towards the target hypothesis by choosing ``helpful" training examples. The choice of the training example depends on how much information she has about the students' learning dynamics.
%By choosing examples randomly from the teaching domain, we can obtain the traditional machine learning paradigm.

\begin{itemize}
\item \textit{Observability}: This represents the information that the teacher possesses about the internal state of each student. We study two cases: i) when the teacher knows the exact value $\bc{w_j^t}_{j=1}^{N}$, and ii) when the teacher has a noisy estimate denoted by $\bc{\tilde{w}_j^t}_{j=1}^{N}$ at any time $t$.

\item \textit{Knowledge}: This represents the information that the teacher has regarding the learning rates of each student. We consider two cases: i) when the learning rate of each student is constant and known to the teacher, and ii) when each student draws a value for the learning rate from a normal distribution at every time step, while the teacher only has access to the past values.
\end{itemize}

%
%In Section \ref{sec:ideal-setting}, we study the ``omniscient" teacher that has full observability and full knowledge. We always assume that the teacher knows the target hypothesis $w^*$ to be taught. In Section \ref{subsec:noise-wjt}, we analyze the case when the teacher has full knowledge, but limited observability. And in Section \ref{sec:noisy-eta}, we consider the case when the teacher has full observability, but limited knowledge.

\paragraph{Teaching objective}
In the abstract machine teaching setting, the objective corresponds to approximately training a predictor. Given an accuracy value $\epsilon$ as input, at time $T$ we say that a student $j \in \bs{N}$ has approximately learnt the target concept $w^*$ when $\norm{w^T_j - w^*} \leq \epsilon$. In the strict sense, the teacher's goal may be to ensure that every student in the classroom converges to the target as quickly as possible, i.e.,
\begin{equation}
\label{teach-obj-all-eq}
\norm{w^T_j - w^*} ~\leq~ \epsilon , \, \forall{j \in \bs{N}} .
\end{equation}
The teacher's goal in the average case however, is to ensure that the classroom as a whole converges to $w^*$ in a minimum number of interactions. More formally, the aim is to find the smallest value $T$ such that the following condition holds:
\begin{equation}
\label{teach-obj-avg-eq}
\frac{1}{N} \sum_{i=1}^N{\norm{w_j^T - w^*}^2} ~\leq~ \epsilon .
\end{equation}

%% file: 3_average-learner.tex
\section{Classroom Teaching}
\label{sec:ideal-setting}
We study the omniscient and synthesis-based teacher, equivalent to the one considered in  \cite{liu2017iterative}, but for the iterative classroom teaching problem under the squared loss given by $\ell\br{\ip{w}{x},y} := \frac{1}{2}\br{\ip{w}{x} - y}^2$. Here the teacher has full  knowledge of the target concept $w^*$, learning rates $\bc{\eta_j}_{j=1}^N$ (assumed constant), and internal states $\bc{w^t_j}_{j=1}^N$ of all the students in the classroom.

\vspace{-4mm}
\paragraph{Teaching protocol}
At every time step $t$, the teacher uses all the information she has to choose a training example $x^t \in \mathcal{X}$ and the corresponding label $y^t = \ip{w^*}{x^t} \in \mathcal{Y}$ (for linear regression). The idea is to pick the example which minimizes the average distance between the students' internal states and the target hypothesis at every time step. Formally,
\begin{align*}
x^t =& \argmin_{x \in \mathcal{X}} \frac{1}{N} \sum_{j=1}^{N}{\norm{w^{t}_j - \eta_j \frac{\partial \ell\br{\ip{w^t_j}{x}, \ip{w^*}{x}}}{\partial w^t_j} - w^*}^2} .
\end{align*}
Note that constructing the optimal example at time $t$, is in general a non-convex optimization problem. For the squared loss function, we present a closed-form solution below.
%as described 
%where $G\br{w_j^t;x,y} = \frac{\partial \ell\br{\ip{w^t_j}{x}, y}}{\partial w^t_j}$.

% The students perform linear regression with squared loss error as defined in Section \ref{sec.model}. The student $i$`s prediction for $y^t$ is given by $y_i^t = \ip{w^t_i}{x^t}$.

% In this section we study the iterative classroom teaching problem under the squared loss given by $\ell\br{\ip{w}{x},y} := \frac{1}{2}\br{\ip{w}{x} - y}^2$. We consider the synthesis-based teaching \cite{liu2017iterative}, where the teacher can provide any samples from $\mathcal{X} = \bc{x \in \mathbb{R}^d, \norm{x} \leq R}$ and $\mathcal{Y} = \mathbb{R}$. We further assume that the students and the teacher are linear predictors -- that is at time $t$, the teacher generates the example as $y^t = \ip{w^*}{x^t}$, and the student $i \in \bs{N}$ makes the prediction as $y_i^t = \ip{w^t_i}{x^t}$. Define $G \br{w} := \frac{\partial \ell\br{\ip{w}{x}, y}}{\partial w}$. For the squared loss, we get $G \br{w} = \br{\ip{w}{x} - y} \cdot x$.

% In the ideal setting, the teacher has full knowledge of the students' learning rates $\bc{\eta_i}_{i=1}^N$ and full observability of the students' states $\bc{w^t_i}_{i=1}^N$ at every time step. This is the case of the ``omniscient" teacher studied by \cite{liu2017iterative}, for teaching a single student. The teacher uses the current state of the classroom, learning rates, and teaching target to choose an appropriate training example.

\begin{algorithm}[tb]
   \caption{CT: Classroom teaching algorithm}
   \label{icta-pca-algo}
\begin{algorithmic}
   \STATE {\bfseries Input:} target $w^* \in \mathcal{W}$; students' learning rates $\bc{\eta_j}_{j=1}^N$
   \STATE {\bfseries Goal:} accuracy $\epsilon$
   \STATE Initialize $t = 0$
   \STATE Observe $\bc{w_j^0}_{j=1}^N$
   \WHILE{$\frac{1}{N} \sum_{j=1}^N{\norm{w_j^t - w^*}^2} > \epsilon$}
   		\STATE Observe $\bc{w_j^t}_{j=1}^N$
		\STATE Choose $\gamma_t$ s.t. $\gamma_t \leq D_{\mathcal{X}}$, and $2 - \eta_j \gamma_t^2 \geq 0 , \forall{j \in \bs{N}}$
		\STATE Construct $W^t$ given by Eq.~\eqref{pca-main-mat}
   		\STATE Pick example $x^t = \gamma_t \cdot e_1\br{W^t}$ and $y^t = \ip{w^*}{x^t}$
		\STATE Provide the labeled example $\br{x^t , y^t}$ to the classroom
		\STATE Students' update: $\forall{j \in \bs{N}}$, \[ w_j^{t+1} \leftarrow \texttt{Proj}_{\mathcal{W}} \br{w_j^t - \eta_j \br{\ip{w_j^t}{x^t} - y^t} x^t}. \]
		\STATE $t \leftarrow t+1$
   \ENDWHILE
\end{algorithmic}
\end{algorithm}

\vspace{-4mm}
\paragraph{Example construction}
Define, $\hat{w}_j^t ~:=~ {w_j^t - w^*}$, for $j \in \bs{N}$. Then the teacher constructs the common example for the whole classroom at time $t$ as follows:
\begin{enumerate}
\item the feature vector $x^t = \gamma_t \hat{x}^t$ such that
\begin{enumerate}
\item the magnitude $\gamma_t$: \begin{equation}
\label{gamma-t-conditions}
\gamma_t \leq D_{\mathcal{X}}, \text{ and } 2 - \eta_j \gamma_t^2 ~\geq~ 0 , \forall{j \in \bs{N}} .
\end{equation}
\item the direction $\hat x^t$ (with $\norm{\hat x^t} = 1$):
\begin{equation}
\hat x^t ~:=~ \argmax_{x: \norm{x}=1} x^\top W^t x ~=~ e_1\br{W^t}, \label{first-pc-eq}
\end{equation}
\vspace{-1mm}
where
\vspace{-1mm}
\begin{align}
\alpha_j^t ~:=~& \eta_j \gamma_t^2 \br{2 - \eta_j \gamma_t^2} \label{alpha-j-eq} \\
W^t ~:=~& \frac{1}{N} \sum_{j=1}^N \alpha_j^t \hat w_j^t \br{\hat w_j^t}^\top . \label{pca-main-mat}
\end{align}
%This specific choice of example is justified along the lines of the proof of Theorem~\ref{main-exp-theorem}.
\end{enumerate}
 \item the label $y^t = \ip{w^*}{x^t}$.
 \end{enumerate}

%At time $t$, $\forall j \in \bs{N}$, we denote the following parameters:
%\begin{itemize}
%
%\item Current distance of student $j$ from target.
%\begin{equation}\hat{w}_j^t ~:=~ {w_j^t - w^*}\end{equation}
%\item Magnitude parameter
%\begin{equation} \gamma_t \leq D_{\mathcal{X}} \text{ such that } 2 - \eta_j \gamma_t^2 ~\geq~ 0 \end{equation} \label{gamma-t-conditions}
%\item Coefficient parameter
%\begin{equation} \alpha_j^t ~:=~ \eta_j \gamma_t^2 \br{2 - \eta_j \gamma_t^2} \end{equation} \label{alpha-j-eq}
%\item Distance matrix
%\begin{equation} W^t ~:=~ \frac{1}{N} \sum_{j=1}^N \alpha_j^t \hat w_j^t \br{\hat w_j^t}^\top \end{equation} \label{pca-main-mat}
%
%\end{itemize}
%
%\textcolor{red}{TODO: improve names with discussions of their meaning/implications}
%
%Then the teacher constructs the common example for the whole classroom at time $t$ as follows
%\begin{enumerate}
%\item feature vector $x^t = \gamma_t \hat{x}^t$ where
%  \begin{align}
%  \hat x^t ~:=~&  \argmax_{x: \norm{x}=1} x^\top W^t x \nonumber\\
%  ~=~& e_1\br{W^t} , \label{first-pc-eq}
%  \end{align}
%  $\gamma_t$ and $\hat{x}^t$ (with $\norm{\hat{x}^t} = 1$) represent the magnitude and direction of the feature vector respectively.
%\item true label $y^t = \ip{w^*}{x^t}$
%\end{enumerate}

Algorithm~\ref{icta-pca-algo} puts together our omniscient classroom teaching strategy. Theorem \ref{main-exp-theorem} provides the number of examples required to teach the target concept.\footnote{Proofs are given in the Appendix.}

\begin{theorem}
\label{main-exp-theorem}
Consider the teaching strategy given in Algorithm~\ref{icta-pca-algo}. Let $k := \max_t\bc{\mathrm{rank}\br{W^t}}$, where $W^t$ is given by Eq.~\eqref{pca-main-mat}. Define $\alpha_j := \min_{t} \alpha_j^t$, $\alpha_{\mathrm{min}} := \min_{t,j} \alpha_j^t$, and $\alpha_{\mathrm{max}} := \max_{t,j} \alpha_j^t$, where $\alpha_j^t$ is given by Eq.~\eqref{alpha-j-eq}. Then after $t = \mathcal{O}\br{\br{\log \frac{1}{1 - \frac{\alpha_{\mathrm{min}}}{k}}}^{-1} \log \frac{1}{\epsilon}}$ rounds, we have $\frac{1}{N} \sum_{j=1}^N{\norm{w_j^t - w^*}^2} \leq \epsilon$. Furthermore, after $t = \mathcal{O}\br{\max \bc{\br{\log \frac{1}{1 - \alpha_j}}^{-1} \log{\frac{1}{\epsilon}}  ,  \br{\log \frac{1}{1 - \frac{\alpha_{\mathrm{min}}}{k}}}^{-1} \log{\frac{1}{\epsilon}}}}$
rounds, we have $\norm{w_j^t - w^*}^2 \leq \epsilon, \forall{j \in \bs{N}}$.
\end{theorem}

\begin{remark}
\label{main-exp-remark}
By using the fact that $\exp\br{-2x} \leq \log\br{1-x}$, for $2 x \leq 1.59$, we can show that for $k \geq \frac{2 \alpha_{\mathrm{min}}}{1.59}$, $\mathcal{O}\br{\br{\log \frac{1}{1 - \frac{\alpha_{\mathrm{min}}}{k}}}^{-1} \log \frac{1}{\epsilon}} \approx \mathcal{O}\br{ \frac{k}{\alpha_{\mathrm{min}}} \log \frac{1}{\epsilon}}$.
\end{remark}

Based on Theorem~\ref{main-exp-theorem} and Remark~\ref{main-exp-remark}, note that the teaching strategy given in Algorithm~\ref{icta-pca-algo} converges in $\mathcal{O} \br{k \log \frac{1}{\epsilon}}$ samples, where $k = \max_t\bc{\mathrm{rank}\br{W^t}} \leq \min\bc{d,N}$. This is in fact a significant improvement (especially when $N \gg d$) over the sample complexity $\mathcal{O} \br{N \log \frac{1}{\epsilon}}$ of a teaching strategy which constructs personalized training examples for each student in the classroom. \\ 

%In the iterative teaching of a single student setting \cite{liu2017iterative} under squared loss, at the time step $t$, the teacher provides the example $x^t = \gamma_t \cdot \frac{w^t-w^*}{\norm{w^t-w^*}}$. By generalizing this result, we considered training examples of the form $x^t = \gamma_t \cdot \frac{\frac{1}{N} \sum_{j=1}^{N}{\br{w_j^t-w^*}}}{\norm{\frac{1}{N} \sum_{j=1}^{N}{\br{w_j^t-w^*}}}}$. But in the experiments with this choice of examples, all the students converge on a plane which is close to the target $w^*$.

\vspace{-5mm}
\paragraph{Choice of magnitude $\gamma_t$}
We consider the following two choices of $\gamma_t$:
\begin{enumerate}
\item static $\gamma_t = \min\bc{\frac{1}{\max_{j \in \bs{N}} \sqrt{\eta_j}}, D_{\mathcal{X}}}$: This ensures that the classroom can converge without being partitioned into small groups of students. However, the value $\alpha_{\mathrm{min}}$ becomes small and as a result, the sample complexity increases.
\item dynamic $\gamma_t = \min\bc{\sqrt{\frac{\sum_{j=1}^N{\eta_j \norm{w_j^t - w^*}^2}}{\sum_{j=1}^N{\eta_j^2 \norm{w_j^t - w^*}^2}}}, D_{\mathcal{X}}}$: This provides an optimal constant for the sample complexity, but requires that for effective teaching the classroom is partitioned appropriately. This value of $\gamma_t$ is obtained by maximizing the term $\sum_{j=1}^N {\eta_j \gamma_t^2 \br{2 - \eta_j \gamma_t^2} \norm{w_j^t - w^*}^2}$.
\end{enumerate}

\paragraph{Natural partitioning based on learning rates}
In order to satisfy the requirements given in Eq.~\eqref{gamma-t-conditions}, for every student $j \in \bs{N}$, we require (for dynamic $\gamma_t$):
\begin{align}
\eta_j \gamma_t^2 ~\leq~& \eta_{\text{max}} \frac{\sum_{j=1}^N{\eta_j \norm{w_j^t - w^*}^2}}{\sum_{j=1}^N{\eta_j^2 \norm{w_j^t - w^*}^2}} \nonumber \\
~\leq~& \eta_{\text{max}} \frac{\sum_{j=1}^N{\eta_j \norm{w_j^t - w^*}^2}}{\eta_{\text{min}} \sum_{j=1}^N{\eta_j \norm{w_j^t - w^*}^2}} ~\leq~ 2,
\end{align}
where $\eta_{\text{max}} = \max_j \eta_j$, and $\eta_{\text{min}} = \min_j \eta_j$. That is, if $\eta_{\text{max}} \leq 2 \eta_{\text{min}}$, we can safely use the above optimal $\gamma_t$. This observation also suggests a natural partitioning of the classroom: $\bc{[\eta_{\text{min}}, 2 \eta_{\text{min}}), [2 \eta_{\text{min}}, 4 \eta_{\text{min}}), \dots , [2^m \eta_{\text{min}}, 2 \eta_{\text{max}})}$, where $m = \floor{\log_2 \frac{\eta_{\text{max}}}{\eta_{\text{min}}}}$. 

%% file: 4_worst-learner.tex
\section{Robust Classroom Teaching}
\label{sec:robust-teaching}

In this section, we study the robustness of our teaching strategy in cases when the teacher can access the current state of the classroom only through a noisy oracle, or when the learning rates of the students vary with time.
%stochastically

%conducts a quiz, i.e. gives $m$ examples $Q^t = \{q^t_1, q^t_2, \ldots q^t_m\}$ where $\forall j \in [m], q^t_j \in \mathcal{X}$ and receives feedback from the classroom. With this information, the teacher constructs a noisy estimate $\bc{\tilde{w}^t_i}_{i=1}^N$ of $\bc{w^t_i}_{i=1}^N$ (or $\tilde{W}^t$ of $W^t$) and uses it to choose an appropriate training example.

%We present this idea below in Algorithm~\ref{icta-pca-lo-algo}.
%
%\begin{algorithm}[tb]
%   \caption{ICTLO: Iterative Classroom Teaching with Limited Observability}
%   \label{icta-pca-lo-algo}
%\begin{algorithmic}
%   \STATE {\bfseries Input:} target $w^* \in \mathbb{R}^d$; students' learning rates $\bc{\eta_i}_{i=1}^N$
%   \STATE {\bfseries Goal:} accuracy $\epsilon$
%   \STATE Initialize $t = 0$.
%   \STATE Obtain $\tilde{W}^0$ by tracking via quiz.
%   \WHILE{$\frac{1}{N} \sum_{i=1}^N{\norm{w_i^t - w^*}^2} > \epsilon$}
%   		\STATE Obtain $\tilde{W}^t$ by tracking via quiz.
%		\STATE Choose $\gamma_t$ s.t. $\gamma_t \leq D_{\mathcal{X}}$, and $2 - \eta_j \gamma_t^2 \geq 0 , \forall{j \in \bs{N}}$.
%   		\STATE Pick example $x^t = \gamma_t \cdot e_1\br{\tilde{W}^t}$ and $y^t = \ip{w^*}{x^t}$.
%		\STATE Provide the labeled example $\br{x^t , y^t}$ to the classroom.
%		\STATE Students' update: $\forall{j \in \bs{N}}$, \[ w_j^{t+1} \leftarrow \texttt{Proj}_{\mathcal{W}} \br{w_j^t - \eta_j \br{\ip{w_j^t}{x^t} - y^t} x^t}. \]
%		\STATE $t \leftarrow t+1$
%   \ENDWHILE
%\end{algorithmic}
%\end{algorithm}

\subsection{Noise in $w_j^t$'s}
\label{subsec:noise-wjt}
Here, we consider the setting where the teacher cannot directly observe the students' internal states $\bc{w^t_j}_{j=1}^N, \forall t$ but has full knowledge of students' learning rates $\bc{\eta_j}_{j=1}^N$. Define $\alpha_{\mathrm{min}} := \min_{t,j} \alpha_j^t$, and $\alpha_{\mathrm{avg}} := \max_t{\frac{1}{N} \sum_{j=1}^N{\alpha_j^t}}$, where $\alpha_j^t$ is given by Eq.~\eqref{alpha-j-eq}. At every time step $t$, the teacher only observes a noisy estimate of $w_j^t$ (for each $j \in \bs{N}$) given by
\begin{equation}
\label{noisy-wjt-model-eq}
\tilde{w}_j^t ~:=~ w_j^t + \delta_j^t ,
\end{equation}
where $\delta_j^t$ is a random noise vector such that $\norm{\delta_j^t} \leq \frac{\epsilon}{4 \br{\frac{\alpha_\mathrm{avg}}{\alpha_\mathrm{min}} d + 1} D_{\mathcal{W}}}$. Then the teacher constructs the example as follows:
\begin{align}
\hat  x^t ~:=~& \argmax_{x: \norm{x}=1} x^\top \bc{\frac{1}{N} \sum_{j=1}^N{\alpha_j^t {\hat w_j^t}\br{\hat w_j^t}^\top}} x \nonumber \\
x^t ~:=~& \gamma_t \hat  x^t \text{ and } y^t = \ip{w^*}{x^t} , \label{noise-wt-example}
\end{align}
where $\hat w^{t}_j := {\tilde{w}^{t}_j - w^*}$, and $\gamma_t$ satisfies the condition given in Eq.~\eqref{gamma-t-conditions}. The following theorem shows that even under this noisy observation setting Eq.~\eqref{noisy-wjt-model-eq}, with the example construction strategy described in Eq.~\eqref{noise-wt-example}, the teacher can teach the classroom with linear convergence.
\begin{theorem}
\label{main-exp-theorem-noise-wjt}
Consider the noisy observation setting given by Eq.~\eqref{noisy-wjt-model-eq}.
Let $k := \max_t\bc{\mathrm{rank}\br{W^t}}$ where $W^t = \frac{1}{N} \sum_{j=1}^N{\alpha_j^t \br{\tilde{w}^{t}_j - w^*}\br{\tilde{w}^{t}_j - w^*}^\top}$. Then for the robust teaching strategy given by Eq.~\eqref{noise-wt-example}, after $t = \mathcal{O}\br{\br{\log \frac{1}{1 - \frac{\alpha_{\mathrm{min}}}{k}}}^{-1} \log \frac{1}{\epsilon}}$ rounds, we have $\frac{1}{N} \sum_{j=1}^N{\norm{w_j^t - w^*}^2} \leq \epsilon$.
\end{theorem}

%Assume that the teacher obtains/observes the current state of any student through a noisy oracle \emph{i.e.} for any student $j \in \bs{N}$, the teacher has access to $\tilde{w}_j^t := w_j^t + \delta$ with $\norm{\delta} \leq \frac{\epsilon}{4 \br{\frac{d \alpha_\mathrm{avg}}{\alpha_\mathrm{min}} + 1} D_{\mathcal{W}}}$, where $\alpha_{\mathrm{avg}} := \max_t{\frac{1}{N} \sum_{j=1}^N{\alpha_j^t}}$.

%\subsection{Stochastic $\eta_j$}
%\label{subsec:noisy-eta}

%\section{Classroom of Stochastic Learners}

%%%%%%%%%%%%%%%%%%%%%%%%%%%%%%%%%%%%%%%%%%%%%%%%%%%%%%%%% copied from next section
 \begin{figure*}[t!]
     \captionsetup[subfigure]{font=scriptsize,labelfont=scriptsize}
     \centering
     \begin{subfigure}[t]{0.26\textwidth}
         \includegraphics[width=\textwidth]{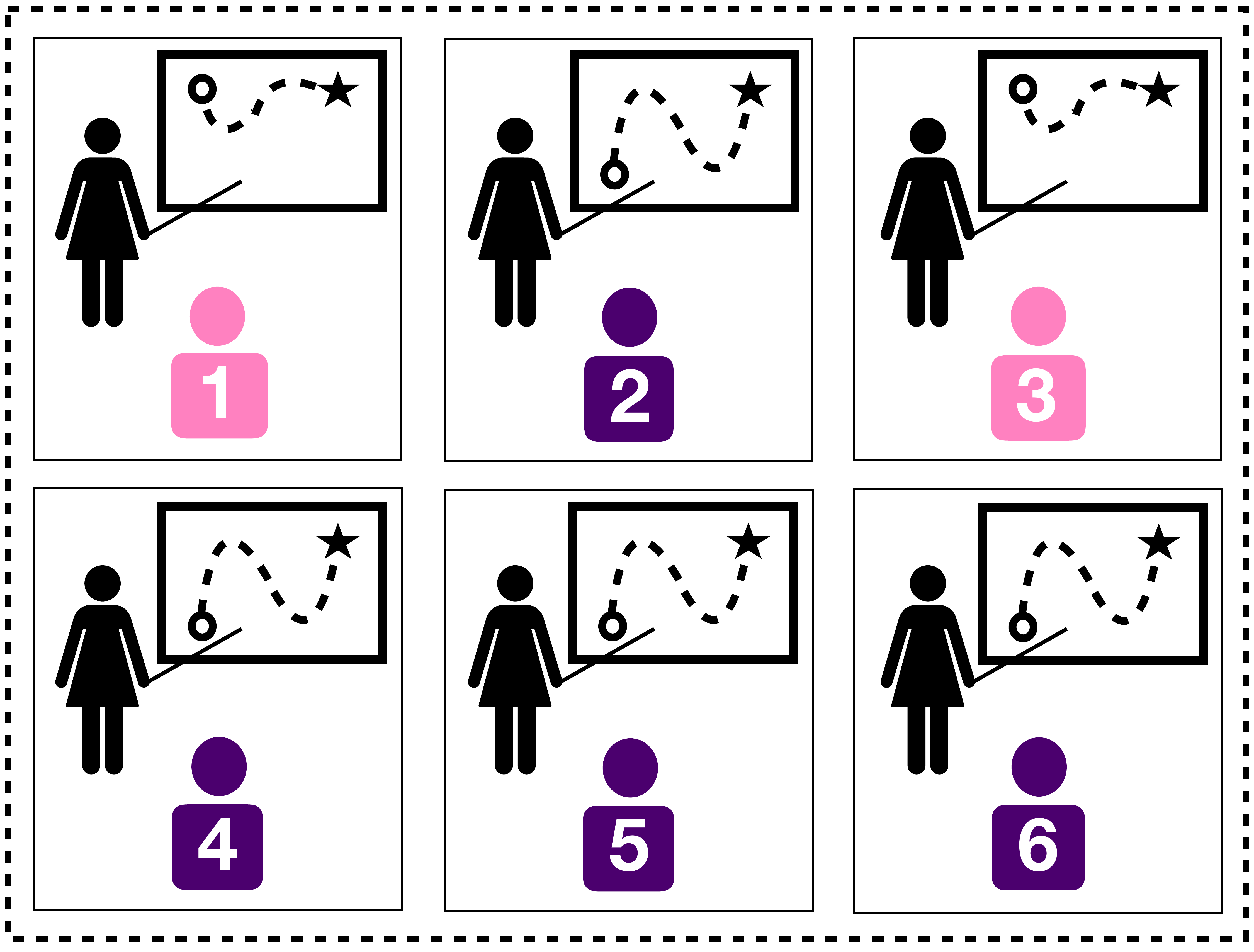}
         \captionsetup{format=hang}
         \caption{Individual teaching (IT)}
         \label{fig-cartoon-it}
     \end{subfigure}
     \hspace{7mm}
     \begin{subfigure}[t]{0.26\textwidth}
         \includegraphics[width=\textwidth]{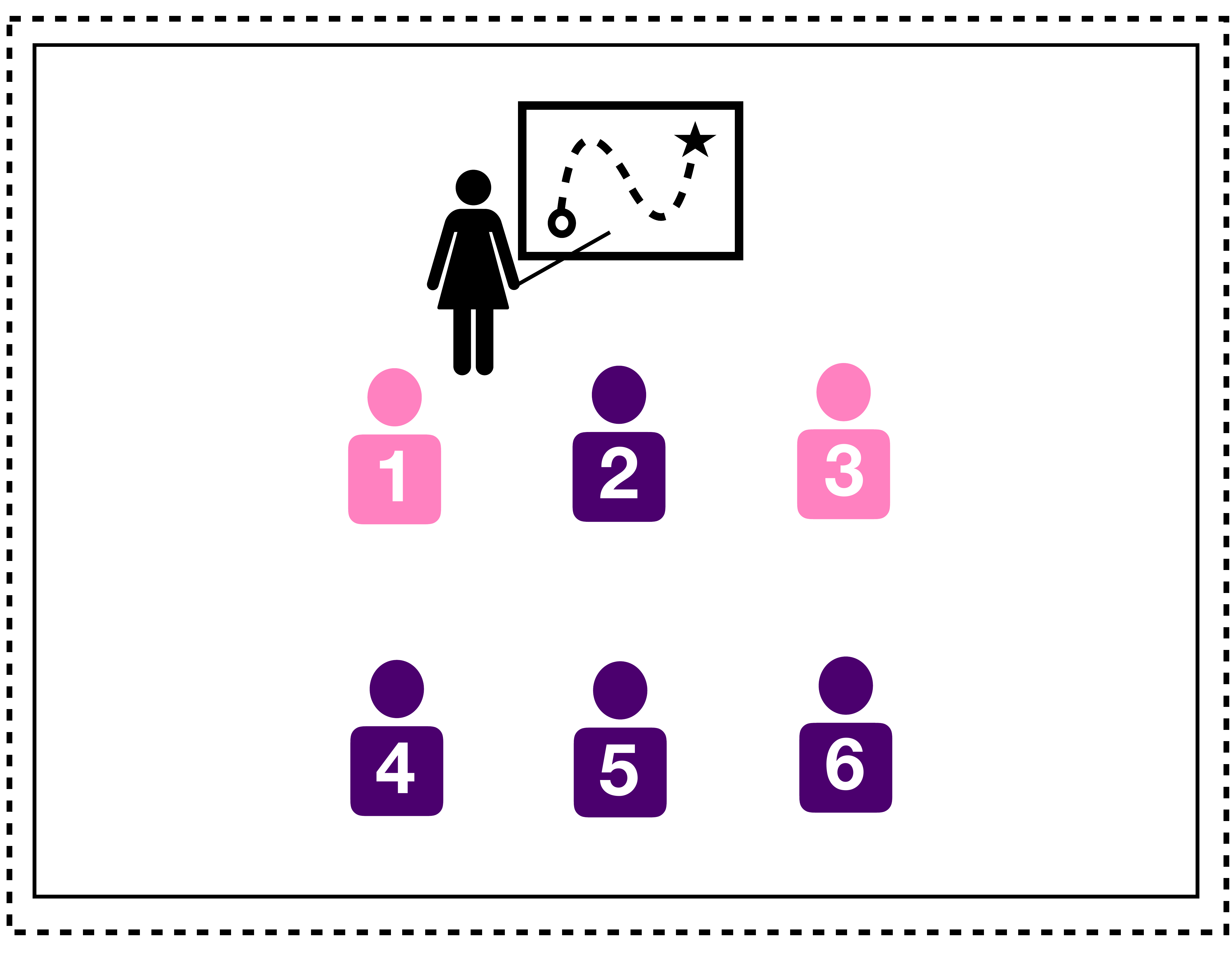}
         \captionsetup{format=hang}
         \caption{Classroom teaching (CT)}
         \label{fig-cartoon-ct}
     \end{subfigure}
     \hspace{7mm}
     \begin{subfigure}[t]{0.26\textwidth}
         \includegraphics[width=\textwidth]{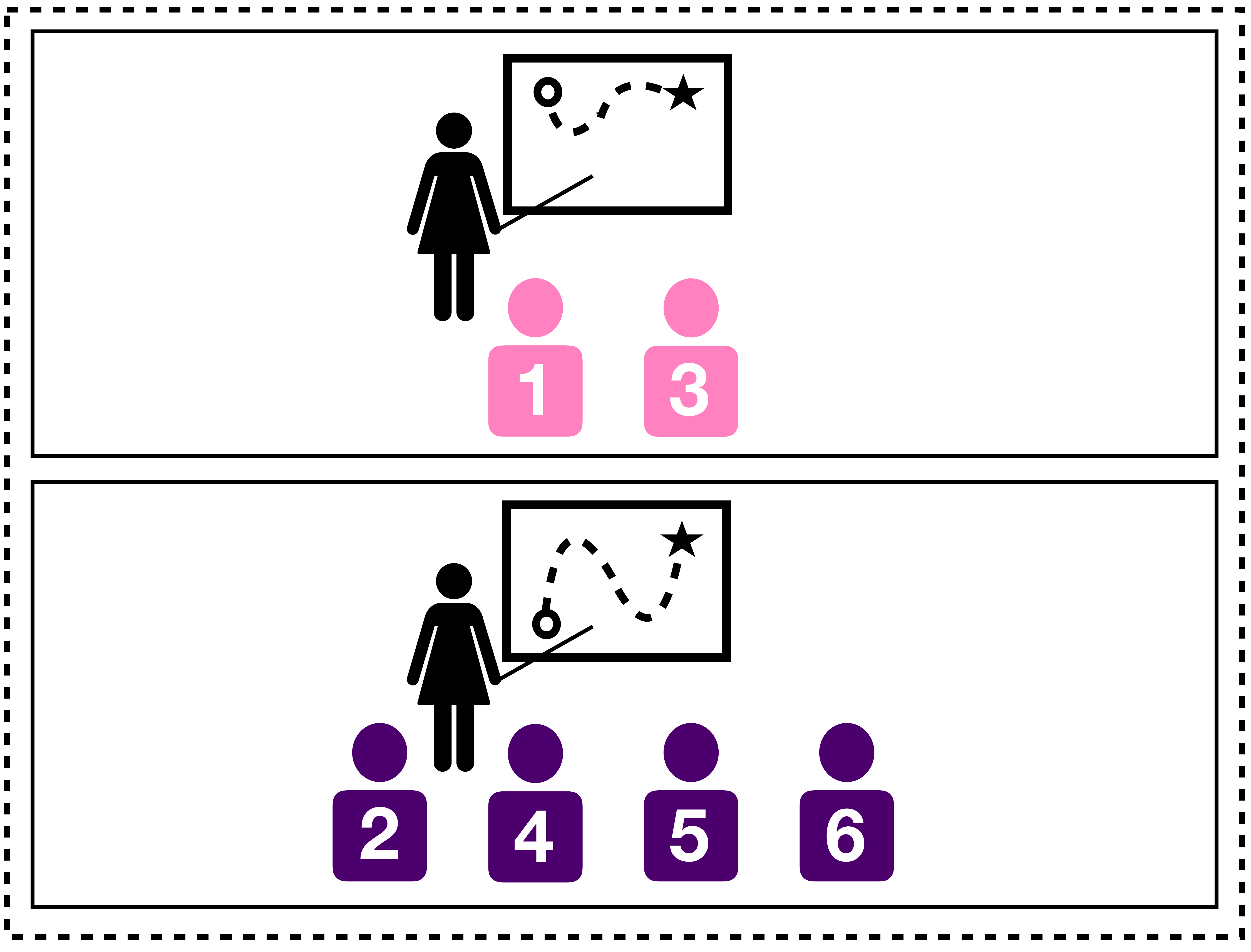}
         \captionsetup{format=hang}
         \caption{Classroom teaching with partitions (CTwP)}
         \label{fig-cartoon-ct-opt-part}
     \end{subfigure}
     \caption{Comparisons between individual teaching (IT) and classroom teaching (CT and CTwP) paradigms.}
         \vspace{-3mm}
     \label{fig-cartoon-representations}
 \end{figure*}
%%%%%%%%%%%%%%%%%%%%%%%%%%%%%%%%%%%%%%%%%%%%%%%%%%%%%%%%%

\subsection{Noise in $\eta_j^t$}
\label{sec:noisy-eta}

Here, we consider a classroom of online projected gradient descent learners with learning rates $\bc{\eta_j^t}_{j=1}^N$, where $\eta_j^t \sim \mathcal{N} \br{\eta_j,\sigma}$. We assume that the teacher knows $\sigma$ (which is constant across all the students) and $\bc{\eta_j}_{j=1}^N$, but doesn't know $\bc{\eta^t_j}_{j=1}^N$.  
% For a given example  $(x, y) \in \mathcal{X} \times \mathcal{Y}$ at time $t$, the update rule of the student $j \in \bs{N}$ is given by
% \begin{equation}
% \label{noisy-eta-classroom-model}
% w^{t+1}_j ~=~ \texttt{Proj}_{\mathcal{W}}\br{w^{t}_j - \eta_j^t \frac{\partial \ell\br{\ip{w^t_j}{x}, y}}{\partial w^t_j}} .
% \end{equation}
Further, we assume that the teacher has full observability of $\bc{w_j^t}_{j=1}^N$. At time $t$, the teacher has access to the history $H_t := \br{\bc{w_j^s}_{s=1}^t , \bc{\eta_j^s}_{s=1}^{t-1} : \forall{j \in \bs{N}}}$. Then the teacher constructs the example as follows (depending only on $H_t$):
\begin{align}
\hat x^t ~:=~& \argmax_{x: \norm{x} = 1} x^\top \bar W^t x ~=~ e_1 \br{\bar W^t} \nonumber \\
x^t ~:=~& \gamma_t \hat  x^t \text{ and } y^t = \ip{w^*}{x^t} , \label{noise-eta-example}
\end{align}
where
\begin{align}
\gamma_t^2 ~\leq~& \frac{2 \eta_j}{\sigma^2 + \eta_j^2}, \forall{j \in \bs{N}} \label{noise-eta-gammat} \\
\bar \eta_j^t ~:=~& \frac{1}{t-1}\sum_{s=1}^{t-1}{\eta_j^s} \label{noise-eta-mujt} \\
\bar \alpha_j^t ~:=~& 2\gamma_t^2 \bar \eta_j^t - \gamma_t^4 \br{\frac{t-2}{t-1}\sigma^2+\br{\bar \eta_j^t}^2} \label{noise-eta-alphajt} \\
\hat w_j^t ~:=~& {w^{t}_j - w^*} \text{ and } \label{noise-eta-wjt} \\
\bar W^t ~:=~& \frac{1}{N} \sum_{j=1}^N \bar \alpha_j^t \hat w_j^t \br{\hat w_j^t}^\top . \label{bar_Wt}
\end{align}
The following theorem shows that, in this setting, the teacher can teach the classroom in expectation with linear convergence.
\begin{theorem}
\label{main-exp-theorem-noise-eta}
Let $k := \max_t\bc{\mathrm{rank}\br{\bar W^t}}$ where $\bar W^t$ is given by Eq.~\eqref{bar_Wt}. Define $\bar \alpha_{\mathrm{min}} := \min_{t,j} \bar \alpha_j^t$, and $\beta_{\mathrm{min}} := \min_{j,t}{\frac{\alpha_j^t}{\bar \alpha_j^t}}$, where $\alpha_j^t := 2\gamma_t^2\eta_j - \gamma_t^4 \br{\sigma^2+\eta_j^2}$ and $\bar \alpha_j^t$ given by Eq.~\eqref{noise-eta-alphajt}. Then for the teaching strategy given by Eq.~\eqref{noise-eta-example}, after $t = \mathcal{O}\br{\br{\log \frac{1}{1 - \frac{\beta_{\mathrm{min}} \bar \alpha_{\mathrm{min}}}{k}}}^{-1} \log \frac{1}{\epsilon}}$ rounds, we have $\mathbb{E} \bs{\frac{1}{N} \sum_{j=1}^N{\norm{w_j^t - w^*}^2}} \leq \epsilon$.
\end{theorem}

%% file: 5_partition.tex
\section{Classroom Partitioning}\label{sec:partitioning}
% Individual teaching is very expensive and sometimes not even feasible due to the effort required in producing personalized education resources. The primary motivation behind classroom teaching is to reduce this orchestration cost. At the same time, classroom teaching has its drawbacks. It increases the students' workload substantially compared to individual teaching because it requires catering to the needs of academically diverse learners. Moreover, providing redundant examples can lower a student's engagement with the teaching process.

% The main motivation of classroom teaching is to reduce the orchestration cost (which is the effort for producing specific education resources) as individual/personalized teaching is very expensive and sometimes not even feasible. However, classroom teaching increases the students' workload substantially (due to the diversity of the learners) compared to individual teaching. Also, providing redundant examples would lower a student's engagement with the teaching process. %process.
%For example, in the crowdsourcing application, the workers will stop learning if we keep providing them redundant examples.
Individual teaching can be very expensive due to the effort required in producing personalized education resources. At the same time, classroom teaching increases the students' workload substantially because it requires catering to the needs of academically diverse learners. We overcome these pitfalls by partitioning the given classroom of $N$ students into $K$ groups such that the orchestration cost of the teacher and the workload of students is balanced. Figure~\ref{fig-cartoon-representations} illustrates these three different teaching paradigms.

Let $T(K)$ be the total number of examples required by the teacher to teach all the groups. Let $S(K)$ be the average number of examples needed by a student to converge to the target. We study the total cost defined as:
% \vspace{-2.5mm}
\begin{align*}
\mathrm{cost}(K) ~:=~ T(K) + \lambda \cdot S(K) ,
\end{align*}
%\[
%\mathrm{cost}(K) ~:=~ T(K) + \lambda \cdot S(K) ,
%\]
%\vspace{-1mm}
where $\lambda$ quantifies the trade-off factor, and its value is application dependent. In particular, for any given $\lambda$, we are interested in that value $K$ that minimizes $\mathrm{cost}(K)$. For example, when $\lambda=\infty$, the focus is on the student workload; thus the optimal teaching strategy is individual teaching, \emph{i.e.}, $K=N$. Likewise, when $\lambda=0$, the focus is on the orchestration cost; thus the optimal teaching strategy is classroom teaching without partitioning, i.e., $K=1$. In this paper, we explore two homogeneous partitioning strategies: (a) based on learning rates of the students $\bc{\eta_j}_{j=1}^N$, (b) based on prior knowledge of the students $\bc{w_j^0}_{j=1}^N$.

%% file: 6_experiments.tex
\section{Experiments}\label{sec:all-experiments}
\input{6.1_experiments-simulation}
\input{6.2_experiments-realworld}
\input{6.3_experiments-realworld-cowriter}

%% file: 6.1_experiments-simulation.tex
%!TEX root = main.tex
%%%%%%%%%%%%%%%%%%%%%%%%%%%%%%%%%%%%%%%%%%%%%%%%%%%%%%%%%
%%%%%%%%%%%%%%%%%%%%%%%%%%%%%%%%%%%%%%%%%%%%%%%%%%%%%%%%%
\begin{figure*}[h]
    \captionsetup[subfigure]{font=scriptsize,labelfont=scriptsize}
    \centering
    \begin{subfigure}[t]{0.29\textwidth}
        \includegraphics[width=\textwidth]{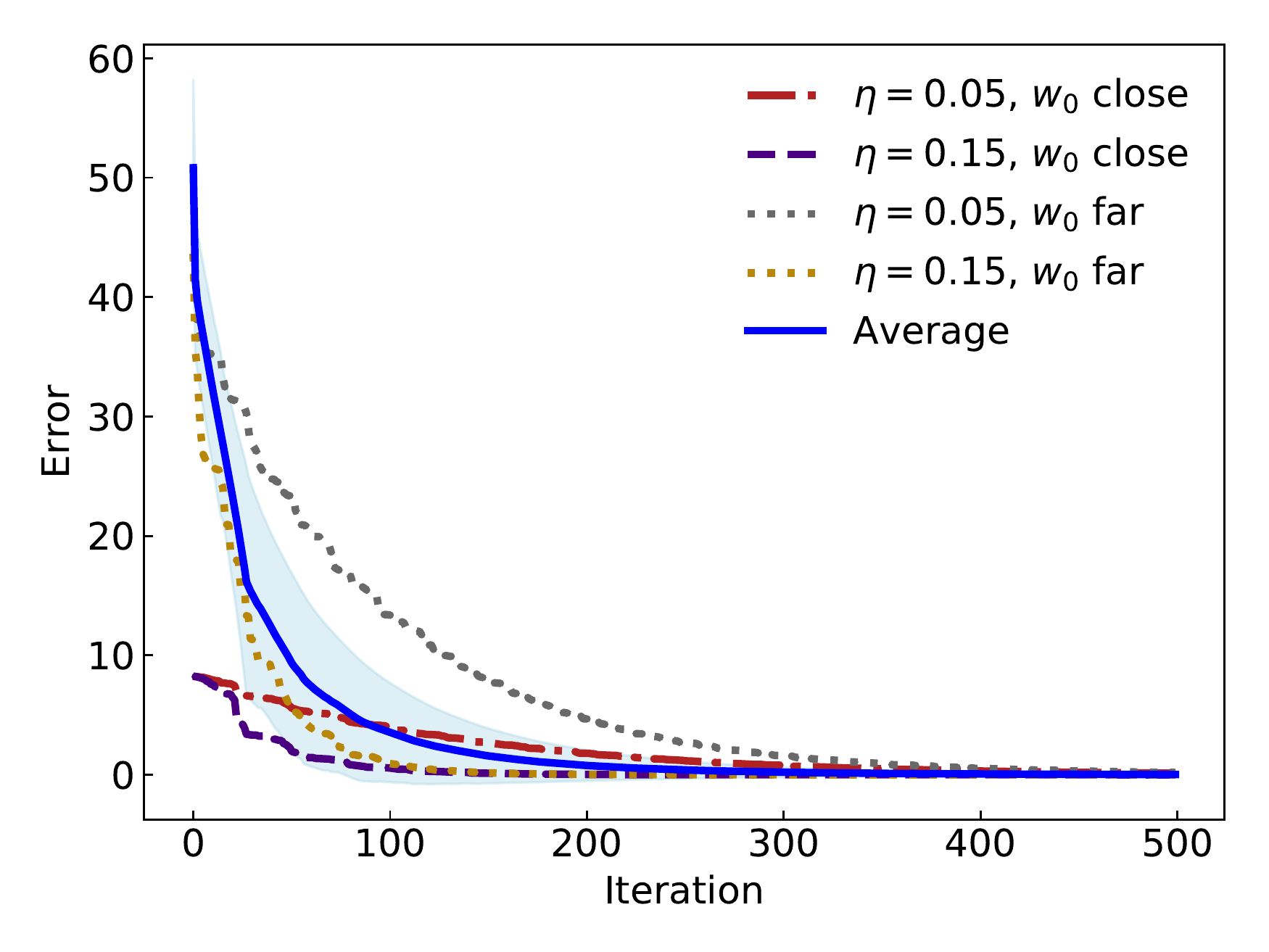}
        \caption{Error plot for CT; average error of the classroom and the error of four selected learners.}
        \label{fig-error-noise-free}
    \end{subfigure}
	\hspace{1mm}
    \begin{subfigure}[t]{0.29\textwidth}
        \includegraphics[width=\textwidth]{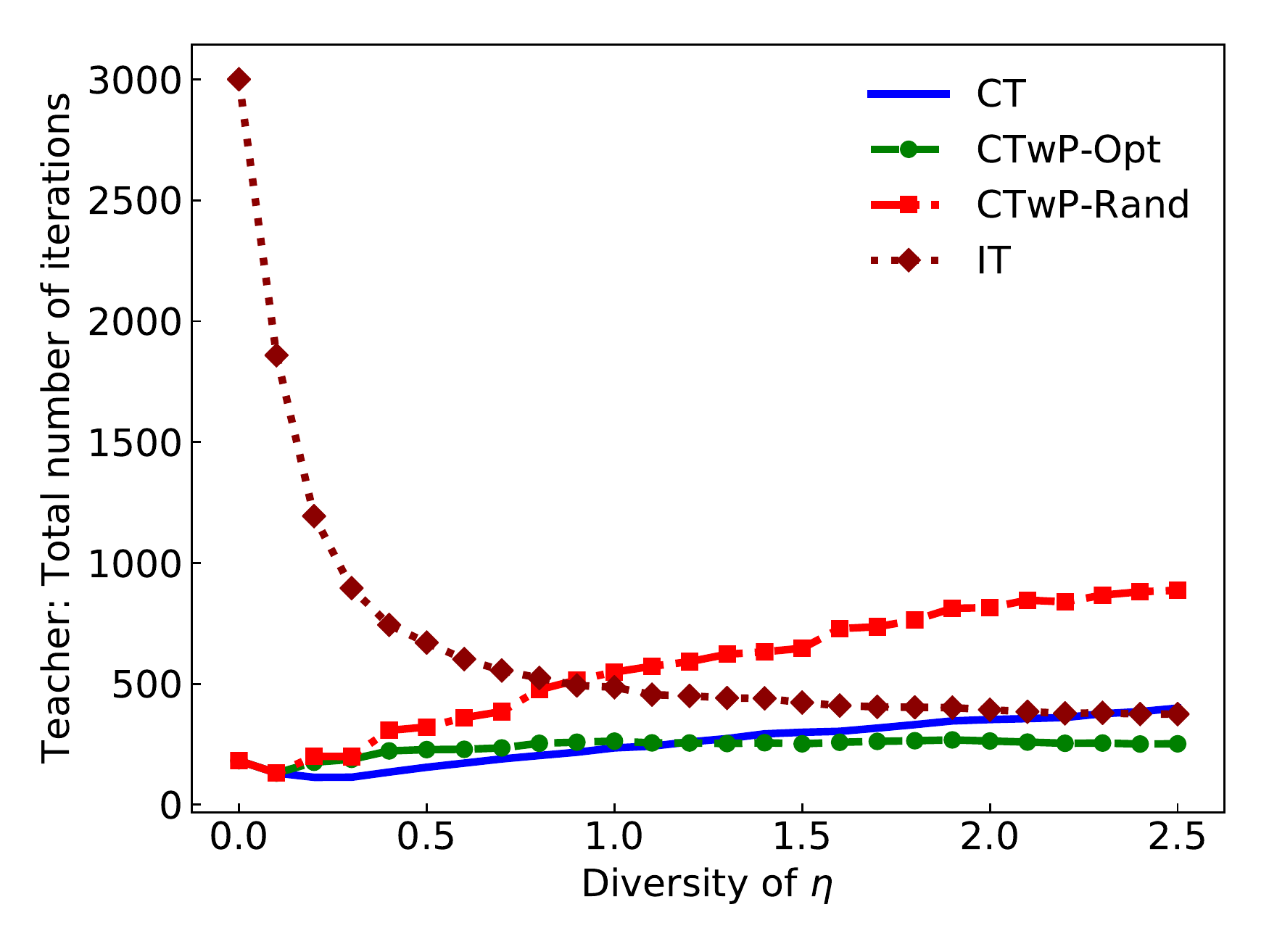}
        \captionsetup{format=hang}
        \caption{Total iterations needed for convergence from the teacher's perspective.}
        \label{fig-cvg-noise-free-eta}
    \end{subfigure}
	\hspace{1mm}
    \begin{subfigure}[t]{0.29\textwidth}
        \includegraphics[width=\textwidth]{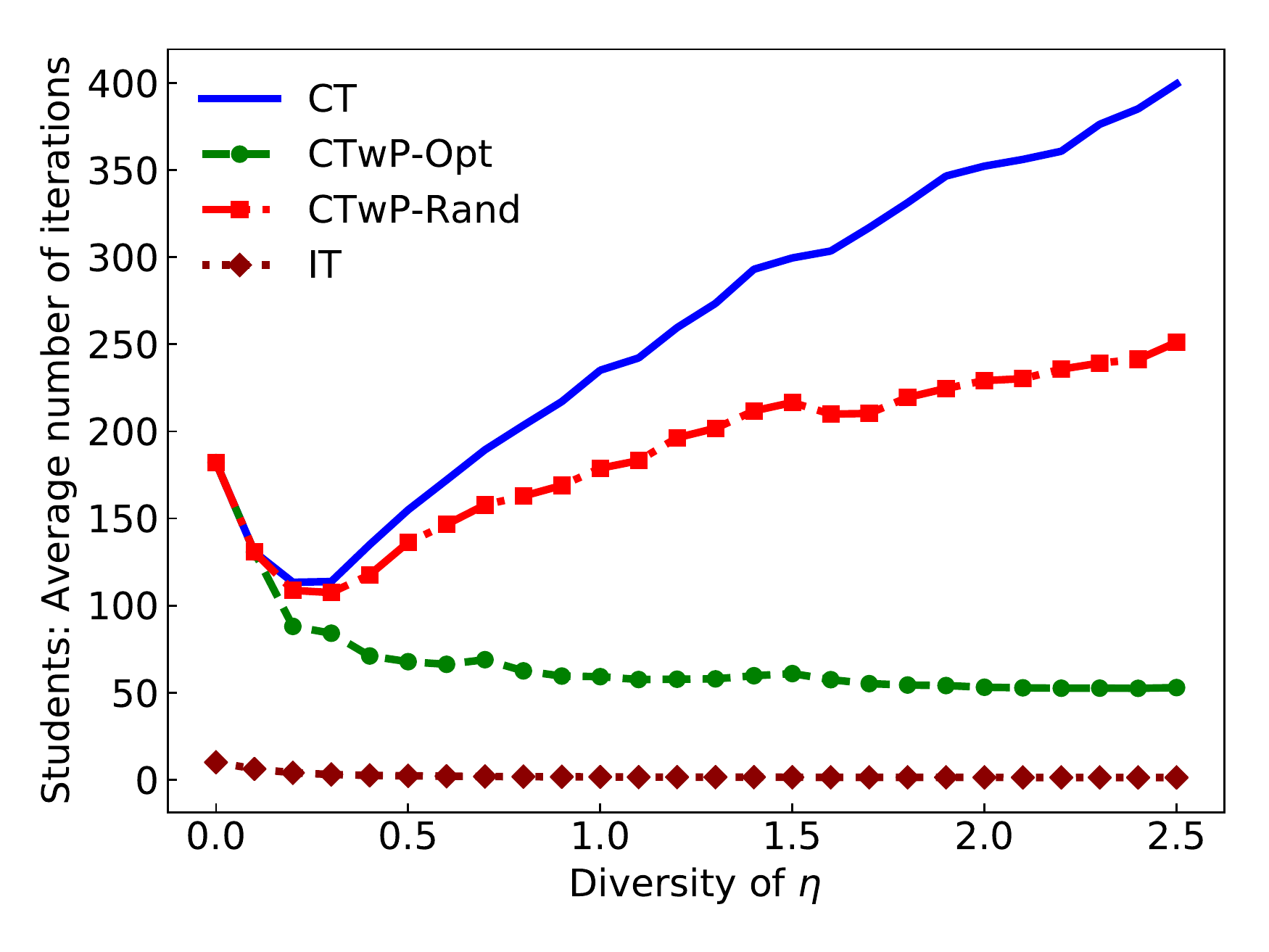}
        \captionsetup{format=hang}
        \caption{Total iterations per student needed for convergence from the students' perspective.}
        \label{fig-cvg-noise-free-eta-student}
    \end{subfigure}
	%%%%%%%%%%%%%%%%%%%
    \begin{subfigure}[t]{0.29\textwidth}
        \includegraphics[width=\textwidth]{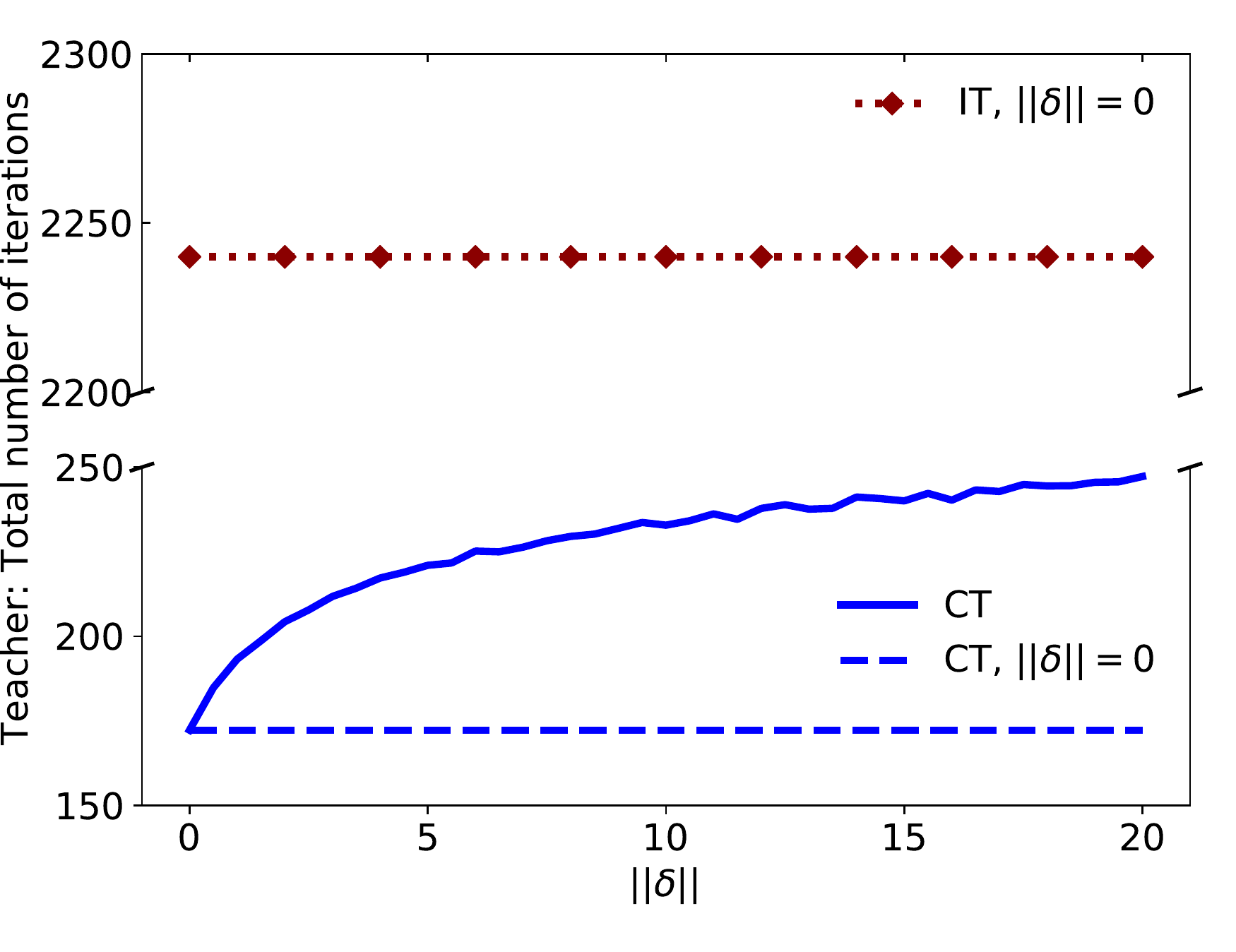}
        \captionsetup{format=hang}
        \caption{Total iterations needed for convergence in noisy $w_t$ case as the noise, $\delta$, increases.}
        \label{fig-cvg-noisy-delta}
    \end{subfigure}
    \hspace{1mm}    
    \begin{subfigure}[t]{0.29\textwidth}
        \includegraphics[width=\textwidth]{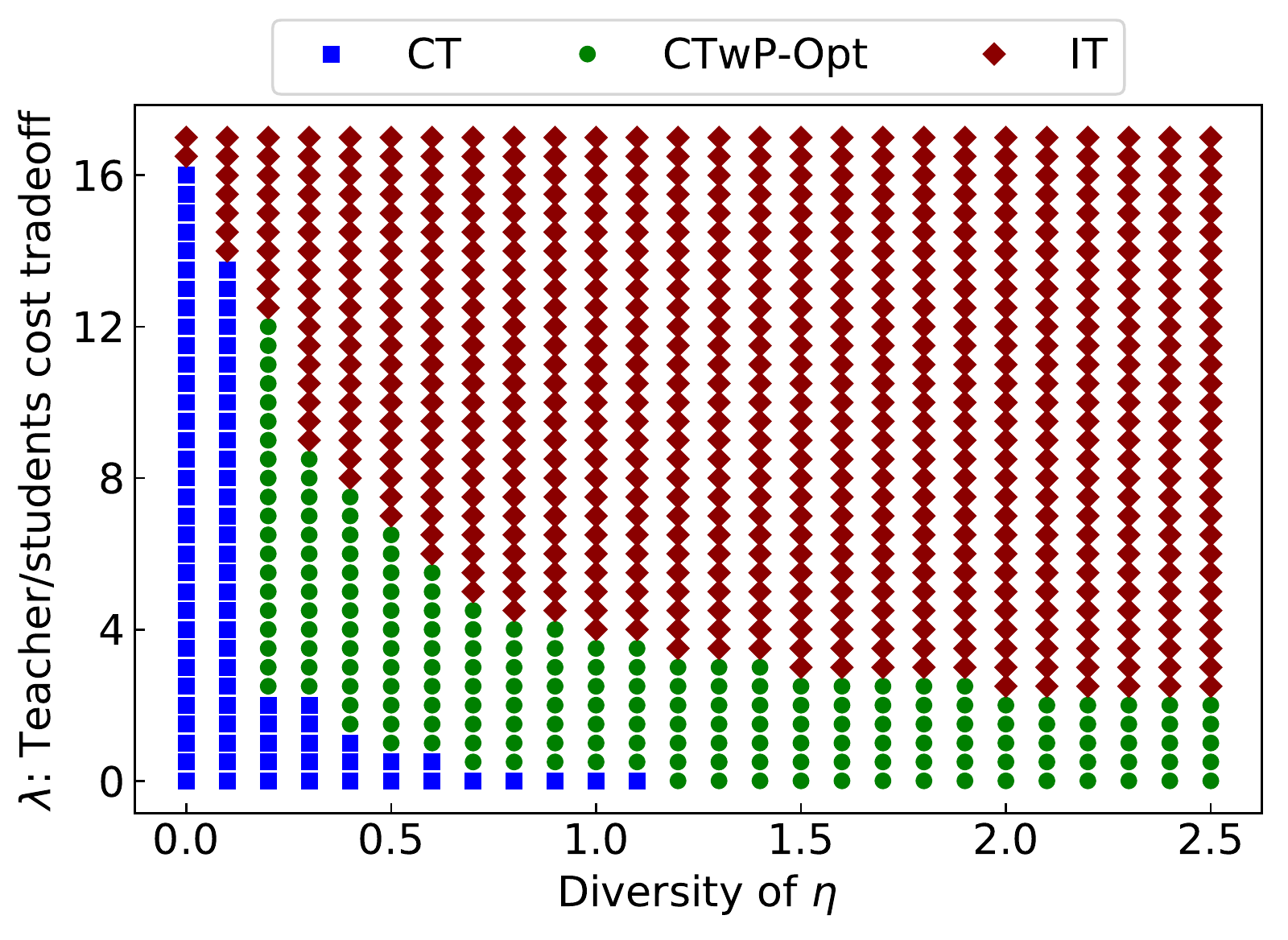}
        \captionsetup{format=hang}
        \caption{$\lambda$: Trade-off between teacher's and students' cost with increasing $\eta$ diversity.}
        \label{fig-cvg-noise-free-lambda-eta}
    \end{subfigure}
    \hspace{1mm}
    \begin{subfigure}[t]{0.29\textwidth}
        \includegraphics[width=\textwidth]{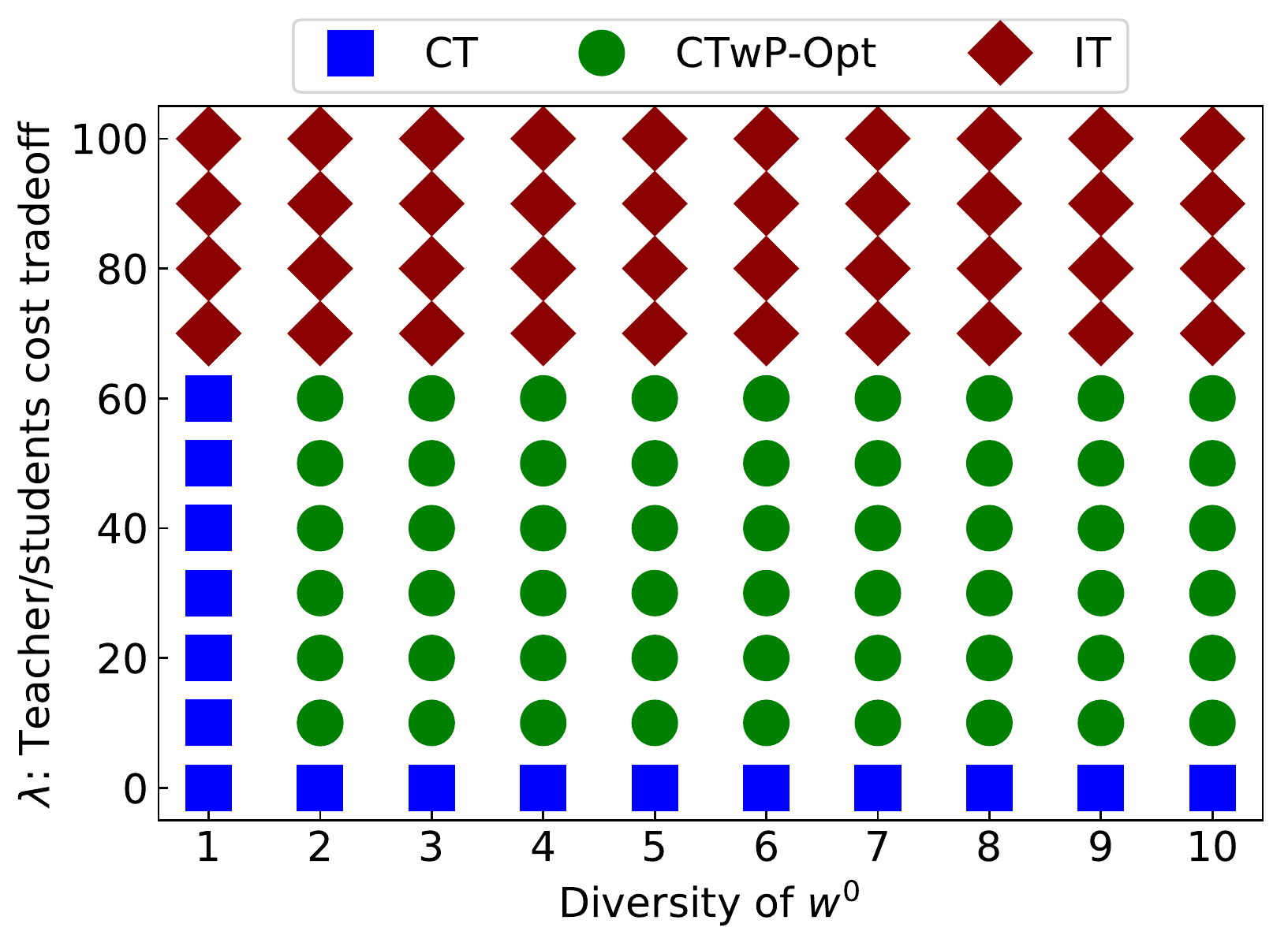}
        \captionsetup{format=hang}
        \caption{$\lambda$: Trade-off between teacher's and students' cost with increasing $w^0$ diversity.}
        \label{fig-cvg-noise-free-lambda-w0}
    \end{subfigure}    
    \caption{(\ref{fig-error-noise-free}) and (\ref{fig-cvg-noisy-delta}) show the convergence results for the noise-free and noisy settings. CT is robust and exhibits linear convergence. (\ref{fig-cvg-noise-free-eta}), (\ref{fig-cvg-noise-free-eta-student}) and (\ref{fig-cvg-noise-free-lambda-eta}) show the convergence results and trade-off for a classroom with diverse $\eta$. (\ref{fig-cvg-noise-free-lambda-w0}) shows the trade-off for a classroom with diverse $w^0$.}
    \label{fig-sim-cvg}
\end{figure*}

%    \caption{(\ref{fig-error-noise-free}) show the convergence results for the noise free. CT exhibits linear convergence. (\ref{fig-cvg-noise-free-eta}), (\ref{fig-cvg-noise-free-eta-student}) and (\ref{fig-cvg-noise-free-lambda-eta}) show the convergence results and trade-off for a classroom with diverse $\eta$.}
%    
%	(\ref{fig-error-noise-free}) shows this in terms of error plot, while (\ref{fig-cvg-noise-free-eta}) to (\ref{fig-cvg-noisy-delta}) shows this in terms of number of iterations required to achieve $\epsilon$ convergence. CT is robust to noise and performs much better than IT. Our proposed algorithms perform better that IT from the teachers perspective. However, from the students' perspective IT outperforms. (\ref{fig-cvg-noise-free-lambda-eta}) and (\ref{fig-cvg-noise-free-lambda-w0}) show how the optimal algorithm changes as we vary $\lambda$, the tradeoff parameter, and diversity of $\eta$ and $w^0$.
%				
%    ((\ref{fig-cvg-noisy-delta}) show the convergence results for the noisy setting. CT is robust and exhibits linear convergence. (\ref{fig-cvg-noise-free-lambda-w0}) shows the trade-off for a classroom with diverse $w^0$.

\subsection{Teaching Linear Models with Synthetic Data}\label{sec.experiments}
We first examine the performance of our teaching algorithms on simulated learners.

%\subsubsection{Setup}\label{sec.experiments.setup}
% As described in Section~\ref{sec.model}, we study the linear regression task under the squared loss function.
\textbf{Setup\ }
We evaluate the following algorithms: (i) classroom teaching (CT) - the teacher gives an example to the entire class at each iteration, (ii) CT with optimal partitioning (CTwP-Opt) - the class is partitioned as defined in Section~\ref{sec:partitioning}, (iii) CT with random partitioning (CTwP-Rand) - the class is randomly assigned to groups, and (iv) individual teaching (IT) - the teacher gives a tailored example to each student. An algorithm is said to converge when $\frac{1}{N}\sum_i\lVert w_i^t-w^*\rVert_2^2 \leq \epsilon$. We set the number of learners $N=300$ and accuracy parameter $\epsilon=0.1$.

\textbf{Average error and robustness of CT\ }
We first consider the noise free classroom setting with $d=25$, learning rates between $[0.05,0.25]$, and $D_\mathcal{X} = 2$. The plot of the error over time is shown in Figure~\ref{fig-error-noise-free}, together with the performance of four selected learners. Our algorithm exhibits linear convergence, as per Theorem \ref{main-exp-theorem}. The slower the learners and the further away they are from $w^*$, the longer they take to converge. Figure~\ref{fig-cvg-noisy-delta} shows how convergence is affected as the noise level, $\delta$, increases in the robust classroom teaching setting as described in Section~\ref{sec:robust-teaching}. Although the number of iterations required for convergence increases, it is still significantly lower than the noise-free IT.

% Two learners have the same initial starting point far away from $w^*$ while the other two have starting points closer to $w^*$. In each of these pairs, one learner has a faster learning rate of 0.15, the other a slower rate of 0.05. The results are as expected,

\textbf{Convergence for classroom with diverse $\eta$\ }
We study the effect of partitioning by $\eta$ on the performance of the algorithms described. The diversity of the classroom varies from 0 (where all learners in the classroom have $\eta=0.1$) to 0.5 (where for all learners $\eta \in [0.1,0.6]$ chosen randomly), and so on. Figure~\ref{fig-cvg-noise-free-eta} and Figure~\ref{fig-cvg-noise-free-eta-student} depict the number of iterations and number of examples needed by the teacher and students respectively to achieve convergence. As expected, IT performs best, and CTwP-Opt consistently outperforms CT. For a class with low diversity, partitioning is costly. However as diversity increases, partitioning is beneficial from the teachers' perspective. Note that the dip at a diversity of 0.15 for both plots is due to the value of $D_\mathcal{X}$. For the static $\gamma_t$, with $D_\mathcal{X}=2$, all learners with rates less than 0.25 will be negatively affected. As the minimum value of $\eta$ is 0.1, at zero diversity, all learners are affected the most. As diversity increases to 0.15, all learners are affected but to a lesser degree. Figure~\ref{fig-cvg-noise-free-lambda-eta} shows how the optimal algorithm, the one that minimizes cost, changes with $\lambda$ and diversity of $\eta$. When diversity is low and there is a low trade-off factor on the students' workload, CT performs best. At high values, IT has the lowest cost. CTwP-Opt falls between these two regimes.

\textbf{Convergence for classroom with diverse $w^0$\ }
Next, we study partitioning based on prior knowledge. We generate each cluster from a Gaussian distribution centered on a point along different axes. At diversity 1, all 300 learners are centered on a point on one axis, whereas at diversity 2, 150 learners are centered on one axis and the other 150 on another. Thus at 10, we have 30 learners around a point at each of the 10 axes. Each cluster represents one partition. Although the convergence plots from the teacher and students' perspective are not presented, they exhibit the same behaviour as partitioning by $\eta$. Figure~\ref{fig-cvg-noise-free-lambda-w0} shows the cost trade off plot in 10 dimensions as the number of clusters of $w^0$ increase. The results are the same as with $\eta$ partitioning and CTwP-Opt outperforms in most regimes.

%% file: 6.2_experiments-realworld.tex
\subsection{Teaching How to Classify Butterflies and Moths}\label{sec.experiments.realworld}

We now demonstrate the performance of our teaching algorithms on a binary image classification task for identifying insect species, a prototypical task in crowdsourcing applications and an important component in citizen science projects such as eBird \cite{sullivan2009ebird}.

\begin{figure*}[h]
    \captionsetup[subfigure]{font=scriptsize,labelfont=scriptsize}
    \centering
    \begin{subfigure}[t]{0.27\textwidth}
        \includegraphics[width=\textwidth]{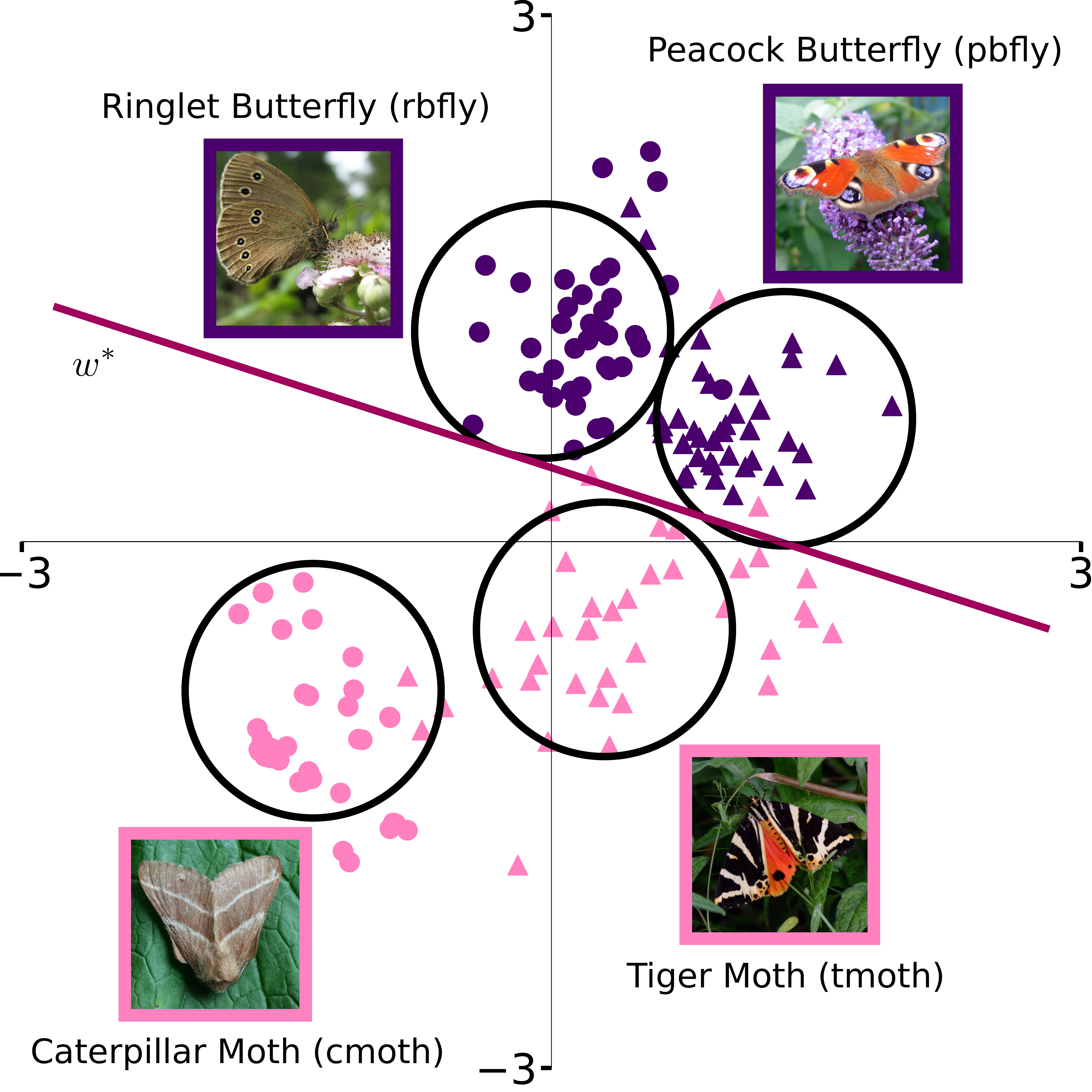}
        \caption{Dataset of images $\mathcal{X}$ and target $w^*$}
        \label{fig-bm-initial-dataset}
    \end{subfigure}
    \begin{subfigure}[t]{0.27\textwidth}
        \includegraphics[width=\textwidth]{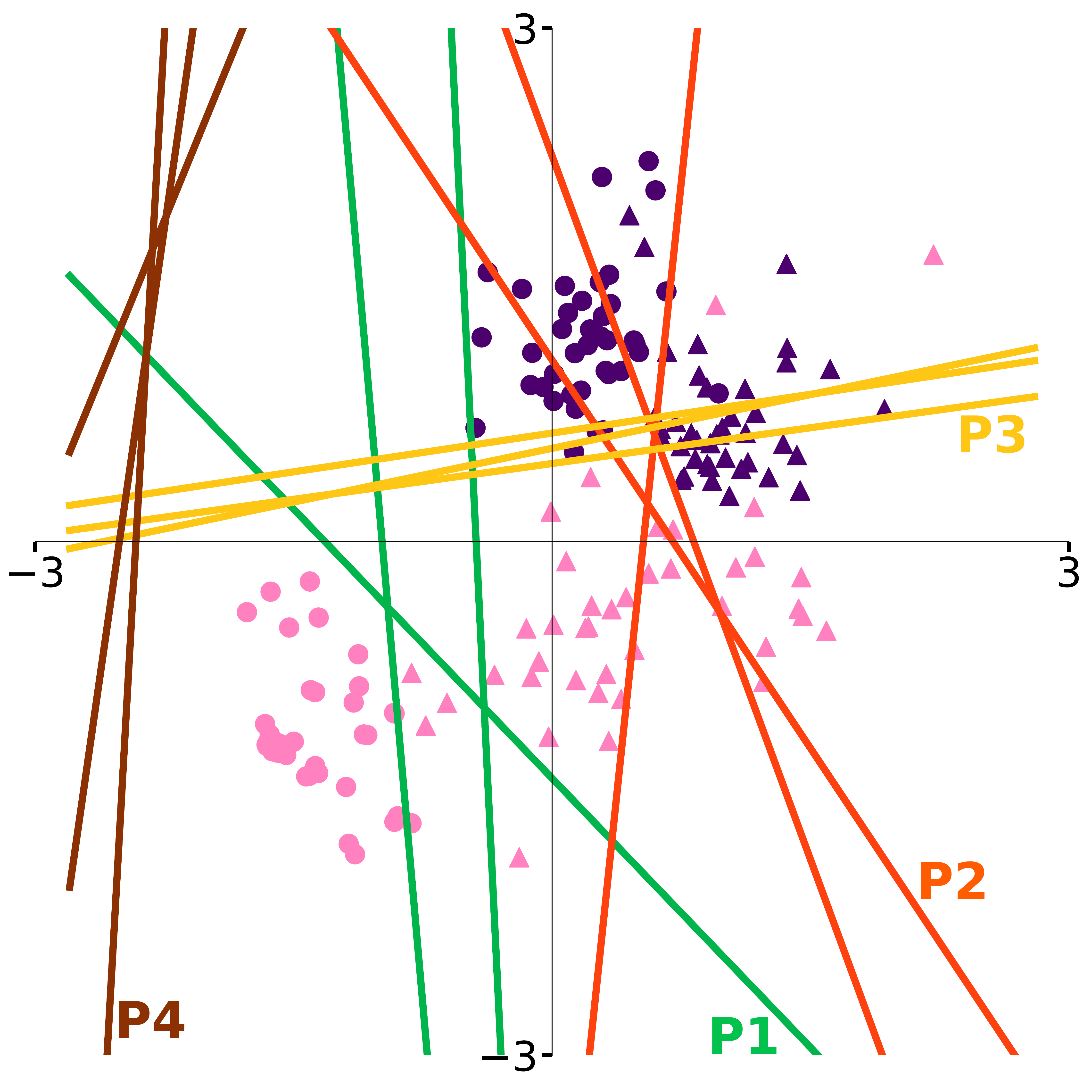}
        \captionsetup{format=hang}
        \caption{Initial $w^0$ of 4 types of learners}
        \label{fig-bm-initial-learners}
    \end{subfigure}
    \begin{subfigure}[t]{0.42\textwidth}
        \includegraphics[width=\textwidth]{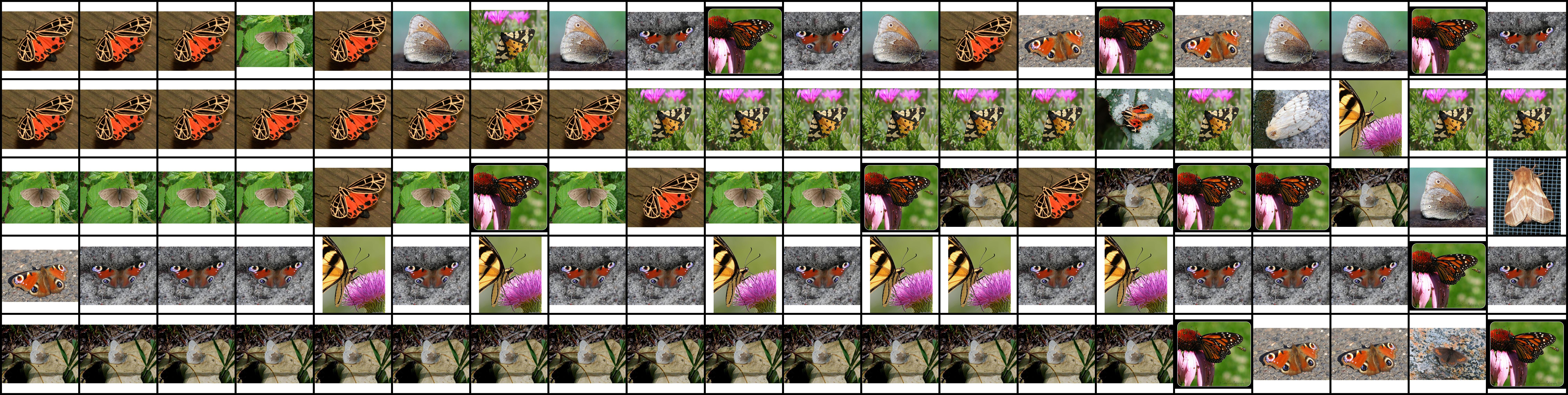}
        \caption{Teaching examples, visualized twice every 10 iterations.}
        \label{fig-bm-thumbnails}
    \end{subfigure}
    \hspace{1mm}
    \begin{subfigure}[t]{0.29\textwidth}
        \includegraphics[width=\textwidth]{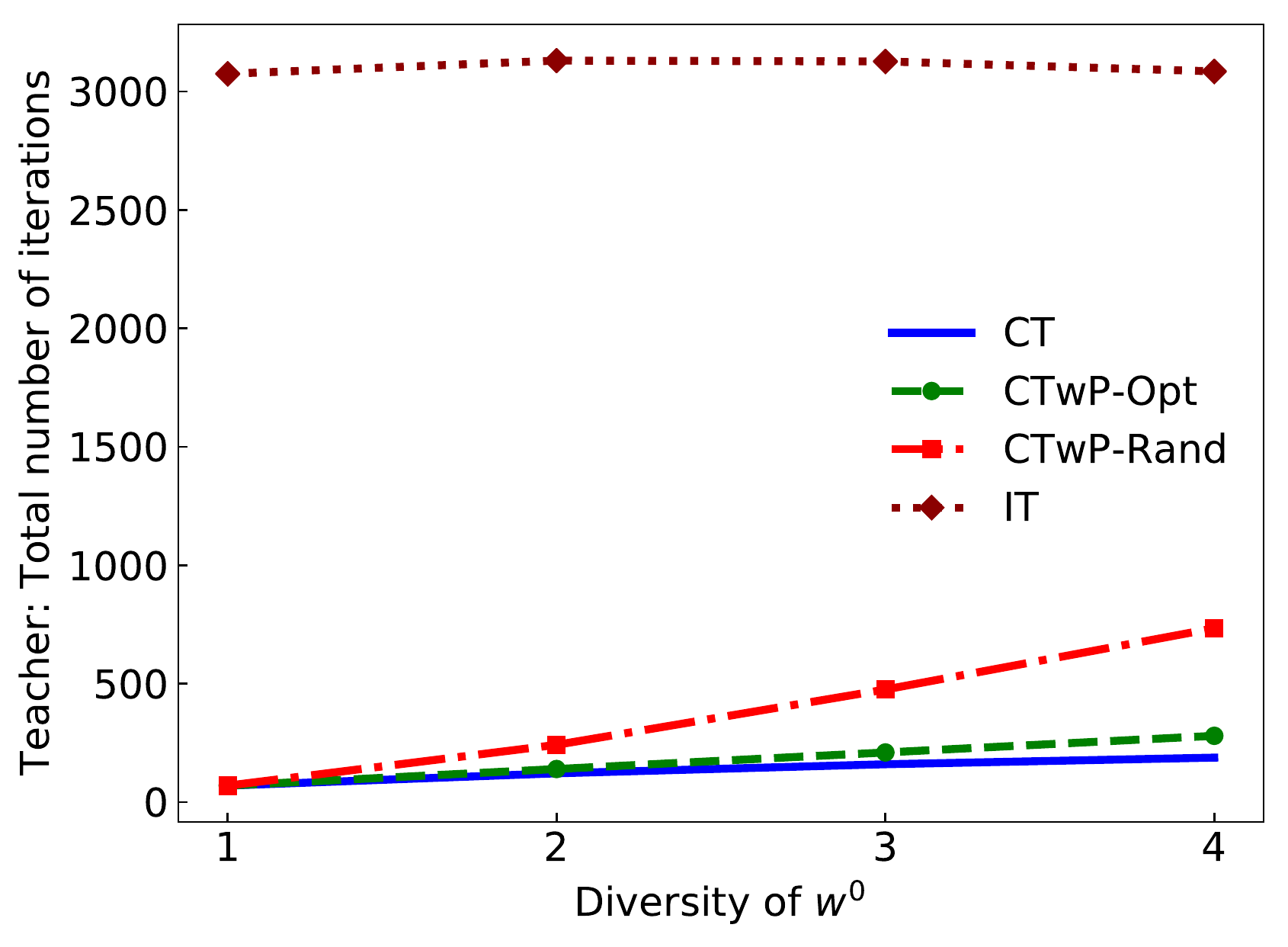}
        \captionsetup{format=hang}
        \caption{Total iterations needed for convergence from the teacher's perspective}
        \label{fig-bm-cvg-t}
    \end{subfigure}   
   \hspace{1mm}
    \begin{subfigure}[t]{0.29\textwidth}
        \includegraphics[width=\textwidth]{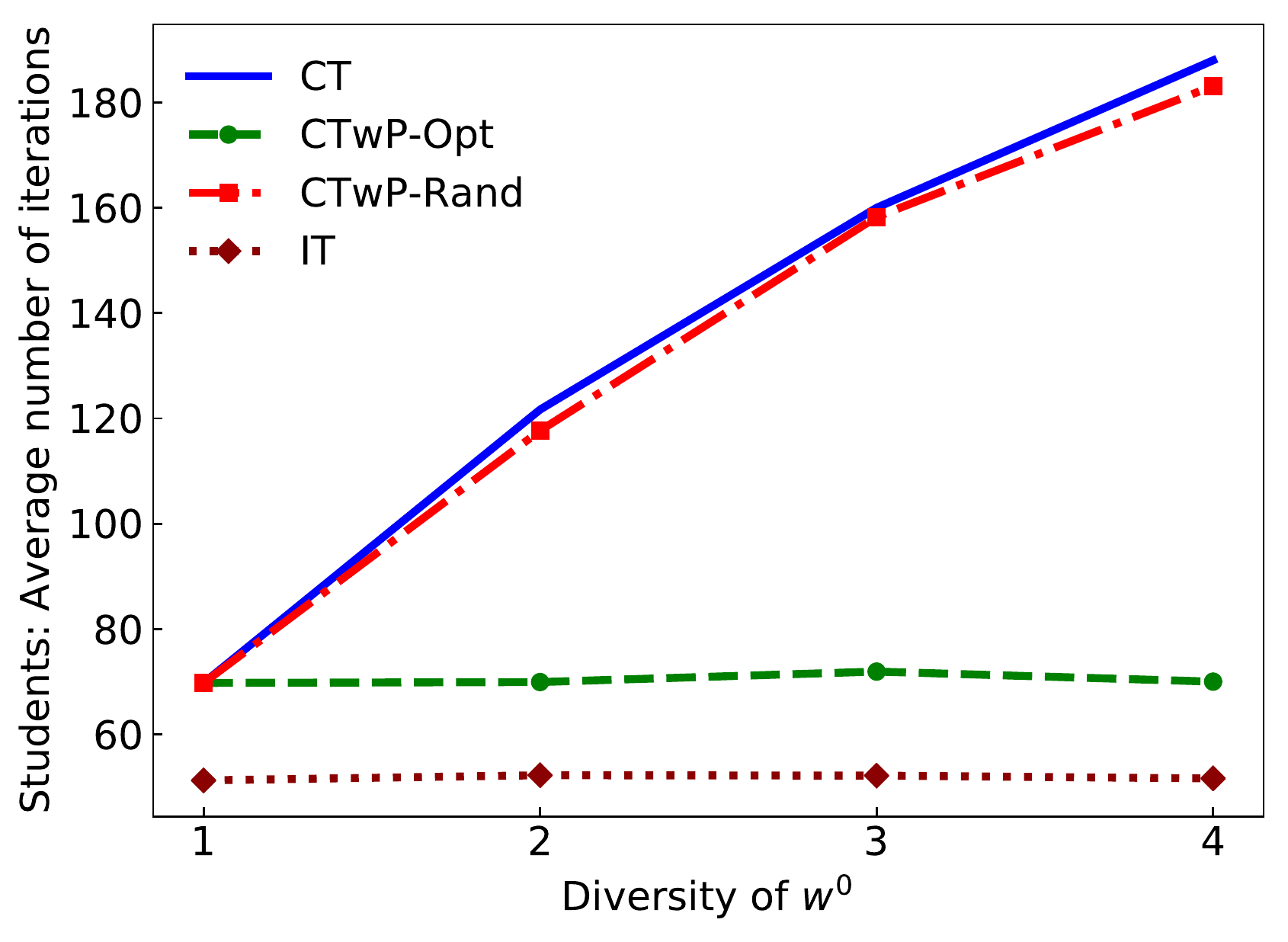}
        \caption{Total iterations needed for convergence from the student's perspective}
        \label{fig-bm-cvg-s}
    \end{subfigure}    
   \hspace{1mm}
    \begin{subfigure}[t]{0.29\textwidth}
        \includegraphics[width=\textwidth]{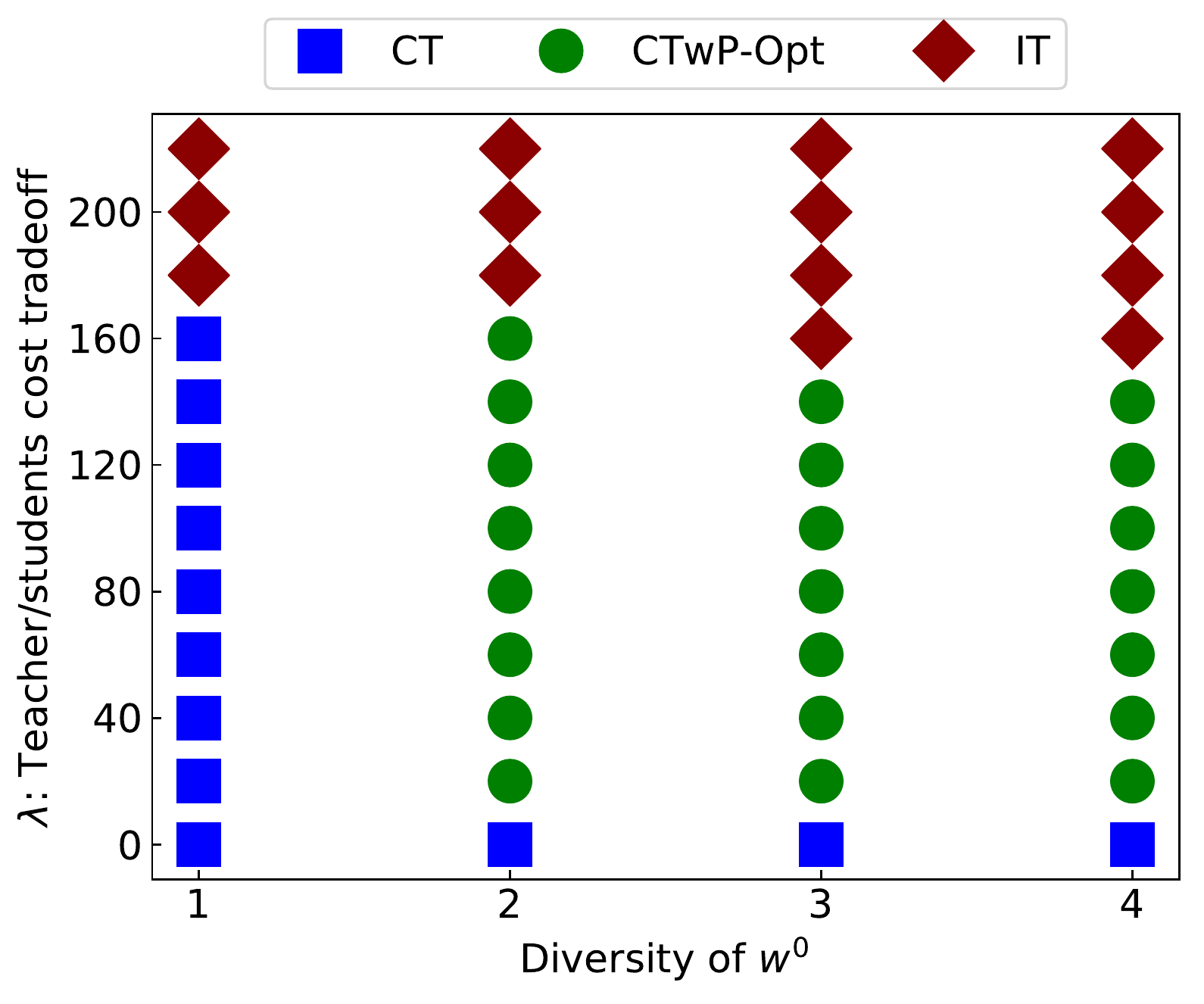}
        \captionsetup{format=hang}
        \caption{$\lambda$: Trade-off between teacher's and students' cost with increasing $w_0$ diversity}
        \label{fig-bm-cvg-lambda}
    \end{subfigure}
    \caption{(\ref{fig-bm-initial-dataset}) shows a low-dimensional embedding of the dataset and the target concept. (\ref{fig-bm-initial-learners}) shows an embedding of the initial states of three learners of each of the 4 types. (\ref{fig-bm-thumbnails}) are  training examples selected by CT and CTwP-Opt teachers when the class has diversity 4. (\ref{fig-bm-cvg-t}) and (\ref{fig-bm-cvg-s}) show the number of iterations required to achieve $\epsilon$-convergence from the teacher and student perspectives. (\ref{fig-bm-cvg-lambda}) shows how the optimal algorithm changes as we vary the trade-off parameter $\lambda$, and diversity of the class.}    
    \label{fig:cvg}
\end{figure*}

\textbf{Images and Euclidean embedding\ }
We use a collection of 160 images (40 each) of four species of insects, namely (a) Caterpillar Moth (cmoth), (b) Tiger Moth (tmoth), (c) Ringlet Butterfly (rbfly), and (d) Peacock Butterfly (pbfly), to form the teaching set $\mathcal{X}$. Given an image, the task is to classify if it is a butterfly or a moth. However, we need a Euclidean embedding of these images so that they can be used by a teaching algorithm. Based on the data collected by \cite{singla2014near}, we obtained binary labels (whether a given image is a butterfly or not) for $\mathcal{X}$ from a set of 67 workers from Amazon Mechanical Turk. Using this annotation data, the Bayesian inference algorithm of \cite{welinder2010multidimensional} allows us to obtain an embedding, shown in Figure~\ref{fig-bm-initial-dataset}, along with the target $w^*$ (the best fitted linear hypothesis for $\mathcal{X}$).

%These images form the teaching set $\mathcal{X}$. However, we need a Euclidean embedding of these images so that they can be used by a teaching algorithm. Based on the data collected by \cite{singla2014near}, we obtained binary labels (whether a given image is a butterfly or not) for these 160 images from a set of 67 workers from Amazon Mechanical Turk (AMT). Using this annotation data, the Bayesian inference algorithm of \cite{welinder2010multidimensional} allows us to obtain an embedding---Figure~\ref{fig-bm-initial-dataset} represents such a 2D embedding for $\mathcal{X}$ along with the target $w^*$ (the best fitted linear hypothesis for $\mathcal{X}$).

%describe a Bayesian inference algorithm t
%By using the We used the data coll approach of \citet{welinder2010multidimensional} that allows to obtain an embedding
%We obtained  Based on the data collected by authors of \cite{singla2014near} , we obtained these images

%\citet{welinder2010multidimensional} describe a Bayesian inference algorithm to estimate a low-dimensional embedding of the data labelled by a multitude of annotators. Figure \ref{fig-bm-initial-dataset} represents such a 2D embedding for our data along with the target concept $w^*$.

%%\vspace{-2mm}
% class
%The dataset consists of linear hypotheses of 67 students obtained from Amazon Mechanical Turk. These are determined from the binary labels provided by each student classifying images in $\mathcal{X}$ as moths or butterflies.
%\vspace{-4mm}
%\paragraph{Learners' hypotheses}
{\bfseries Learners' hypotheses\ }
The process described above to obtain the embedding in Figure~\ref{fig-bm-initial-dataset} simultaneously generates an embedding of each of the $67$ annotators as linear hypotheses in the same 2D space. Termed as ``schools of thought" by \cite{welinder2010multidimensional}, these hypotheses capture various real-world idiosyncrasies in the AMT workers' annotation behavior. For our experiments, we identified four types of learners' hypotheses; those who (i) \textit{misclassify tmoth as butterfly (P1)}, (ii) \textit{misclassify rbfly as moth (P2)}, (iii) \textit{misclassify pbfly as moth (P3)} and (iv) \textit{misclassify tmoth and cmoth as butterflies}. Figure \ref{fig-bm-initial-learners} shows an embedding of three distinct hypotheses each of the four types of learners.

{ \bfseries Creating the classroom\ }
 We denote the hypotheses described above as initial states $w^0$ of the learners/students. Due to sparsity of data, we create a supersample of size $60$ for the four types of learners by adding a small noise. We set the classroom size $N = 60$. The diversity of the class, defined by the number of different types of learners present, varies from 1 to 4. Thus, diversity of $1$ refers to the case when all $60$ learners are of same type (randomly picked from P1, P2, P3, or P4), and diversity of $4$ refers to the case when there are $15$ learners of each type. We set a constant learning rate of $\eta = 0.05$ for all students.

 \textbf{Teaching and performance metrics\ }
 We study the performance of CT, CTwP-Rand, and IT teachers. We also examine the CTwP-Opt teacher that partitions the learners of the class based on their types. All teachers are assumed to have complete information about the learners at all times. We set accuracy parameter $\epsilon = 0.2$ and the classroom is said to have converged when $\frac{1}{N}\sum_i\lVert w_i^t-w^*\rVert_2^2 \leq \epsilon$.

\textbf{Teaching examples\ }
Figure \ref{fig-bm-thumbnails} consists of 5 rows of 20 thumbnail images each, depicting the training examples chosen in an actual run of the experiment when the diversity of the classroom is 4. The first row corresponds to the images chosen by CT. For instance, in iteration 1, CT chooses a tmoth example. While this example is most helpful for learners in P1 (confusing tmoths as butterflies), however, learners in P2 and P3 would have benefited more from seeing examples of butterflies. This increases the workload for the learners.  The next four rows in Figure \ref{fig-bm-thumbnails} correspond to the images chosen by CTwP-Opt when teaching partitions P1, P2, P3, and P4 respectively---these thumbnails show the personalized effect given the homogeneity of these partitions. For instance, for P1, the CTwP-Opt focuses on choosing tmoth examples thereby allowing these learners to converge faster while ensuring that the cost for learners in other partitions does not increase.
%We refer the reader to the Appendix\ref{sec.experiments.realworld.app} for further details regarding the experimental setup and discussions of results.

%\vspace{-2mm}

%\vspace{-4mm}

%  \begin{subfigure}[b]{0.9\textwidth}
%  \includegraphics[width=\textwidth, height=3.2cm]{lplots_2d_100l.png}
%  \caption{Hyperplanes for the first and next lowest and highest $\eta$. The intensity of the line indicates the order at which the the student is updating $w^t$. The lightest line denotes the initial $w^0_i$ and darkest the final $w^t_i$. The black line represents $w^*$. The examples given by the teacher are also plotted with the same colour code as the hyperplane.}
%  \label{fig:hyp}
%  \end{subfigure}
%  \caption{Results with 100 students}
%  \label{fig1}
%\end{figure}

% \subsection{Experimental Setup}\label{exp.bm.setup}
% \vspace{-1mm}
 %\vspace{-3pt}

 %\vspace{-2mm}
\textbf{Convergence\ }
Figure \ref{fig-bm-cvg-t} compares the performances of the teachers in terms of the total number of iterations required for convergence. CT performs optimally because every example chosen is provided to the entire class; CTwP-Opt requires only a few examples more, given the homogeneity of the partition and the partitions being of equal size.
%This is because CTwP-Opt constructs examples only for a single partition at any given time step. As a result, learners in other partitions do not make an update.
IT constructs individually tailored examples for each learner in the class. Thus the combined number of iterations is much higher in comparison.

\textbf{Teacher/students cost trade-off\ } On the other hand, Figure \ref{fig-bm-cvg-s} depicts the average number of examples required by each learner to achieve convergence as a function of diversity. This represents the learning cost from the students' persective. IT performs best because the teacher chooses personalized examples for each learner. CTwP-Opt performs considerably better than CT. This happens because partitioning groups together learners of the same type. Figure \ref{fig-bm-cvg-lambda} represents optimal algorithm given the diversity of the class and the trade-off factor $\lambda$ as defined in Section~\ref{sec:partitioning}. As diversity increases, CTwP-Opt outperforms the other teachers in terms of the total cost.

%%%%%%%%%%%%%%%%%%%%%%%%%%%%%%%%%%%%%%%%%%%%%%%%%%%%%%%%% OLD
%%%%%%%%%%%%%%%%%%%%%%%%%%%%%%%%%%%%%%%%%%%%%%%%%%%%%%%%%

%% file: 6.3_experiments-realworld-cowriter.tex
%!TEX root = main.tex
%%%%%%%%%%%%%%%%%%%%%%%%%%%%%%%%%%%%%%%%%%%%%%%%%%%%%%%%%
%%%%%%%%%%%%%%%%%%%%%%%%%%%%%%%%%%%%%%%%%%%%%%%%%%%%%%%%%

%%%%%%%%%%%%%%%%%%%%%%%%%%%%%%%%%%%%%%%%%%%%%%%%%%%%%%%%% copied from 6.3
\begin{figure*}[h]
    \captionsetup[subfigure]{font=scriptsize,labelfont=scriptsize}
    \centering
    \begin{subfigure}[t]{0.323\textwidth}
        \includegraphics[width=\textwidth,height=0.8cm]{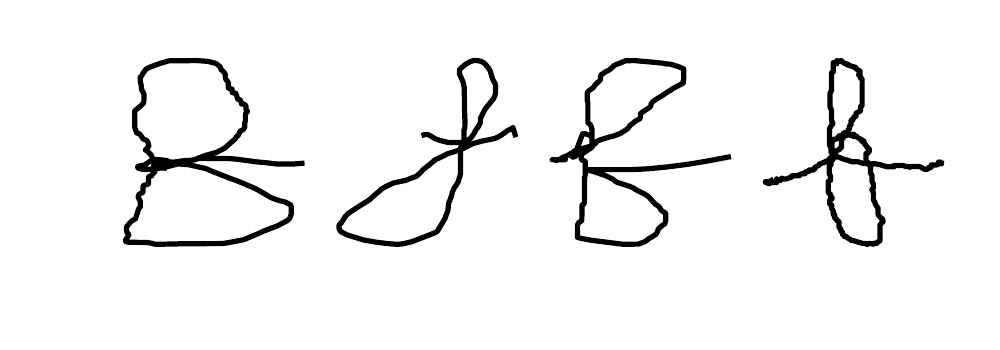}
        \caption{Shaky and distorted handwriting}
        \label{fig-samples-rotated-shaky}
    \end{subfigure}
    \begin{subfigure}[t]{0.33\textwidth}
        \includegraphics[width=\textwidth,height=0.8cm]{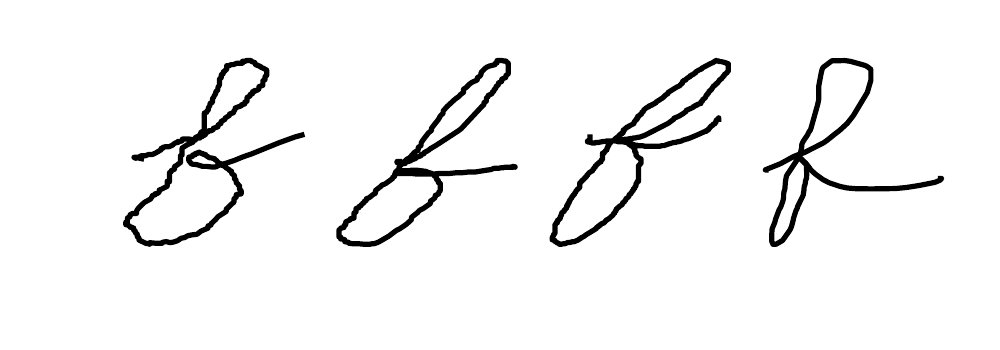}
        \captionsetup{format=hang}
        \caption{Shaky and rotated handwriting}
        \label{fig-samples-distorted-shaky}
    \end{subfigure}
    % \hspace{0.5mm}
    \begin{subfigure}[t]{0.33\textwidth}
        \includegraphics[width=\textwidth,height=0.8cm]{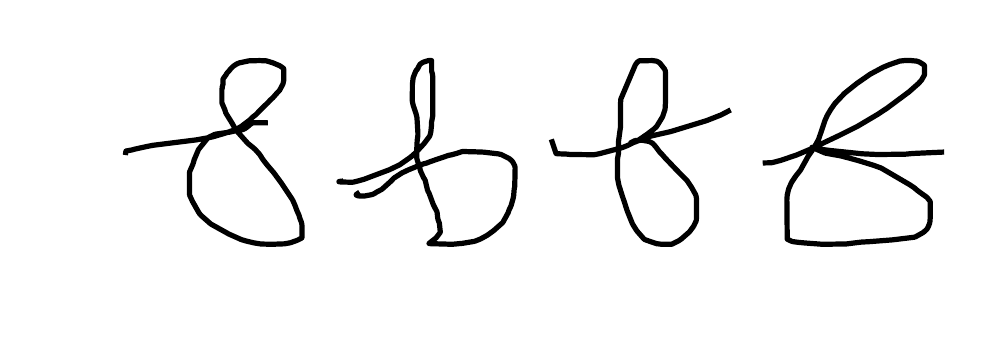}
        \captionsetup{format=hang}
        \caption{Rotated and distorted handwriting}
        \label{fig-samples-distorted-rotated}
    \end{subfigure}
%   \hspace{1mm}
    \caption{(\ref{fig-samples-rotated-shaky}) to (\ref{fig-samples-distorted-rotated}) shows samples of children's handwriting where two of the three defined features are poor and the third is good.}
    \label{fig:cvg}
\end{figure*}

\begin{figure*}[h]
    \captionsetup[subfigure]{font=scriptsize,labelfont=scriptsize}
    \centering
    \begin{subfigure}[t]{0.24\textwidth}
        \includegraphics[width=\textwidth,height=3.2cm]{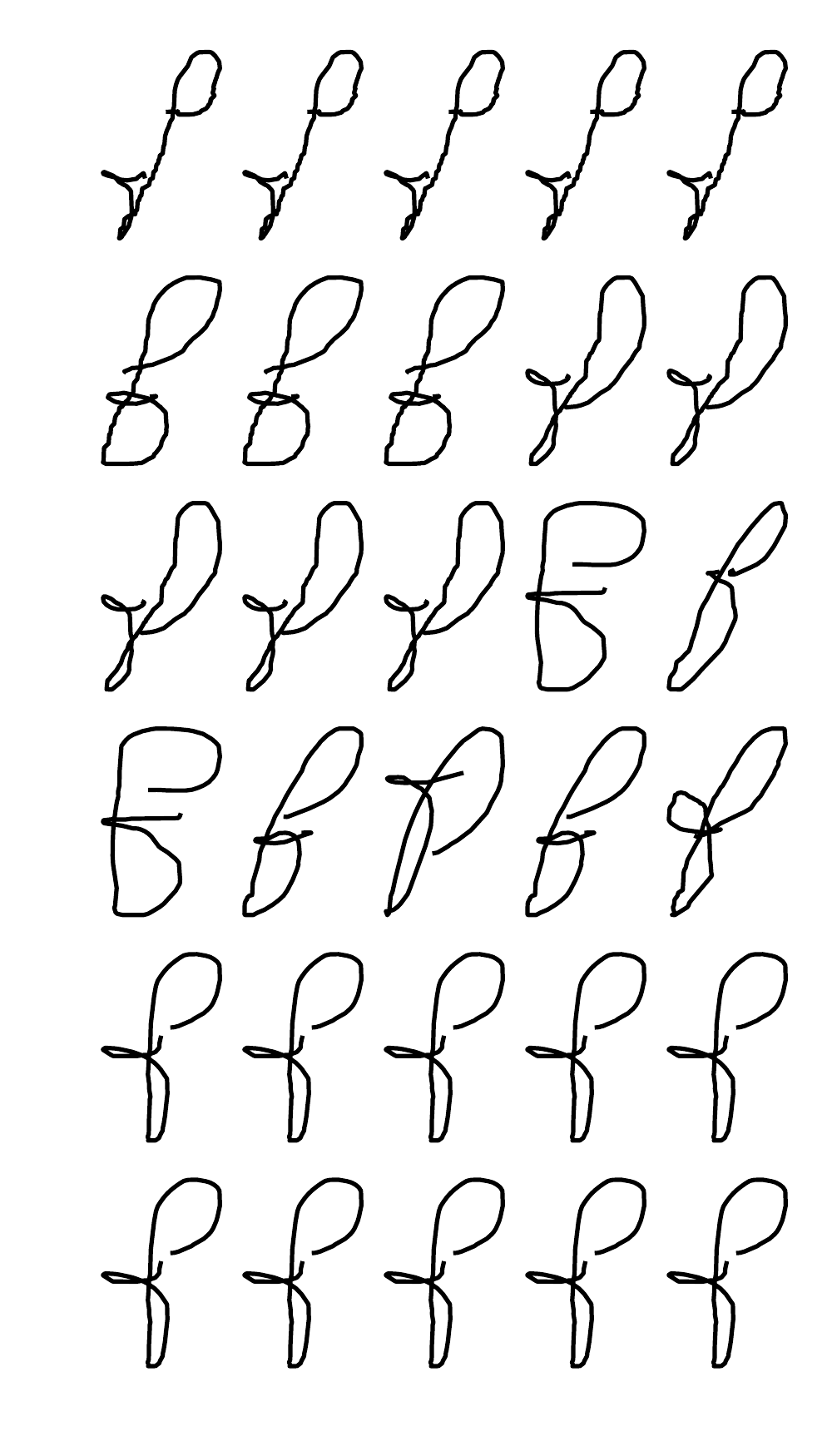}
        \caption{Teaching examples for shaky and rotated handwriting}
        \label{fig-rotated-shaky}
    \end{subfigure}
    \begin{subfigure}[t]{0.24\textwidth}
        \includegraphics[width=\textwidth,height=3.2cm]{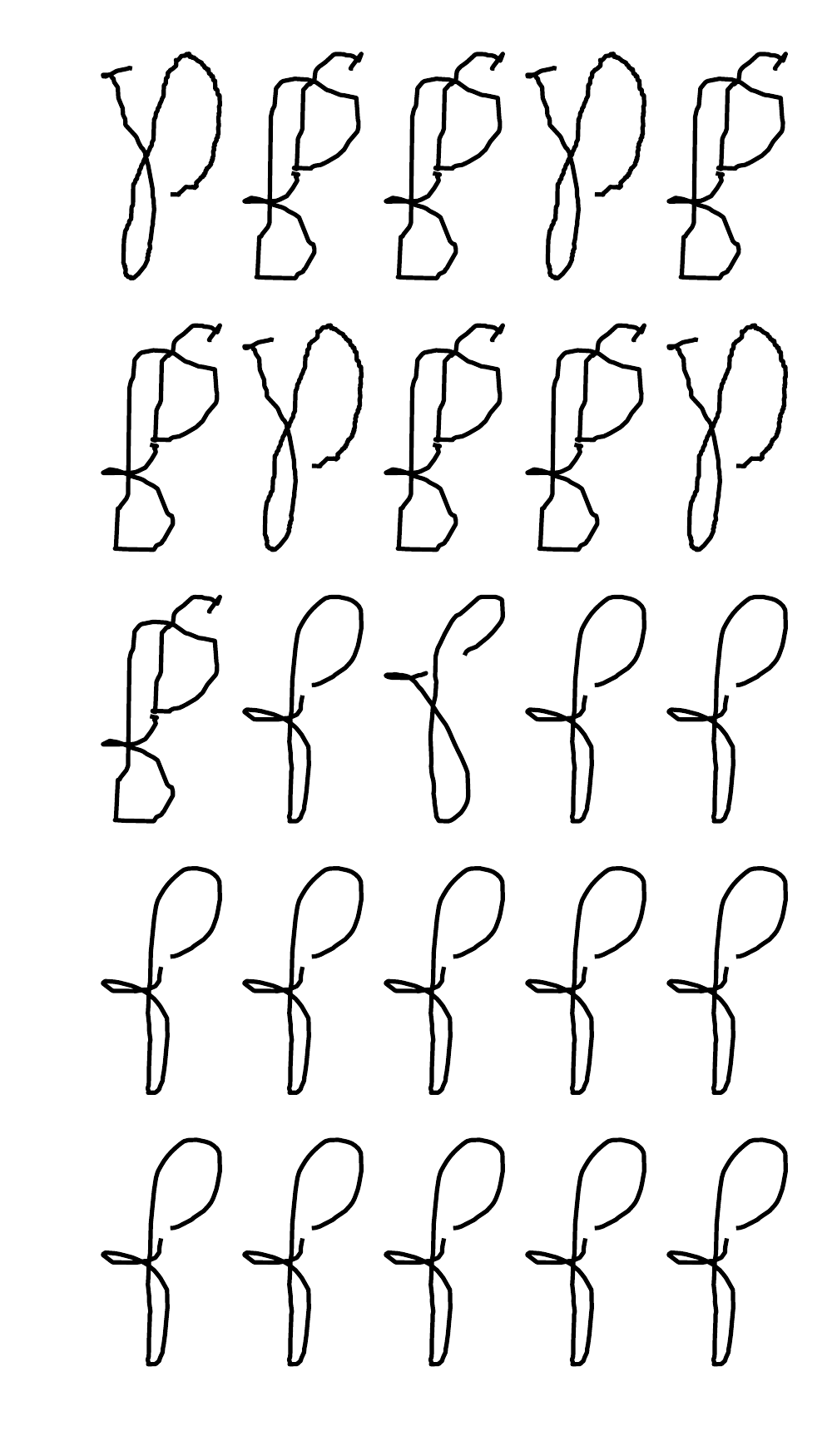}
%        \captionsetup{format=hang}
        \caption{Teaching examples for distorted and shaky handwriting}
        \label{fig-distorted-shaky}
    \end{subfigure}
    % \hspace{0.5mm}
    \begin{subfigure}[t]{0.24\textwidth}
        \includegraphics[width=\textwidth,height=3.2cm]{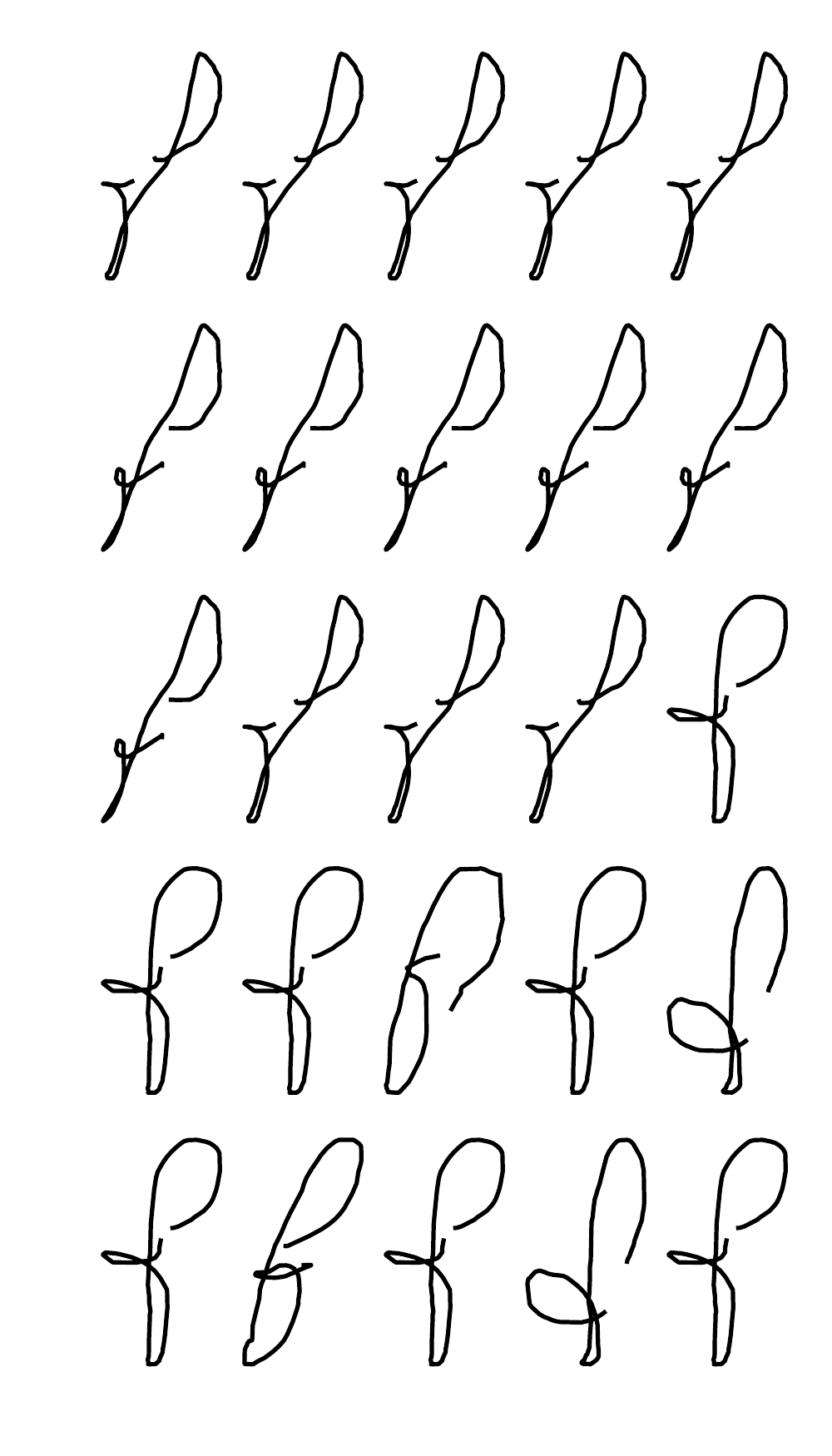}
%        \captionsetup{format=hang}
        \caption{Teaching examples for distorted and rotated handwriting}
        \label{fig-distorted-rotated}
    \end{subfigure}
%   \hspace{1mm}
    \begin{subfigure}[t]{0.24\textwidth}
        \includegraphics[width=\textwidth,height=3.2cm]{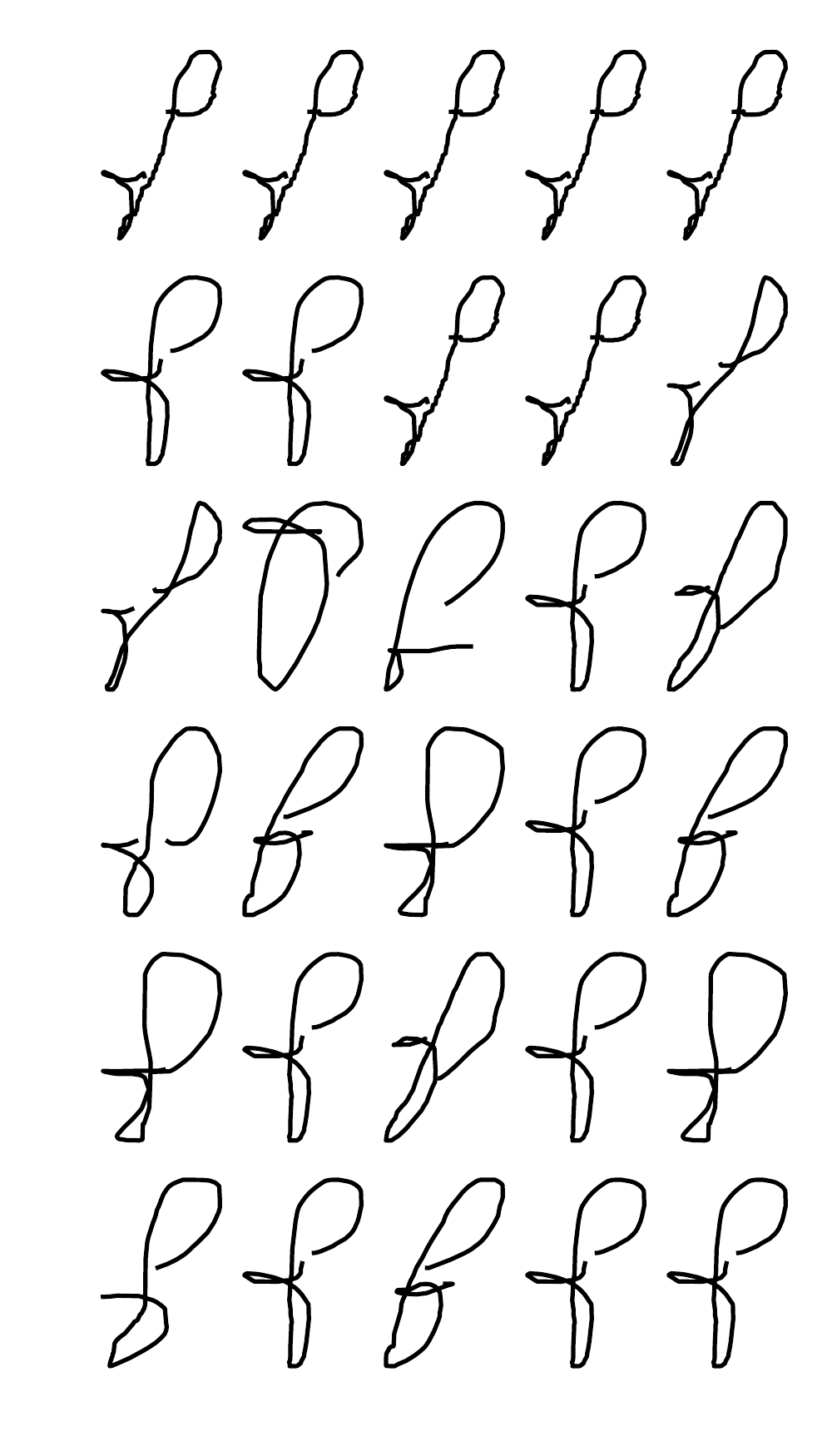}
        \caption{Teaching examples for distorted, rotated, shaky handwriting}
        \label{fig-distorted-rotated-shaky}
    \end{subfigure}

    \caption{(\ref{fig-rotated-shaky}) to (\ref{fig-distorted-rotated-shaky}) shows the sequence of examples, visualized every other iteration, chosen by our algorithm for different initial hypothesis of the children.}
    \label{fig:cvg}
\end{figure*}
%%%%%%%%%%%%%%%%%%%%%%%%%%%%%%%%%%%%%%%%%%%%%%%%%%%%%%%%%

\subsection{Teaching How to Write}\label{sec.experiments.realworld.cowriter}

Despite formal training, between 5\% to 25\% of children struggle to acquire handwriting skills. Being unable to write legibly and rapidly limits a child's ability to simultaneously handle other tasks such as grammar and composition which may lead to general learning difficulties \cite{feder2007handwriting,christensen2009critical}. 
\cite{johal2016child} and \cite{chase2009teachable} adopt an approach where the child plays the role of the ``teacher'' and an agent a ``learner'' that needs help. This method of learning by teaching boosts a child's self esteem and increases their commitment to the task as they are given the role of the one who ``knows and teaches'' \cite{rohrbeck2003peer,chase2009teachable}. In our experiments, a robot iteratively proposes a handwriting adapted to the handwriting profile of the child, that they try to correct (\textit{cf.} Figure \ref{fig-cowriter-seq}). We now demonstrate the performance of our algorithm in choosing this sequence of examples.
%The CoWriter project aims to help these children with an original approach: the child plays the role of the "teacher" and a robot a "learner" that needs help with improving its handwriting \cite{hood2015cowriter}. 

% It is therefore crucial to detect and address handwriting difficulties as early as possible .

%The CoWriter research project aims to help these children with an original approach: the child plays the role of the "teacher" and a robot is a "learner" requiring help to improve its handwriting (Figure \ref{fig-cowriter}) \cite{hood2015cowriter}. This approach is called learning by teaching boosts the child's self esteem as he/she is given the role of the one who "knows and teaches" \cite{rohrbeck2003peer}. With this added responsibility, child also feels more motivated and committed to the task \cite{chase2009teachable}. 

%\subsection{Dataset}\label{exp.bm.dataset}

% \paragraph{Images and euclidean embedding\}
\textbf{Generating handwriting dynamics\ } 
A LSTM is used to learn and generate handwriting dynamics \cite{graves2013generating}. It is trained on children's handwriting data collected from 1014 children from 14 schools.\footnote{The model has 3 layers, 300 hidden units and outputs a 20 component bivariate Gaussian mixture and a Bernoulli variable indicating the end of the letter. Each child was asked to write, in cursive, the 26 letters of the alphabet and the 10 digits on a tablet.} In the Appendix, we showed that the pool of samples has to be rich enough for teaching to be effective.\footnote{The attained result is for the squared loss function, however, the analysis holds for other loss function.} As our generative model outputs a distribution, we can sample from it to get a diverse set of teaching examples. We analyze our results for a cursive ``f'', similar results apply for the other letters.
% (\textit{cf.} Eq. \ref{eq-pooled-based})

\textbf{Handwriting features\ }
%Several handwriting tests were invented to evaluate children's handwriting quality, with the Concise Evaluation Scale (BHK) \cite{hamstra1987concise} being the standard in Europe.
Concise Evaluation Scale (BHK) \cite{hamstra1987concise} is a standard handwriting test used in Europe to evaluate children's handwriting quality. We adopt features such as (i) distortion, (ii) rotation, (iii) shakiness, and label each generated sample with a score for each of these features.

%%\paragraph{Learners' hypotheses} Given a child's handwriting sample, we estimate their initial hypothesis $w_0$ by how well each of the above features have been written, in a similar fashion to the scoring of samples.

%In this experiment, we take the hypothesis that each student will always learn with the input proposed. This assumption will obviously not be the case in real life. However, our model can easily be adapted to take that into account.

\textbf{Creating the classroom\ }
Given a child's handwriting sample, we estimate their initial hypothesis, $w_0$ by how well each of the above features have been written, in a similar fashion to the scoring of samples. As most children fair poorly in two out of the three features, we selected and partitioned them according to the following three types of handwriting characteristics, substantial (i) shakiness and rotation, (ii) distortion and shakiness, and (iii) distortion and rotation. Original samples of each are shown in Figures~\ref{fig-samples-rotated-shaky} to \ref{fig-samples-distorted-rotated}. We set a constant learning rate for all learners.

%\subsection{Results}\label{exp.bm.results}

\textbf{Teaching examples\ }
Figures~\ref{fig-rotated-shaky} to \ref{fig-distorted-rotated} shows the training examples chosen by CTwP-Opt and Figure \ref{fig-distorted-rotated-shaky} by our CT algorithm. Each of the synthesized handwriting samples are labelled as good or bad, based on the average of their normalized scores. We then run a classification algorithm on each partition and the entire class. This returns a sequence of examples that the robot would propose, for the children to correct. For children with handwriting that is shaky and rotated but not distorted, the sequence of examples chosen by our algorithm shows examples that are not distorted but progressively smoother and upright. Similarly, for children with handwriting that is distorted and shaky, the sequence of examples shown is upright with decreasing distortion and shakiness. We did not show the convergence plots as they have similar characteristics as those from the previous experiments.

%% file: 7_conclusion.tex
\section{Conclusion}\label{sec.conclusion}

We studied the problem of constructing an optimal teaching sequence for a classroom of online gradient descent learners. In general, this problem is non-convex, but for the squared loss, we presented and analyzed a teaching strategy with linear convergence. We achieved a sample complexity of $\mathcal{O} \br{\min\bc{d,N} \log \frac{1}{\epsilon}}$, which is a significant improvement over $\mathcal{O} \br{N \log \frac{1}{\epsilon}}$ samples as required by the individual teaching strategy. We also showed that a homogeneous grouping of learners allows us to achieve a good trade-off between the learners' workload and the teacher's orchestration cost.  Further, we compared the individual teaching (IT), classroom teaching (CT), and classroom teaching with partitioning (CTwP): we showed that a homogeneous grouping of learners (based on learning ability or prior knowledge) allows us to achieve a good trade-off between the learners' workload and the teacher's orchestration cost. The sequence of examples returned by our experiments are interpretable and they clearly demonstrate a significant potential in automation for robotics.

%% file: 8.1_appendix_additional-robust-settings.tex
%!TEX root = main.tex
%%%%%%%%%%%%%%%%%%%%%%%%%%%%%%%%%%%%%%%%%%%%%%%%%%%%%%%%%
%%%%%%%%%%%%%%%%%%%%%%%%%%%%%%%%%%%%%%%%%%%%%%%%%%%%%%%%%
\section{Additional Robust Teaching Settings}

\subsection{Noise in $W^t$}
\label{subsec:noise-Wt}
Assume that the teacher only receives the noisy version of $W^t$ given by 
\begin{equation}
\label{noisy-Wt-model-eq}
\tilde{W}^t ~:=~ W^t + \delta^t ,
\end{equation}
where $\delta^t$ is some random noise matrix such that $\lambda_1 \br{\delta^t} \leq \frac{\alpha_{\mathrm{min}} \epsilon}{2 \br{d-1}}$. Then the teacher constructs the example as follows:
\begin{align}
\hat  x^t ~:=~& \argmax_{x: \norm{x}=1} x^\top \tilde{W}^t x ~=~ e_1 \br{\tilde{W}^t} \nonumber \\
x^t ~:=~& \gamma_t \hat  x^t \text{ and } y^t = \ip{w^*}{x^t} , \label{noise-cap-Wt-example}
\end{align}
where $\gamma_t$ satisfies the condition given in \eqref{gamma-t-conditions}. 
In this setting also, the classroom teaching is possible with linear convergence, as shown in the following theorem.
\begin{theorem}
\label{main-exp-theorem-noise-Wt}
Consider the noisy observation setting given by \eqref{noisy-Wt-model-eq}. Let $k := \max_t\bc{\mathrm{rank}\br{\tilde{W}^t}}$ where $\tilde{W}^t$ is given by \eqref{noisy-Wt-model-eq}. Define $\alpha_{\mathrm{min}} := \min_{t,j} \alpha_j^t$, where $\alpha_j^t = \eta_j \gamma_t^2 \br{2 - \eta_j \gamma_t^2}$. Then for the robust teaching strategy given by \eqref{noise-cap-Wt-example}, after $t = \mathcal{O}\br{\br{\log \frac{1}{1 - \frac{\alpha_{\mathrm{min}}}{k}}}^{-1} \log \frac{1}{\epsilon}}$ rounds, we have $\frac{1}{N} \sum_{i=1}^N{\norm{w_i^t - w^*}^2} \leq \epsilon$. 
\end{theorem}

\subsection{SGLD Learners}
\label{sec:noisy-sgld}

Here we consider a classroom of \emph{Stochastic Gradient Langevin Dynamics} (SGLD) learners. For a given example  $(x, y) \in \mathcal{X} \times \mathcal{Y}$ at time $t$, the update rule of the student $i \in \bs{N}$ is given by 
\begin{equation}
\label{noisy-sgld-classroom-model}
w^{t+1}_i ~=~ \texttt{Proj}_{\mathcal{W}}\br{w^{t}_i - \eta_i \frac{\partial \ell\br{\ip{w^t_i}{x}, y}}{\partial w^t_i} + \sqrt{2 \eta_j \beta^{-1}} \xi_j^t} ,
\end{equation}
where $\xi_j^t \sim \mathcal{N}\br{0,I}$ is a standard Gaussian random vector in $\mathbb{R}^d$, and $\beta > 0$ is the inverse temperature parameter. We assume that the teacher has full observability of $\bc{w_i^t}_{i=1}^N$, and has full knowledge of students' learning rates $\bc{\eta_i}_{i=1}^N$, but doesn't know $\beta$. Then the teacher constructs the example as follows (depending on $H_t := \br{\bc{w_j^s}_{s=1}^t : \forall{j \in \bs{N}}}$):
\begin{align}
\hat x^t ~:=~& \argmax_{x: \norm{x} = 1} x^\top W^t x ~=~ e_1 \br{W^t} \nonumber \\
x^t ~:=~& \gamma_t \hat  x^t \text{ and } y^t = \ip{w^*}{x^t} , \label{noise-sgld-example}
\end{align}
where
\begin{align}
\gamma_t^2 ~\leq~& \frac{2}{\eta_j}, \forall{j \in \bs{N}} \label{noise-sgld-gammat} \\
\alpha_j^t ~:=~& \eta_j \gamma_t^2 \br{2 - \eta_j \gamma_t^2} \label{noise-sgld-alphajt} \\
\hat w_j^t ~:=~& {w^{t}_j - w^*} \text{ and } \label{noise-sgld-wjt} \\
W^t ~:=~& \frac{1}{N} \sum_{j=1}^N \alpha_j^t \hat w_j^t \br{\hat w_j^t}^\top . \label{noise-sgld-Wt}
\end{align}
The following theorem shows that, in this setting, the teacher can teach the classroom in expectation with linear convergence.

\begin{theorem}
\label{main-exp-theorem-noise-sgld}
Consider the classroom model given by \eqref{noisy-sgld-classroom-model}. Let $k := \max_t\bc{\mathrm{rank}\br{W^t}}$ where $W^t$ is given by \eqref{noise-sgld-Wt}. Define $\alpha_{\mathrm{min}} := \min_{t,j} \alpha_j^t$, and $\eta_{\mathrm{avg}} := \frac{1}{N}\sum_{j=1}^N{\eta_j}$, where $\alpha_j^t$ given by \eqref{noise-sgld-alphajt}. Then for the teaching strategy given by \eqref{noise-sgld-example} and for $\beta^{-1} \leq \frac{\alpha_{\mathrm{min}}}{4 \eta_{\mathrm{avg}} d^2} \epsilon$, after $t = \mathcal{O}\br{\br{\log \frac{1}{1 - \frac{\alpha_{\mathrm{min}}}{k}}}^{-1} \log \frac{1}{\epsilon}}$ rounds, we have $\mathbb{E} \bs{\frac{1}{N} \sum_{i=1}^N{\norm{w_i^t - w^*}^2}} \leq \epsilon$. 
\end{theorem}

%% file: 8.2_appendix_proofs.tex
%!TEX root = main.tex
%%%%%%%%%%%%%%%%%%%%%%%%%%%%%%%%%%%%%%%%%%%%%%%%%%%%%%%%%
%%%%%%%%%%%%%%%%%%%%%%%%%%%%%%%%%%%%%%%%%%%%%%%%%%%%%%%%%
\section{Proofs}
\begin{reptheorem}{main-exp-theorem}
\label{repthm:main-exp-theorem}
Consider the teaching strategy given in Algorithm~\ref{icta-pca-algo}. Let $k := \max_t\bc{\mathrm{rank}\br{W^t}}$, where $W^t$ is given by Eq.~\eqref{pca-main-mat}. Define $\alpha_j := \min_{t} \alpha_j^t$, $\alpha_{\mathrm{min}} := \min_{t,j} \alpha_j^t$, and $\alpha_{\mathrm{max}} := \max_{t,j} \alpha_j^t$, where $\alpha_j^t$ is given by Eq.~\eqref{alpha-j-eq}. Then after $t = \mathcal{O}\br{\br{\log \frac{1}{1 - \frac{\alpha_{\mathrm{min}}}{k}}}^{-1} \log \frac{1}{\epsilon}}$ rounds, we have $\frac{1}{N} \sum_{j=1}^N{\norm{w_j^t - w^*}^2} \leq \epsilon$. Furthermore, after $t = \mathcal{O}\br{\max \bc{\br{\log \frac{1}{1 - \alpha_j}}^{-1} \log{\frac{1}{\epsilon}}  ,  \br{\log \frac{1}{1 - \frac{\alpha_{\mathrm{min}}}{k}}}^{-1} \log{\frac{1}{\epsilon}}}}$
rounds, we have $\norm{w_j^t - w^*}^2 \leq \epsilon, \forall{j \in \bs{N}}$.
\end{reptheorem}

\begin{proof}
Let $G\br{w;x,y} = \frac{\partial \ell\br{\ip{w}{x}, y}}{\partial w}$. For the student $j \in \bs{N}$ with the update rule $w^{t+1}_j \leftarrow \texttt{Proj}_{\mathcal{W}}\br{w^t_j - \eta_j G \br{w^t_j;x,y}}$ and any input example $\br{x^t,y^t} \in \mathcal{X} \times \mathcal{Y}$ (with $y^t = \ip{w^*}{x^t}$) we have
\begin{align}
\norm{w^{t+1}_j - w^*}^2 ~\overset{(i)}{\leq}~& \norm{w^{t}_j - \eta_j G \br{w^t_j;x^t,y^t} - w^*}^2 \nonumber \\
~=~& \norm{w^{t}_j - w^*}^2 + \eta_j^2 \norm{G \br{w^t_j;x^t,y^t}}^2 - 2 \eta_j \ip{w^{t}_j - w^*}{G \br{w^t_j;x^t,y^t}} \nonumber \\
~\overset{(ii)}{=}~& \norm{w^{t}_j - w^*}^2 + \eta_j^2 \br{\ip{w^{t}_j}{x^t} - y^t}^2 \norm{x^t}^2 - 2 \eta_j \br{\ip{w^{t}_j}{x^t} - y^t} \ip{w^{t}_j - w^*}{x^t} \nonumber \\
~=~& \norm{w^{t}_j - w^*}^2 + \eta_j {\ip{w^{t}_j - w^*}{x^t}}^2 \br{\eta_j \norm{x^t}^2 - 2}  , \label{single-learner-eq}
\end{align}
where $(i)$ is by the property of projection, and $(ii)$ is due to the fact that $G \br{w;x,y} = \br{\ip{w}{x} - y} \cdot x$ for the squared loss function. Then for the example construction strategy described in Algorithm~\ref{icta-pca-algo}, we have 
\begin{align}
\frac{1}{N} \sum_{j=1}^{N}{\norm{w^{t+1}_j - w^*}^2} ~\leq~& \frac{1}{N} \sum_{j=1}^{N}{\norm{w^{t}_j - w^*}^2} + \frac{1}{N} \sum_{j=1}^{N}{\eta_j {\ip{w^{t}_j - w^*}{x^t}}^2 \br{\eta_j \norm{x^t}^2 - 2}}  \nonumber \\
~\overset{(i)}{=}~& \frac{1}{N} \sum_{j=1}^{N}{\norm{w^{t}_j - w^*}^2} + \frac{1}{N} \sum_{j=1}^{N}{\eta_j \gamma_t^2 {\ip{w^{t}_j - w^*}{\hat x^t}}^2 \br{\eta_j \gamma_t^2 - 2}}  \nonumber \\
~\overset{(ii)}{=}~& \frac{1}{N} \sum_{j=1}^{N}{\norm{w^{t}_j - w^*}^2} - \frac{1}{N} \sum_{j=1}^{N}{\alpha_j^t {\ip{{w^{t}_j - w^*}}{\hat  x^t}}^2}  \nonumber \\
~\overset{(iii)}{=}~& \frac{1}{N} \sum_{j=1}^{N}{\norm{w^{t}_j - w^*}^2} - \frac{1}{N} \sum_{j=1}^{N}{\alpha_j^t {\ip{\hat w^{t}_j}{\hat  x^t}}^2}  \nonumber \\
~=~& \frac{1}{N} \sum_{j=1}^{N}{\norm{w^{t}_j - w^*}^2} - \frac{1}{N} \sum_{j=1}^N \alpha_j^t \br{\hat  x^t}^\top \hat w_j^t \br{\hat w_j^t}^\top \hat  x^t  \nonumber \\
~\overset{(iv)}{=}~& \frac{1}{N} \sum_{j=1}^{N}{\norm{w^{t}_j - w^*}^2} - \br{\hat  x^t}^\top W^t \hat  x^t , \label{class-room-eq-12}
\end{align}
where $(i)$ is due to the fact that $x^t = \gamma_t \hat x^t$ with $\norm{\hat x^t} = 1$, $(ii)$ is due to the fact that $\br{\eta_j \gamma_t^2 - 2} < 0, \forall{j \in \bs{N}}$, $(iii)$ is by the definition of $\hat w^{t}_j$, and $(iv)$ is by the definition of $W^t$. Since $\hat x^t$ is the first principal component of $W^t$ \emph{i.e.} eigenvector corresponding to the largest eigenvalue of $W^t$, we have
\begin{align}
\br{\hat x^t}^\top W^t \hat x^t ~=~& \lambda_1 \br{W^t} \nonumber \\
~=~& \frac{\lambda_1 \br{W^t}}{\sum_{j=1}^d\lambda_j \br{W^t}} \cdot \mathrm{tr}\br{W^t} \nonumber \\
~=~& \frac{\lambda_1 \br{W^t}}{\sum_{j=1}^d\lambda_j \br{W^t}} \cdot \frac{1}{N} \sum_{j=1}^N \alpha_j^t \norm{\hat w_j^t}^2 \nonumber \\
~\geq~& \frac{1}{k} \cdot \frac{1}{N} \sum_{j=1}^N \alpha_j^t \norm{\hat w_j^t}^2 \nonumber \\
~=~& \frac{1}{k} \cdot \frac{1}{N} \sum_{j=1}^N {\alpha_j^t \norm{w_j^t - w^*}^2} \nonumber \\
~\geq~& \frac{\alpha_{\mathrm{min}}}{k} \cdot \frac{1}{N} \sum_{j=1}^N {\norm{w_j^t - w^*}^2} , \label{int-result-eq}
\end{align}
where $\alpha_{\mathrm{min}} := \min_{t,j} \alpha_j^t$, and $k := \max_t\bc{\mathrm{rank}\br{W^t}}$. From \eqref{class-room-eq-12} and \eqref{int-result-eq}, we get
\begin{equation}
\label{recurrence-avg-learner}
\frac{1}{N} \sum_{j=1}^{N}{\norm{w^{t+1}_j - w^*}^2} ~\leq~ \br{1 - \frac{\alpha_{\mathrm{min}}}{k}} \frac{1}{N} \sum_{j=1}^{N}{\norm{w^{t}_j - w^*}^2} .
\end{equation}	
That is after
\[
t+1 \geq \br{\log \frac{1}{1 - \frac{\alpha_{\mathrm{min}}}{k}}}^{-1} \log \frac{\frac{1}{N} \sum_{i=1}^N{\norm{w_i^0 - w^*}^2}}{\epsilon}
\]
iterations we get $\frac{1}{N} \sum_{i=1}^N{\norm{w_i^{t+1} - w^*}^2} \leq \epsilon$. This completes the proof of the first part of the theorem.

For any student $j \in \bs{N}$, and the example constructed in Algorithm~\ref{icta-pca-algo}, we have
\begin{align}
\norm{w^{t+1}_j - w^*}^2 ~\leq~& \norm{w^{t}_j - w^*}^2 + \eta_j {\ip{w^{t}_j - w^*}{x^t}}^2 \br{\eta_j \norm{x^t}^2 - 2} \nonumber \\
~=~& \norm{w^{t}_j - w^*}^2 + \eta_j \gamma_t^2 {\ip{w^{t}_j - w^*}{\hat x^t}}^2 \br{\eta_j \gamma_t^2 - 2} \nonumber \\
~=~& \norm{w^{t}_j - w^*}^2 - \alpha_j^t {\ip{{w^{t}_j - w^*}}{\hat  x^t}}^2 \nonumber \\
~=~& \norm{w^{t}_j - w^*}^2 - \alpha_j^t {\ip{\hat w^{t}_j}{\hat  x^t}}^2 \nonumber \\
~=~& \norm{w^{t}_j - w^*}^2 - \br{\hat  x^t}^\top W_j^t \hat  x^t \nonumber \\
~=~& \norm{w^{t}_j - w^*}^2 - \br{\hat  x^t}^\top \br{W^t - \delta_j^t} \hat  x^t \nonumber \\
~\leq~& \norm{w^{t}_j - w^*}^2 - \br{\hat  x^t}^\top W^t \hat  x^t + \max_{x: \norm{x} = 1}{x^\top \delta_j^t x} , \label{worst-case-start-eq-1}
\end{align}
where $W_j^t := \alpha_j^t \hat w_j^t \br{\hat w_j^t}^\top$ and $\delta_j^t := W^t - W_j^t$. Consider 
\begin{align}
\max_{x: \norm{x} = 1}{x^\top \delta_j^t x} ~=~& \lambda_1 \br{\delta_j^t} \nonumber \\
~\leq~& \mathrm{tr}\br{\delta_j^t} \nonumber \\ 
~=~& \mathrm{tr}\br{W^t} - \mathrm{tr}\br{W^t_j} \nonumber \\
~=~& \frac{1}{N} \sum_{i=1}^N{\alpha_i^t \norm{\hat w_i^t}^2} - \alpha_j^t \norm{\hat w_j^t}^2 \nonumber \\
~=~& \frac{1}{N} \sum_{i=1}^N{\alpha_i^t \norm{w_i^t - w^*}^2} - \alpha_j^t \norm{w_j^t - w^*}^2 \nonumber \\
~=~& \alpha_{\mathrm{max}} \cdot \frac{1}{N} \sum_{i=1}^N{\norm{w_i^t - w^*}^2} - \alpha_j \norm{w_j^t - w^*}^2, \label{worst-case-start-eq-2}
\end{align}
where $\alpha_{\mathrm{max}} := \max_{t,i}{\alpha_i^t}$ and $\alpha_j := \min_t \alpha_j^t$. Thus from \eqref{int-result-eq}, \eqref{recurrence-avg-learner}, \eqref{worst-case-start-eq-1}, and \eqref{worst-case-start-eq-2}, we get
\begin{align*}
\norm{w^{t+1}_j - w^*}^2 ~\leq~& \norm{w^{t}_j - w^*}^2 - \br{\hat  x^t}^\top W^t \hat  x^t + \max_{x: \norm{x} = 1}{x^\top \delta_j^t x} \\
~\leq~& \norm{w^{t}_j - w^*}^2 - \frac{\alpha_{\mathrm{min}}}{k} \cdot \frac{1}{N} \sum_{i=1}^N {\norm{w_i^t - w^*}^2} + \alpha_{\mathrm{max}} \cdot \frac{1}{N} \sum_{i=1}^N{\norm{w_i^t - w^*}^2} - \alpha_j \norm{w_j^t - w^*}^2 \\
~=~& \br{1 - \alpha_j} \norm{w^{t}_j - w^*}^2 + \br{\alpha_{\mathrm{max}} - \frac{\alpha_{\mathrm{min}}}{k}} \Delta_t \\
~\leq~& \br{1 - \alpha_j} \norm{w^{t}_j - w^*}^2 + \br{\alpha_{\mathrm{max}} - \frac{\alpha_{\mathrm{min}}}{k}} \br{1 - \frac{\alpha_{\mathrm{min}}}{k}}^t \Delta_0 \\
~\leq~& \br{1 - \alpha_j} \bc{\br{1 - \alpha_j} \norm{w^{t-1}_j - w^*}^2 + \br{\alpha_{\mathrm{max}} - \frac{\alpha_{\mathrm{min}}}{k}} \br{1 - \frac{\alpha_{\mathrm{min}}}{k}}^{t-1} \Delta_0} \\ 
& + \br{\alpha_{\mathrm{max}} - \frac{\alpha_{\mathrm{min}}}{k}} \br{1 - \frac{\alpha_{\mathrm{min}}}{k}}^t \Delta_0 \\
~\leq~& \br{1 - \alpha_j}^{t+1} \norm{w^{0}_j - w^*}^2 + \br{\alpha_{\mathrm{max}} - \frac{\alpha_{\mathrm{min}}}{k}} \Delta_0 \sum_{s=0}^t{\br{1 - \alpha_j}^s \br{1 - \frac{\alpha_{\mathrm{min}}}{k}}^{t-s}} ,
\end{align*}
where $\Delta_t := \frac{1}{N} \sum_{i=1}^N{\norm{w_i^t - w^*}^2}$. Since $1 - \alpha_j \leq {1 - \frac{\alpha_{\mathrm{min}}}{k}}$, we have
\begin{align*}
\sum_{s=0}^t{\br{1 - \alpha_j}^s \br{1 - \frac{\alpha_{\mathrm{min}}}{k}}^{t-s}} ~=~& \br{1 - \frac{\alpha_{\mathrm{min}}}{k}}^t \sum_{s=0}^t{\br{\frac{1 - \alpha_j}{1 - \frac{\alpha_{\mathrm{min}}}{k}}}^s} \\
 ~\leq~& \br{1 - \frac{\alpha_{\mathrm{min}}}{k}}^t \sum_{s=0}^{\infty}{\br{\frac{1 - \alpha_j}{1 - \frac{\alpha_{\mathrm{min}}}{k}}}^s} \\ 
~=~& \br{1 - \frac{\alpha_{\mathrm{min}}}{k}}^t \frac{1}{1 - \frac{1 - \alpha_j}{{1 - \frac{\alpha_{\mathrm{min}}}{k}}}} \\
~=~&\frac{1}{\alpha_j - \frac{\alpha_{\mathrm{min}}}{k}} \br{1 - \frac{\alpha_{\mathrm{min}}}{k}}^{t+1}  .
\end{align*}
Thus we get
\begin{align*}
 \norm{w^{t+1}_j - w^*}^2 ~\leq~& \br{1 - \alpha_j}^{t+1} \norm{w^{0}_j - w^*}^2 + \frac{\alpha_{\mathrm{max}}}{\alpha_j - \frac{\alpha_{\mathrm{min}}}{k}} \Delta_0 \br{1 - \frac{\alpha_{\mathrm{min}}}{k}}^{t+1} \\
~=~& \br{1 - \alpha_j}^{t+1} \norm{w^{0}_j - w^*}^2 + C_{j} \Delta_0 \br{1 - \frac{\alpha_{\mathrm{min}}}{k}}^{t+1} \\
 ~\leq~& \frac{\epsilon}{2} + \frac{\epsilon}{2}  = \epsilon ,
 \end{align*}
 where $C_{j} := \frac{\alpha_{\mathrm{max}}}{\alpha_j - \frac{\alpha_{\mathrm{min}}}{k}}$. That is we get $\norm{w^{t+1}_j - w^*}^2 \leq \epsilon$ after 
 \[
 t_j + 1 = \max \bc{\br{\log \frac{1}{1 - \alpha_j}}^{-1} \log{\frac{2 \norm{w^{0}_j - w^*}^2}{\epsilon}}  ,  \br{\log \frac{1}{1 - \frac{\alpha_{\mathrm{min}}}{k}}}^{-1} \log{\frac{2 C_{j} \Delta^0}{\epsilon}}}
 \]
 iterations. This completes the proof.
\end{proof}

\begin{reptheorem}{main-exp-theorem-noise-wjt}
\label{repthm:main-exp-theorem-noise-wjt}
Consider the noisy observation setting given by Eq.~\eqref{noisy-wjt-model-eq}.
Let $k := \max_t\bc{\mathrm{rank}\br{W^t}}$ where $W^t = \frac{1}{N} \sum_{j=1}^N{\alpha_j^t \br{\tilde{w}^{t}_j - w^*}\br{\tilde{w}^{t}_j - w^*}^\top}$. Then for the robust teaching strategy given by Eq.~\eqref{noise-wt-example}, after $t = \mathcal{O}\br{\br{\log \frac{1}{1 - \frac{\alpha_{\mathrm{min}}}{k}}}^{-1} \log \frac{1}{\epsilon}}$ rounds, we have $\frac{1}{N} \sum_{j=1}^N{\norm{w_j^t - w^*}^2} \leq \epsilon$.
\end{reptheorem}

\begin{proof}
From \eqref{single-learner-eq}, we have 
\[
\norm{w^{t+1}_j - w^*}^2 ~=~ \norm{w^{t}_j - w^*}^2 + \eta_j {\ip{w^{t}_j - w^*}{x^t}}^2 \br{\eta_j \norm{x^t}^2 - 2}  .
\]
Then for the example construction strategy described, we have
\begin{align*}
\norm{w^{t+1}_j - w^*}^2 ~\leq~& \norm{w^{t}_j - w^*}^2 + \eta_j \gamma_t^2 {\ip{w^{t}_j - w^*}{\hat x^t}}^2 \br{\eta_j \gamma_t^2 - 2} \\
~=~& \norm{w^{t}_j - w^*}^2 - \eta_j \gamma_t^2 \br{2 - \eta_j \gamma_t^2} {\ip{\tilde{w}^{t}_j - \delta - w^*}{\hat x^t}}^2 \\
~=~& \norm{w^{t}_j - w^*}^2 - \eta_j \gamma_t^2 \br{2 - \eta_j \gamma_t^2} \bc{{\ip{\tilde{w}^{t}_j - w^*}{\hat x^t}}^2 - 2 \ip{\tilde{w}^{t}_j - w^*}{\hat x^t} \ip{\delta}{\hat x^t} + {\ip{\delta}{\hat x^t}}^2} \\
~\leq~& \norm{w^{t}_j - w^*}^2 - \eta_j \gamma_t^2 \br{2 - \eta_j \gamma_t^2} {\ip{\tilde{w}^{t}_j - w^*}{\hat x^t}}^2 + 2 \eta_j \gamma_t^2 \br{2 - \eta_j \gamma_t^2} \ip{\tilde{w}^{t}_j - w^*}{\hat x^t} \ip{\delta}{\hat x^t} \\
~\leq~& \norm{w^{t}_j - w^*}^2 - \alpha_j^t {\ip{\hat{w}^{t}_j}{\hat x^t}}^2 + 2 \alpha_j^t \norm{\tilde{w}^{t}_j - w^*}\norm{\hat x^t} \norm{\delta}\norm{\hat x^t} \\
~\leq~& \norm{w^{t}_j - w^*}^2 - \alpha_j^t {\ip{\hat{w}^{t}_j}{\hat x^t}}^2 + 2 \alpha_j^t D_{\mathcal{W}}\norm{\delta} .
\end{align*}
Since 
\begin{align*}
\norm{\tilde{w}^{t}_j - w^*}^2 ~=~& \norm{w^{t}_j + \delta - w^*}^2 \\
~=~& \norm{w^{t}_j - w^*}^2 + \norm{\delta}^2 - 2 \ip{w^{t}_j - w^*}{\delta} \\
~\geq~& \norm{w^{t}_j - w^*}^2 + \norm{\delta}^2 - 2 \norm{w^{t}_j - w^*} \norm{\delta} \\
~\geq~& \norm{w^{t}_j - w^*}^2 - 2 D_{\mathcal{W}} \norm{\delta} ,
\end{align*}
we have
\begin{align*}
\frac{1}{N} \sum_{j=1}^N \norm{w^{t+1}_j - w^*}^2 ~\leq~& \frac{1}{N} \sum_{j=1}^N \norm{w^{t}_j - w^*}^2 - \frac{1}{N} \sum_{j=1}^N \alpha_j^t {\ip{\hat{w}^{t}_j}{\hat x^t}}^2 + 2 \frac{1}{N} \sum_{j=1}^N \alpha_j^t  D_{\mathcal{W}}\norm{\delta} \\
~\leq~& \frac{1}{N} \sum_{j=1}^N \norm{w^{t}_j - w^*}^2 - \frac{\alpha_\mathrm{min}}{k} \frac{1}{N} \sum_{j=1}^N \norm{\tilde{w}^{t}_j - w^*}^2 + 2 \alpha_\mathrm{avg} D_{\mathcal{W}}\norm{\delta} \\
~\leq~& \frac{1}{N} \sum_{j=1}^N \norm{w^{t}_j - w^*}^2 - \frac{\alpha_\mathrm{min}}{k} \frac{1}{N} \sum_{j=1}^N \br{\norm{w^{t}_j - w^*}^2 - 2 D_{\mathcal{W}} \norm{\delta}} + 2 \alpha_\mathrm{avg} D_{\mathcal{W}}\norm{\delta} \\
~=~& \br{1 - \frac{\alpha_\mathrm{min}}{k}} \frac{1}{N} \sum_{j=1}^N \norm{w^{t}_j - w^*}^2 + 2 \br{\frac{\alpha_\mathrm{min}}{k} + \alpha_\mathrm{avg}} D_{\mathcal{W}}\norm{\delta} \\
~\leq~& \br{1 - \frac{\alpha_\mathrm{min}}{k}}^{t+1} \frac{1}{N} \sum_{j=1}^N \norm{w^{0}_j - w^*}^2 + 2 \br{\frac{\alpha_\mathrm{min}}{k} + \alpha_\mathrm{avg}} D_{\mathcal{W}}\norm{\delta} \sum_{s=0}^{t}{\br{1 - \frac{\alpha_\mathrm{min}}{k}}^{s}} \\
~\leq~& \br{1 - \frac{\alpha_\mathrm{min}}{k}}^{t+1} \frac{1}{N} \sum_{j=1}^N \norm{w^{0}_j - w^*}^2 + 2 \br{\frac{\alpha_\mathrm{min}}{k} + \alpha_\mathrm{avg}} D_{\mathcal{W}}\norm{\delta} \frac{1}{1 - \br{1 - \frac{\alpha_\mathrm{min}}{k}}} \\
~=~& \br{1 - \frac{\alpha_\mathrm{min}}{k}}^{t+1} \frac{1}{N} \sum_{j=1}^N \norm{w^{0}_j - w^*}^2 + 2 \br{\frac{k \alpha_\mathrm{avg}}{\alpha_\mathrm{min}} + 1} D_{\mathcal{W}} \norm{\delta} \\
~\leq~& \frac{\epsilon}{2} + \frac{\epsilon}{2} = \epsilon , 
\end{align*}
where $\alpha_{\mathrm{min}} := \min_{t,j} \alpha_j^t$, and $\alpha_{\mathrm{avg}} := \max_t{\frac{1}{N} \sum_{j=1}^N{\alpha_j^t}}$. Thus for
\[
t+1 ~\geq~ \br{\log \frac{1}{1 - \frac{\alpha_{\mathrm{min}}}{k}}}^{-1} \log \frac{\frac{2}{N} \sum_{i=1}^N{\norm{w_i^0 - w^*}^2}}{\epsilon}
\]
and $\norm{\delta} \leq \frac{\epsilon}{4 \br{\frac{k \alpha_\mathrm{avg}}{\alpha_\mathrm{min}} + 1} D_{\mathcal{W}}}$, we have $\frac{1}{N} \sum_{i=1}^N{\norm{w_i^{t+1} - w^*}^2} \leq \epsilon$.
\end{proof}

\begin{reptheorem}{main-exp-theorem-noise-eta}
\label{repthm:main-exp-theorem-noise-eta}
Let $k := \max_t\bc{\mathrm{rank}\br{\bar W^t}}$ where $\bar W^t~:=~ \frac{1}{N} \sum_{j=1}^N \bar \alpha_j^t \hat w_j^t \br{\hat w_j^t}^\top$. Define $\bar \alpha_{\mathrm{min}} := \min_{t,j} \bar \alpha_j^t$, and $\beta_{\mathrm{min}} := \min_{j,t}{\frac{\alpha_j^t}{\bar \alpha_j^t}}$, where $\alpha_j^t := 2\gamma_t^2\eta_j - \gamma_t^4 \br{\sigma^2+\eta_j^2}$ and $\bar \alpha_j^t$ given by Eq.~\eqref{noise-eta-alphajt}. Then for the teaching strategy given by Eq.~\eqref{noise-eta-example}, after $t = \mathcal{O}\br{\br{\log \frac{1}{1 - \frac{\beta_{\mathrm{min}} \bar \alpha_{\mathrm{min}}}{k}}}^{-1} \log \frac{1}{\epsilon}}$ rounds, we have $\mathbb{E} \bs{\frac{1}{N} \sum_{j=1}^N{\norm{w_j^t - w^*}^2}} \leq \epsilon$.
\end{reptheorem} 

\begin{proof}
For the student $j \in \bs{N}$, with update rule $w_j^{t+1} \leftarrow \texttt{Proj}_{\mathcal{W}} \br{w_j^t - \eta_j^t G\br{w_j^t;x,y}}$, where $\eta_j^t \sim \mathcal{N}\br{\eta_j, \sigma}$, and for any example $\br{x^t , y^t}$, from (\ref{single-learner-eq}), we have
\[
\norm{w^{t+1}_j - w^*}^2 ~\leq~ \norm{w^{t}_j - w^*}^2 + \eta_j^t {\ip{w^{t}_j - w^*}{x^t}}^2 \br{\eta_j^t \norm{x^t}^2 - 2}  .
\]
Let the history up to time $t$ be $H_t := \br{\bc{w_j^s}_{s=1}^t , \bc{\eta_j^s}_{s=1}^{t-1} : \forall{j \in \bs{N}}}$, and define $\mathbb{E}_t\bs{\cdot} := \mathbb{E}\bs{\cdot \mid H_t}$. Suppose the teacher constructs the example $x^t = \gamma_t \hat x^t$ (with $\norm{\hat x^t} = 1$) based on the history $H_t$. Then given $H_t$, only $\eta_j^t$ and $w_j^{t+1}$ are random variables in the above equation \emph{i.e.} we have 
\begin{align*}
\mathbb{E}_t\bs{{\norm{w^{t+1}_j - w^*}^2}} ~\leq~&  {\norm{w^{t}_j - w^*}^2} + \mathbb{E}_t\bs{{\eta}_j^t \gamma_t^2 \br{{\eta}_j^t\gamma_t^2-2}}{\ip{w^{t}_j - w^*}{\hat x^t}}^2  \\
~=~&  {\norm{w^{t}_j - w^*}^2} +\bc{\gamma_t^4\br{\sigma^2+\eta_j^2}-2\gamma_t^2\eta_j}{\ip{w^{t}_j - w^*}{\hat x^t}}^2 \\
~=~&{\norm{w^{t}_j - w^*}^2} - {\alpha_j^t {\ip{{w^{t}_j - w^*}}{\hat  x^t}}^2} ,
\end{align*}
where $\alpha_j^t := 2\gamma_t^2\eta_j - \gamma_t^4 \br{\sigma^2+\eta_j^2}$ and $\gamma_t^2 \leq \frac{2 \eta_j}{\sigma^2 + \eta_j^2}, \forall{j \in \bs{N}}$. Thus for the classroom of students, we have
\begin{align}
\mathbb{E}_t\bs{\frac{1}{N} \sum_{j=1}^{N}{\norm{w^{t+1}_j - w^*}^2}} ~\leq~&
\frac{1}{N} \sum_{j=1}^{N}{\norm{w^{t}_j - w^*}^2} - \frac{1}{N} \sum_{j=1}^{N}{\alpha_j^t{\ip{{w^{t}_j - w^*}}{\hat  x^t}}^2}  \nonumber \\
~=~& \frac{1}{N} \sum_{j=1}^{N}{\norm{w^{t}_j - w^*}^2} - \frac{1}{N} \sum_{j=1}^{N}{\alpha_j^t{\ip{\hat w^{t}_j}{\hat  x^t}}^2}  \nonumber \\
~=~& \frac{1}{N} \sum_{j=1}^{N}{\norm{w^{t}_j - w^*}^2} - \frac{1}{N} \sum_{j=1}^N \alpha_j^t \br{\hat  x^t}^\top \hat w_j^t \br{\hat w_j^t}^\top \hat  x^t  \nonumber \\
~=~& \frac{1}{N} \sum_{j=1}^{N}{\norm{w^{t}_j - w^*}^2} - \br{\hat  x^t}^\top W^t \hat  x^t , \label{noisy-eta-meta-eq}
\end{align}
where $\hat w_j^t := \br{w^{t}_j - w^*}$ and $W^t := \frac{1}{N} \sum_{j=1}^N \alpha_j^t \hat w_j^t \br{\hat w_j^t}^\top$. The teacher constructs the example $\hat x^t$ as follows (depending only on $H_t$):
\begin{align*}
\hat x^t ~:=~& \argmax_{x: \norm{x} = 1} x^\top \bc{\frac{1}{N} \sum_{j=1}^N \bar \alpha_j^t \hat w_j^t \br{\hat w_j^t}^\top} x \\
~=~& \argmax_{x: \norm{x} = 1} x^\top \bar W^t x ~=~ e_1 \br{\bar W^t} , 
\end{align*}
where $\bar \alpha_j^t := 2\gamma_t^2 \bar \eta_j^t - \gamma_t^4 \br{\frac{t-2}{t-1}\sigma^2+\br{\bar \eta_j^t}^2}$, $\bar \eta_j^t := \frac{1}{t-1}\sum_{s=1}^{t-1}{\eta_j^s}$ and $\bar W^t := \frac{1}{N} \sum_{j=1}^N \bar \alpha_j^t \hat w_j^t \br{\hat w_j^t}^\top$. Note that $\mathbb{E}\bs{\bar \alpha_j^t} = \alpha_j^t$. For this example, from (\ref{int-result-eq}), we have 
\[
\br{\hat x^t}^\top \bar W^t \hat x^t ~\geq~ \frac{\bar \alpha_{\mathrm{min}}}{k} \cdot \frac{1}{N} \sum_{j=1}^N {\norm{w_j^t - w^*}^2},
\]
where $\bar \alpha_{\mathrm{min}} := \min_{t,j} \bar \alpha_j^t$. Since $\hat w_j^t \br{\hat w_j^t}^\top$ is a positive semidefinite matrix, we have
\begin{align*}
\br{\hat  x^t}^\top W^t \hat  x^t ~=~& \br{\hat  x^t}^\top \bc{\frac{1}{N} \sum_{j=1}^N \alpha_j^t \hat w_j^t \br{\hat w_j^t}^\top} \hat  x^t \\
~\geq~& \min_{j,t}{\frac{\alpha_j^t}{\bar \alpha_j^t}} \br{\hat  x^t}^\top \bc{\frac{1}{N} \sum_{j=1}^N \bar \alpha_j^t \hat w_j^t \br{\hat w_j^t}^\top} \hat  x^t \\
~=~& \beta_{\mathrm{min}} \br{\hat  x^t}^\top \bar W^t  \hat  x^t \\
~\geq~& \frac{\beta_{\mathrm{min}} \bar \alpha_{\mathrm{min}}}{k} \cdot \frac{1}{N} \sum_{j=1}^N {\norm{w_j^t - w^*}^2},
\end{align*}
where $\beta_{\mathrm{min}} := \min_{j,t}{\frac{\alpha_j^t}{\bar \alpha_j^t}}$. By using the above inequality in \eqref{noisy-eta-meta-eq}, we get
\[
\mathbb{E}_t\bs{\frac{1}{N} \sum_{j=1}^{N}{\norm{w^{t+1}_j - w^*}^2}} ~\leq~ \br{1 - \frac{\beta_{\mathrm{min}} \bar \alpha_{\mathrm{min}}}{k}} \frac{1}{N} \sum_{j=1}^{N}{\norm{w^{t}_j - w^*}^2} . 
\]
Then by the law of total expectation and the above recurrence relationship, we have 
\begin{align*}
\mathbb{E} \bs{\frac{1}{N} \sum_{j=1}^{N}{\norm{w^{t+1}_j - w^*}^2}} ~=~& \mathbb{E}_0 \bs{\frac{1}{N} \sum_{j=1}^{N}{\norm{w^{t+1}_j - w^*}^2}} \\
~=~& \mathbb{E}_0 \mathbb{E}_1 \cdots \mathbb{E}_{t-1} \mathbb{E}_t \bs{\frac{1}{N} \sum_{j=1}^{N}{\norm{w^{t+1}_j - w^*}^2}} \\
~\leq~& \br{1 - \frac{\beta_{\mathrm{min}} \bar \alpha_{\mathrm{min}}}{k}} \mathbb{E}_0 \mathbb{E}_1 \cdots \mathbb{E}_{t-1} \bs{\frac{1}{N} \sum_{j=1}^{N}{\norm{w^{t}_j - w^*}^2}} \\
~\leq~& \br{1 - \frac{\beta_{\mathrm{min}} \bar \alpha_{\mathrm{min}}}{k}}^{t+1} \frac{1}{N} \sum_{j=1}^{N}{\norm{w^{0}_j - w^*}^2} .
\end{align*}
That is after
\[
t+1 ~\geq~ \br{\log \frac{1}{1 - \frac{\beta_{\mathrm{min}} \bar \alpha_{\mathrm{min}}}{k}}}^{-1} \log \frac{\frac{1}{N} \sum_{j=1}^N{\norm{w_j^0 - w^*}^2}}{\epsilon}
\]
iterations we get $\mathbb{E} \bs{\frac{1}{N} \sum_{j=1}^{N}{\norm{w^{t+1}_j - w^*}^2}} \leq \epsilon$. \\

\end{proof}

\begin{reptheorem}{main-exp-theorem-noise-Wt}
\label{repthm:main-exp-theorem-noise-Wt}
Consider the noisy observation setting given by \eqref{noisy-Wt-model-eq}. Let $k := \max_t\bc{\mathrm{rank}\br{\tilde{W}^t}}$ where $\tilde{W}^t$ is given by \eqref{noisy-Wt-model-eq}. Define $\alpha_{\mathrm{min}} := \min_{t,j} \alpha_j^t$, where $\alpha_j^t = \eta_j \gamma_t^2 \br{2 - \eta_j \gamma_t^2}$. Then for the robust teaching strategy given by \eqref{noise-cap-Wt-example}, after $t = \mathcal{O}\br{\br{\log \frac{1}{1 - \frac{\alpha_{\mathrm{min}}}{k}}}^{-1} \log \frac{1}{\epsilon}}$ rounds, we have $\frac{1}{N} \sum_{i=1}^N{\norm{w_i^t - w^*}^2} \leq \epsilon$.  
\end{reptheorem}

\begin{proof}
From \eqref{class-room-eq-12}, we have
\begin{align*}
\frac{1}{N} \sum_{j=1}^{N}{\norm{w^{t+1}_j - w^*}^2} ~\leq~& \frac{1}{N} \sum_{j=1}^{N}{\norm{w^{t}_j - w^*}^2} - \br{\hat  x^t}^\top W^t {\hat  x^t},
\end{align*}
where $\alpha_j^t := \eta_j \gamma_t^2 \br{2 - \eta_j \gamma_t^2}$, $\hat w^{t}_j := {w^{t}_j - w^*}$ and $W^t := \frac{1}{N} \sum_{j=1}^N \alpha_j^t \hat w_j^t \br{\hat w_j^t}^\top$. Then for the example construction strategy described in section~\ref{subsec:noise-Wt}, we have
\begin{align*}
\frac{1}{N} \sum_{j=1}^{N}{\norm{w^{t+1}_j - w^*}^2} ~\leq~& \frac{1}{N} \sum_{j=1}^{N}{\norm{w^{t}_j - w^*}^2} - \br{\hat  x^t}^\top W^t {\hat  x^t} \\
~=~& \frac{1}{N} \sum_{j=1}^{N}{\norm{w^{t}_j - w^*}^2} - \br{\hat  x^t}^\top \br{\tilde{W}^t - \delta} {\hat  x^t} \\
~=~& \frac{1}{N} \sum_{j=1}^{N}{\norm{w^{t}_j - w^*}^2} - \br{\hat  x^t}^\top \tilde{W}^t {\hat  x^t} + \br{\hat  x^t}^\top \delta {\hat  x^t} \\
~=~& \frac{1}{N} \sum_{j=1}^{N}{\norm{w^{t}_j - w^*}^2} - \lambda_1\br{\tilde{W}^t} + \br{\hat  x^t}^\top \delta {\hat  x^t} \\
~=~& \frac{1}{N} \sum_{j=1}^{N}{\norm{w^{t}_j - w^*}^2} - \frac{\lambda_1\br{\tilde{W}^t}}{\sum_{j=1}^{d}{\lambda_j\br{\tilde{W}^t}}} \mathrm{tr}\br{\tilde{W}^t} + \br{\hat  x^t}^\top \delta {\hat  x^t} \\
~\leq~& \frac{1}{N} \sum_{j=1}^{N}{\norm{w^{t}_j - w^*}^2} - \frac{1}{k} \mathrm{tr}\br{\tilde{W}^t} + \br{\hat  x^t}^\top \delta {\hat  x^t} \\
~\leq~& \frac{1}{N} \sum_{j=1}^{N}{\norm{w^{t}_j - w^*}^2} - \frac{1}{k} \mathrm{tr}\br{W^t + \delta} + \lambda_1 \br{\delta} \\
~\leq~& \frac{1}{N} \sum_{j=1}^{N}{\norm{w^{t}_j - w^*}^2} - \frac{\alpha_{\mathrm{min}}}{k} \frac{1}{N} \sum_{j=1}^{N}{\norm{w^{t}_j - w^*}^2} + \lambda_1 \br{\delta} - \frac{1}{k} \lambda_1 \br{\delta} \\
~=~& \br{1 - \frac{\alpha_{\mathrm{min}}}{k}} \frac{1}{N} \sum_{j=1}^{N}{\norm{w^{t}_j - w^*}^2} + \br{1 - \frac{1}{k}} \lambda_1 \br{\delta} \\
~\leq~& \br{1 - \frac{\alpha_{\mathrm{min}}}{k}}^{t+1} \frac{1}{N} \sum_{j=1}^{N}{\norm{w^{0}_j - w^*}^2} + \br{1 - \frac{1}{k}} \lambda_1 \br{\delta} \sum_{s=0}^{t}{\br{1 - \frac{\alpha_{\mathrm{min}}}{k}}^{s}} \\
~\leq~& \br{1 - \frac{\alpha_{\mathrm{min}}}{k}}^{t+1} \frac{1}{N} \sum_{j=1}^{N}{\norm{w^{0}_j - w^*}^2} + \br{1 - \frac{1}{k}} \lambda_1 \br{\delta} \frac{1}{1 - \br{1 - \frac{\alpha_{\mathrm{min}}}{k}}} \\
~=~& \br{1 - \frac{\alpha_{\mathrm{min}}}{k}}^{t+1} \frac{1}{N} \sum_{j=1}^{N}{\norm{w^{0}_j - w^*}^2} + \frac{k-1}{\alpha_{\mathrm{min}}} \lambda_1 \br{\delta} \\
~\leq~& \frac{\epsilon}{2} + \frac{\epsilon}{2} ~=~ \epsilon ,
\end{align*}
where $\alpha_{\mathrm{min}} := \min_{t,j} \alpha_j^t$. Thus for
\[
t+1 ~\geq~ \br{\log \frac{1}{1 - \frac{\alpha_{\mathrm{min}}}{k}}}^{-1} \log \frac{\frac{2}{N} \sum_{i=1}^N{\norm{w_i^0 - w^*}^2}}{\epsilon}
\]
and $\lambda_1 \br{\delta} \leq \frac{\alpha_{\mathrm{min}} \epsilon}{2 \br{k-1}}$, we have $\frac{1}{N} \sum_{i=1}^N{\norm{w_i^{t+1} - w^*}^2} \leq \epsilon$.
\end{proof}

\begin{reptheorem}{main-exp-theorem-noise-sgld}
\label{repthm:main-exp-theorem-noise-sgld}
Consider the classroom model given by \eqref{noisy-sgld-classroom-model}. Let $k := \max_t\bc{\mathrm{rank}\br{W^t}}$ where $W^t$ is given by \eqref{noise-sgld-Wt}. Define $\alpha_{\mathrm{min}} := \min_{t,j} \alpha_j^t$, and $\eta_{\mathrm{avg}} := \frac{1}{N}\sum_{j=1}^N{\eta_j}$, where $\alpha_j^t$ given by \eqref{noise-sgld-alphajt}. Then for the teaching strategy given by \eqref{noise-sgld-example} and for $\beta^{-1} \leq \frac{\alpha_{\mathrm{min}}}{4 \eta_{\mathrm{avg}} d^2} \epsilon$, after $t = \mathcal{O}\br{\br{\log \frac{1}{1 - \frac{\alpha_{\mathrm{min}}}{k}}}^{-1} \log \frac{1}{\epsilon}}$ rounds, we have $\mathbb{E} \bs{\frac{1}{N} \sum_{i=1}^N{\norm{w_i^t - w^*}^2}} \leq \epsilon$.
\end{reptheorem}

\begin{proof}
For the student $j \in \bs{N}$ with the update rule $w^{t+1}_j \leftarrow \texttt{Proj}_{\mathcal{W}} \br{w^t_j - \eta_j G \br{w^t_j;x,y} + \sqrt{2 \eta_j \beta^{-1}} \xi_j^t}$ (where $\xi_j^t \sim \mathcal{N}\br{0,I}$, and $\beta > 0$) and any input example $\br{x^t,y^t} \in \mathcal{X} \times \mathcal{Y}$ (with $y^t = \ip{w^*}{x^t}$) we have
\begin{align}
\norm{w^{t+1}_j - w^*}^2 ~\overset{(i)}{\leq}~& \norm{w^{t}_j - \eta_j G \br{w^t_j;x^t,y^t} + \sqrt{2 \eta_j \beta^{-1}} \xi_j^t - w^*}^2 \nonumber \\
~=~& \norm{w^{t}_j - w^*}^2 + \norm{\sqrt{2 \eta_j \beta^{-1}} \xi_j^t - \eta_j G \br{w^t_j;x^t,y^t}}^2 + 2 \ip{w^{t}_j - w^*}{\sqrt{2 \eta_j \beta^{-1}} \xi_j^t - \eta_j G \br{w^t_j;x^t,y^t}} \nonumber \\
~=~& \norm{w^{t}_j - w^*}^2 + 2 \eta_j \beta^{-1} \norm{\xi_j^t}^2 + \eta_j^2 \norm{G \br{w^t_j;x^t,y^t}}^2 - 2 \ip{\sqrt{2 \eta_j \beta^{-1}} \xi_j^t}{\eta_j G \br{w^t_j;x^t,y^t}} \nonumber \\
& + 2 \ip{w^{t}_j - w^*}{\sqrt{2 \eta_j \beta^{-1}} \xi_j^t} - 2 \ip{w^{t}_j - w^*}{\eta_j G \br{w^t_j;x^t,y^t}} \nonumber \\
~\overset{(ii)}{=}~& \norm{w^{t}_j - w^*}^2 + 2 \eta_j \beta^{-1} \norm{\xi_j^t}^2 + \eta_j^2 \br{\ip{w^{t}_j}{x^t} - y^t}^2 \norm{x^t}^2 - 2 \eta_j \sqrt{2 \eta_j \beta^{-1}} \br{\ip{w^{t}_j}{x^t} - y^t} \ip{\xi_j^t}{x^t} \nonumber \\
& + 2 \sqrt{2 \eta_j \beta^{-1}} \ip{w^{t}_j - w^*}{\xi_j^t} - 2 \eta_j \br{\ip{w^{t}_j}{x^t} - y^t} \ip{w^{t}_j - w^*}{x^t} \nonumber \\
~=~& \norm{w^{t}_j - w^*}^2 + \eta_j {\ip{w^{t}_j - w^*}{x^t}}^2 \br{\eta_j \norm{x^t}^2 - 2} + 2 \eta_j \beta^{-1} \norm{\xi_j^t}^2 \nonumber \\
& - 2 \eta_j \sqrt{2 \eta_j \beta^{-1}} \ip{w^{t}_j - w^*}{x^t} \ip{\xi_j^t}{x^t} + 2 \sqrt{2 \eta_j \beta^{-1}} \ip{w^{t}_j - w^*}{\xi_j^t}  , \label{single-learner-eq-sgld}
\end{align}
where $(i)$ is by the property of projection, and $(ii)$ is due to the fact that $G \br{w;x,y} = \br{\ip{w}{x} - y} \cdot x$ for the squared loss function. Let the history up to time $t$ be $H_t := \br{\bc{w_j^s}_{s=1}^t : \forall{j \in \bs{N}}}$, and define $\mathbb{E}_t\bs{\cdot} := \mathbb{E}\bs{\cdot \mid H_t}$. Suppose the teacher constructs the example $x^t = \gamma_t \hat x^t$ (with $\norm{\hat x^t} = 1$) and $y^t = \ip{w^*}{x^t}$ based on the history $H_t$. Then given $H_t$, only $\xi_j^t$ and $w_j^{t+1}$ are random variables in the above equation \emph{i.e.} we have 
\begin{align*}
\mathbb{E}_t\bs{{\norm{w^{t+1}_j - w^*}^2}} ~\leq~& \norm{w^{t}_j - w^*}^2 + \eta_j {\ip{w^{t}_j - w^*}{x^t}}^2 \br{\eta_j \norm{x^t}^2 - 2} + 2 \eta_j \beta^{-1} \mathbb{E}_t\bs{\norm{\xi_j^t}^2} \\
& - 2 \eta_j \sqrt{2 \eta_j \beta^{-1}} \ip{w^{t}_j - w^*}{x^t} \ip{\mathbb{E}_t\bs{\xi_j^t}}{x^t} + 2 \sqrt{2 \eta_j \beta^{-1}} \ip{w^{t}_j - w^*}{\mathbb{E}_t\bs{\xi_j^t}} \\
~\overset{(i)}{=}~& \norm{w^{t}_j - w^*}^2 + \eta_j {\ip{w^{t}_j - w^*}{x^t}}^2 \br{\eta_j \norm{x^t}^2 - 2} + 2 \eta_j \beta^{-1} d \\
~=~& \norm{w^{t}_j - w^*}^2 - \eta_j \gamma_t^2 \br{2 - \eta_j \gamma_t^2} {\ip{w^{t}_j - w^*}{\hat x^t}}^2 + 2 \eta_j \beta^{-1} d \\
~\overset{(ii)}{=}~& \norm{w^{t}_j - w^*}^2 - \alpha_j^t {\ip{w^{t}_j - w^*}{\hat x^t}}^2 + 2 \eta_j \beta^{-1} d  ,
\end{align*}
where $(i)$ is by the facts that $\mathbb{E}_t\bs{\norm{\xi_j^t}^2} = \mathrm{tr}\br{I} = d$ and $\mathbb{E}_t\bs{\xi_j^t} = 0$, and $(ii)$ is due to $\alpha_j^t := \eta_j \gamma_t^2 \br{2 - \eta_j \gamma_t^2}$ and $\gamma_t^2 \leq \frac{2}{\eta_j}, \forall{j \in \bs{N}}$. Thus for the classroom of students, we have
\begin{align}
\mathbb{E}_t\bs{\frac{1}{N} \sum_{j=1}^{N}{\norm{w^{t+1}_j - w^*}^2}} ~\leq~&
\frac{1}{N} \sum_{j=1}^{N}{\norm{w^{t}_j - w^*}^2} - \frac{1}{N} \sum_{j=1}^{N}{\alpha_j^t{\ip{{w^{t}_j - w^*}}{\hat  x^t}}^2} + \frac{2 \beta^{-1} d}{N} \sum_{j=1}^{N}{\eta_j} \nonumber \\
~{=}~& \frac{1}{N} \sum_{j=1}^{N}{\norm{w^{t}_j - w^*}^2} - \frac{1}{N} \sum_{j=1}^{N}{\alpha_j^t{\ip{\hat w^{t}_j}{\hat  x^t}}^2}  + \frac{2 \beta^{-1} d}{N} \sum_{j=1}^{N}{\eta_j} \nonumber \\
~=~& \frac{1}{N} \sum_{j=1}^{N}{\norm{w^{t}_j - w^*}^2} - \frac{1}{N} \sum_{j=1}^N \alpha_j^t \br{\hat  x^t}^\top \hat w_j^t \br{\hat w_j^t}^\top \hat  x^t  + \frac{2 \beta^{-1} d}{N} \sum_{j=1}^{N}{\eta_j} \nonumber \\
~{=}~& \frac{1}{N} \sum_{j=1}^{N}{\norm{w^{t}_j - w^*}^2} - \br{\hat  x^t}^\top W^t \hat  x^t + \frac{2 \beta^{-1} d}{N} \sum_{j=1}^{N}{\eta_j} , \label{noisy-eta-meta-eq-sgld}
\end{align}
where $\hat w_j^t := \br{w^{t}_j - w^*}$ and $W^t := \frac{1}{N} \sum_{j=1}^N \alpha_j^t \hat w_j^t \br{\hat w_j^t}^\top$. The teacher constructs the example $\hat x^t$ as follows (depending only on $H_t$):
\begin{align*}
\hat x^t ~:=~& \argmax_{x: \norm{x} = 1} x^\top W^t x ~=~ e_1 \br{W^t} . 
\end{align*}
For this example, from (\ref{int-result-eq}), we have 
\[
\br{\hat x^t}^\top W^t \hat x^t ~\geq~ \frac{\alpha_{\mathrm{min}}}{k} \cdot \frac{1}{N} \sum_{j=1}^N {\norm{w_j^t - w^*}^2},
\]
where $\alpha_{\mathrm{min}} := \min_{t,j} \alpha_j^t$. By using the above inequality in \eqref{noisy-eta-meta-eq-sgld}, we get
\[
\mathbb{E}_t\bs{\frac{1}{N} \sum_{j=1}^{N}{\norm{w^{t+1}_j - w^*}^2}} ~\leq~ \br{1 - \frac{\alpha_{\mathrm{min}}}{k}} \frac{1}{N} \sum_{j=1}^{N}{\norm{w^{t}_j - w^*}^2} + \frac{2 \beta^{-1} d}{N} \sum_{j=1}^{N}{\eta_j}. 
\]
Then by the law of total expectation and the above recurrence relationship, we have 
\begin{align*}
\mathbb{E} \bs{\frac{1}{N} \sum_{j=1}^{N}{\norm{w^{t+1}_j - w^*}^2}} ~=~& \mathbb{E}_0 \bs{\frac{1}{N} \sum_{j=1}^{N}{\norm{w^{t+1}_j - w^*}^2}} \\
~=~& \mathbb{E}_0 \mathbb{E}_1 \cdots \mathbb{E}_{t-1} \mathbb{E}_t \bs{\frac{1}{N} \sum_{j=1}^{N}{\norm{w^{t+1}_j - w^*}^2}} \\
~\leq~& \br{1 - \frac{\alpha_{\mathrm{min}}}{k}} \mathbb{E}_0 \mathbb{E}_1 \cdots \mathbb{E}_{t-1} \bs{\frac{1}{N} \sum_{j=1}^{N}{\norm{w^{t}_j - w^*}^2}} + \frac{2 \beta^{-1} d}{N} \sum_{j=1}^{N}{\eta_j} \\
~\leq~& \br{1 - \frac{\alpha_{\mathrm{min}}}{k}}^{t+1} \frac{1}{N} \sum_{j=1}^{N}{\norm{w^{0}_j - w^*}^2} + \frac{2 \beta^{-1} d}{N} \sum_{j=1}^{N}{\eta_j} \frac{1}{1 - \br{1 - \frac{\alpha_{\mathrm{min}}}{k}}} \\
~=~& \br{1 - \frac{\alpha_{\mathrm{min}}}{k}}^{t+1} \frac{1}{N} \sum_{j=1}^{N}{\norm{w^{0}_j - w^*}^2} + \frac{2 \beta^{-1} k d \eta_{\mathrm{avg}}}{\alpha_{\mathrm{min}}} \\
~\leq~& \frac{\epsilon}{2} + \frac{\epsilon}{2} ~=~ \epsilon .
\end{align*}
That is for $\beta^{-1} \leq \frac{\alpha_{\mathrm{min}}}{4 \eta_{\mathrm{avg}} k d} \epsilon$ and after
\[
t+1 ~\geq~ \br{\log \frac{1}{1 - \frac{\alpha_{\mathrm{min}}}{k}}}^{-1} \log \frac{\frac{2}{N} \sum_{j=1}^N{\norm{w_j^0 - w^*}^2}}{\epsilon}
\]
iterations we get $\mathbb{E} \bs{\frac{1}{N} \sum_{j=1}^{N}{\norm{w^{t+1}_j - w^*}^2}} \leq \epsilon$. 
\end{proof}

%Since $\hat w_j^t \br{\hat w_j^t}^\top$ is a positive semidefinite matrix, we have
%\begin{align*}
%\br{\hat  x^t}^\top W^t \hat  x^t ~=~& \br{\hat  x^t}^\top \bc{\frac{1}{N} \sum_{j=1}^N \alpha_j^t \hat w_j^t \br{\hat w_j^t}^\top} \hat  x^t \\
%~\geq~& \min_{j,t}{\frac{\alpha_j^t}{\bar \alpha_j^t}} \br{\hat  x^t}^\top \bc{\frac{1}{N} \sum_{j=1}^N \bar \alpha_j^t \hat w_j^t \br{\hat w_j^t}^\top} \hat  x^t \\
%~=~& \beta_{\mathrm{min}} \br{\hat  x^t}^\top \bar W^t  \hat  x^t \\
%~\geq~& \frac{\beta_{\mathrm{min}} \bar \alpha_{\mathrm{min}}}{k} \cdot \frac{1}{N} \sum_{j=1}^N {\norm{w_j^t - w^*}^2},
%\end{align*}
%where $\beta_{\mathrm{min}} := \min_{j,t}{\frac{\alpha_j^t}{\bar \alpha_j^t}}$. 

%% file: 8.3_appendix_rescalable-pool-based-teaching.tex
%!TEX root = main.tex
%%%%%%%%%%%%%%%%%%%%%%%%%%%%%%%%%%%%%%%%%%%%%%%%%%%%%%%%%
%%%%%%%%%%%%%%%%%%%%%%%%%%%%%%%%%%%%%%%%%%%%%%%%%%%%%%%%%
\section{Re-scalable Pool based Teaching under Squared Loss}\label{sec:pool-based}
Here we restrict the teacher to select examples only from 
\begin{align*}
\mathcal{X} ~:=~& \bc{x: \norm{x} \leq R, x = \gamma x_i, x_i \in \mathcal{D}, \gamma \in \mathbb{R}} \\
\mathcal{Y} ~:=~& \mathbb{R} \text{ (regression) or } \bc{-1,1} \text{ (classification) } ,
\end{align*}
where $\mathcal{D} := \bc{x_1,\dots,x_m : \norm{x_i} = 1, \forall{i \in \bs{m}}}$ is a pool of directions. For teaching to be effective, the pool should contain \emph{rich enough} directions.

\subsection{Single Learner}
For the student $j \in \bs{N}$ with the update rule $w^{t+1}_j \leftarrow \texttt{Proj}_{\mathcal{W}}\br{w^t_j - \eta_j G \br{w^t_j;x,y}}$ and any input example $\br{x^t,y^t} \in \mathcal{X} \times \mathcal{Y}$ (with $x^t = \gamma_t \hat x^t$, $y^t = \ip{w^*}{x^t}$, and $\norm{\hat x^t} = 1$) we have
\begin{equation}
\label{pool-square-single-eq}
\norm{w^{t+1}_j - w^*}^2 ~\leq~ \norm{w^{t}_j - w^*}^2 - \eta_j \gamma_t^2 \br{2 - \eta_j \gamma_t^2} {\ip{w^{t}_j - w^*}{\hat x^t}}^2 .
\end{equation}
Given a pool of unit vector directions $\mathcal{D} := \bc{x_1,\dots,x_m : \norm{x_i} = 1, \forall{i \in \bs{m}}}$, the teacher constructs the example as follows 
\[
\hat x^t ~:=~ \argmax_{x \in \mathcal{D}}{\ip{w^{t}_j - w^*}{x}} .
\]
Let the optimal example of synthesis based teaching be $\hat x_{\mathrm{syn}}^t = \frac{w^{t}_j - w^*}{\norm{w^{t}_j - w^*}}$. Then for some $a_t,b_t \in \mathbb{R}$, we can decompose the example $\hat x^t$ as follows
\[
\hat x^t ~=~ \frac{a_t}{\sqrt{a_t^2 + b_t^2}} \hat x_{\mathrm{syn}}^t + \frac{b_t}{\sqrt{a_t^2 + b_t^2}} \br{\hat x_{\mathrm{syn}}^t}_{\perp} .
\]
If the pool is rich enough we would have $\abs{\frac{a_t}{\sqrt{a_t^2 + b_t^2}}} \approx 1$. Consider 
\begin{align*}
{\ip{w^{t}_j - w^*}{\hat x^t}}^2 ~=~& {\ip{w^{t}_j - w^*}{\frac{a_t}{\sqrt{a_t^2 + b_t^2}} \hat x_{\mathrm{syn}}^t + \frac{b_t}{\sqrt{a_t^2 + b_t^2}} \br{\hat x_{\mathrm{syn}}^t}_{\perp}}}^2 \\ 
~=~& {\ip{w^{t}_j - w^*}{\frac{a_t}{\sqrt{a_t^2 + b_t^2}} \hat x_{\mathrm{syn}}^t}}^2 \\
~=~& \frac{a_t^2}{{a_t^2 + b_t^2}}  {\ip{w^{t}_j - w^*}{\hat x_{\mathrm{syn}}^t}}^2 \\
~=~& \frac{a_t^2}{{a_t^2 + b_t^2}}  \norm{w^{t}_j - w^*}^2 .
\end{align*}
Then by applying the above equality in \eqref{pool-square-single-eq}, we get 
\begin{align*}
\norm{w^{t+1}_j - w^*}^2 ~\leq~& \norm{w^{t}_j - w^*}^2 - \eta_j \gamma_t^2 \br{2 - \eta_j \gamma_t^2} \frac{a_t^2}{{a_t^2 + b_t^2}}  \norm{w^{t}_j - w^*}^2 \\
~\leq~& \norm{w^{t}_j - w^*}^2 - \alpha_j^t \norm{w^{t}_j - w^*}^2 \\
~\leq~& \norm{w^{t}_j - w^*}^2 - \alpha_j \norm{w^{t}_j - w^*}^2 \\
~\leq~& \br{1 - \alpha_j}^{t+1} \norm{w^{0}_j - w^*}^2 ,
\end{align*}
where $\alpha_j^t := \frac{a_t^2}{{a_t^2 + b_t^2}}  \eta_j \gamma_t^2 \br{2 - \eta_j \gamma_t^2}$, and $\alpha_j := \min_t \alpha_j^t$.

\subsection{Classroom Setting}
For the classroom, from \eqref{class-room-eq-12}, we have 
\begin{equation}
\label{pool-square-class-eq}
\frac{1}{N} \sum_{j=1}^{N}{\norm{w^{t+1}_j - w^*}^2} ~\leq~ \frac{1}{N} \sum_{j=1}^{N}{\norm{w^{t}_j - w^*}^2} - \br{\hat  x^t}^\top W^t \hat  x^t .
\end{equation}

Given a pool of unit vector directions $\mathcal{D} := \bc{x_1,\dots,x_m : \norm{x_i} = 1, \forall{i \in \bs{m}}}$, the teacher constructs the example as follows 
\[
\hat x^t ~:=~ \argmax_{x \in \mathcal{D}}{x^\top W^t x} .
\]
Since $W^t$ is a symmetric positive semidefinite matrix:
\begin{itemize}
\item it has orthogonal eigenvectors, \emph{i.e.}, $e_i\br{W^t} \perp e_j\br{W^t}, \forall{i,j \in \bs{d}; i \neq j}$.
\item the eigenvectors span $\mathbb{R}^d$ i.e. $\mathrm{span}\bc{e_i\br{W^t}: i \in \bs{d}} = \mathbb{R}^d$.
\item for $i \neq j$: $e_i\br{W^t}^\top W^t e_j\br{W^t} = e_i\br{W^t}^\top \lambda_j\br{W^t} e_j\br{W^t} = 0$.
\end{itemize}
Let the optimal example of synthesis based teaching be $\hat x_{\mathrm{syn}}^t = e_1\br{W^t}$. Then for some $a_{t,i} \in \mathbb{R}, i \in \bs{d}$, we can decompose the example $\hat x^t$ as follows
\[
\hat x^t ~=~ \frac{1}{\sqrt{\sum_{i=1}^{d}{a_{t,i}^2}}} \sum_{i=1}^{d}{a_{t,i} e_i\br{W^t}} .
\]
If the pool is rich enough we would have $\abs{\frac{a_{t,1}}{\sqrt{\sum_{i=1}^{d}{a_{t,i}^2}}}} \approx 1$. Consider
\begin{align*}
\br{\hat  x^t}^\top W^t \hat  x^t ~=~& \frac{1}{{\sum_{i=1}^{d}{a_{t,i}^2}}} \br{\sum_{i=1}^{d}{a_{t,i} e_i\br{W^t}}}^\top W^t \br{\sum_{i=1}^{d}{a_{t,i} e_i\br{W^t}}} \\
~=~& \frac{1}{{\sum_{i=1}^{d}{a_{t,i}^2}}} \bc{\sum_{i=1}^{d}{a_{t,i}^2 e_i\br{W^t}^\top W^t e_i\br{W^t}} + \sum_{i \neq j}{a_{t,i} a_{t,j} e_i\br{W^t}^\top W^t e_j\br{W^t}}} \\
~=~& \frac{1}{{\sum_{i=1}^{d}{a_{t,i}^2}}} \sum_{i=1}^{d}{a_{t,i}^2 e_i\br{W^t}^\top W^t e_i\br{W^t}} \\
~\geq~& \frac{a_{t,1}^2}{{\sum_{i=1}^{d}{a_{t,i}^2}}} e_1\br{W^t}^\top W^t e_1\br{W^t} \\
~=~& \frac{a_{t,1}^2}{{\sum_{i=1}^{d}{a_{t,i}^2}}} \br{\hat x_{\mathrm{syn}}^t}^\top W^t \hat x_{\mathrm{syn}}^t \\
~\geq~& \frac{a_{t,1}^2}{{\sum_{i=1}^{d}{a_{t,i}^2}}} \cdot \frac{\alpha_{\mathrm{min}}}{k} \cdot \frac{1}{N} \sum_{j=1}^N {\norm{w_j^t - w^*}^2} ,
\end{align*}
where last inequality is from \eqref{int-result-eq}. Then by applying the above inequality in \eqref{pool-square-class-eq}, we get
\begin{align}
\frac{1}{N} \sum_{j=1}^{N}{\norm{w^{t+1}_j - w^*}^2} ~\leq~& \br{1 - \frac{a_{t,1}^2}{{\sum_{i=1}^{d}{a_{t,i}^2}}} \cdot \frac{\alpha_{\mathrm{min}}}{k}} \frac{1}{N} \sum_{j=1}^N {\norm{w_j^t - w^*}^2}  \nonumber\\
~\leq~& \br{1 - a_{\mathrm{min}} \cdot \frac{\alpha_{\mathrm{min}}}{k}} \frac{1}{N} \sum_{j=1}^N {\norm{w_j^t - w^*}^2} \nonumber\\
~\leq~& \br{1 - a_{\mathrm{min}} \cdot \frac{\alpha_{\mathrm{min}}}{k}}^{t+1} \frac{1}{N} \sum_{j=1}^N {\norm{w_j^0 - w^*}^2}  , \label{eq-pooled-based}
\end{align}
where $a_{\mathrm{min}} := \min_t{\frac{a_{t,1}^2}{{\sum_{i=1}^{d}{a_{t,i}^2}}}}$.

%\section{Synthesis based Teaching under other Loss Functions}

%\subsection{Absolute Loss}

%\subsection{Hinge Loss}